%% file: main.tex
\documentclass{article}

\usepackage{arxiv}

\usepackage[utf8]{inputenc}
\usepackage[T1]{fontenc}
\usepackage[numbers,square,sort&compress]{natbib}
\usepackage{hyperref}
\usepackage{url}
\usepackage{booktabs}
\usepackage[above,below]{placeins}
\usepackage{float}
\usepackage{amsfonts}
\usepackage{nicefrac}
\usepackage{microtype}
\usepackage{xcolor}
\usepackage{graphicx}
\usepackage{enumitem}
\usepackage{algorithm}
\usepackage{algorithmic}
\usepackage{tikz}
\usetikzlibrary{positioning,arrows.meta,fit,calc}
\usepackage{subcaption}
\usepackage{threeparttable}
\usepackage{multirow}
\usepackage{longtable}
\usepackage{tabularx}
\usepackage{amsmath}
\usepackage{amssymb}
\usepackage{mathtools}
\usepackage{amsthm}
\usepackage{bbm}
\usepackage[capitalize,noabbrev]{cleveref}


\newcommand{\method}{\mbox{CausalArena}}
\newcommand{\scorestd}[1]{{\textcolor{black!55}{\fontsize{5}{6}\selectfont\,(#1)}}}

\theoremstyle{plain}

\theoremstyle{definition}

\theoremstyle{remark}

\title{\texorpdfstring{\method{}: Benchmarking Causal Discovery \\in the Foundation Model Era}{CausalArena: Benchmarking Causal Discovery in the Foundation Model Era}}

\renewcommand{\shorttitle}{\method{}: Benchmarking Causal Discovery in the Foundation Model Era}

\author{%
  Zi-Rong Li, Si-Yang Liu, Tian-Zuo Wang, Han-Jia Ye\thanks{Corresponding author.} \\
  School of Artificial Intelligence, Nanjing University \\
  E-mail: \{lizr, liusy, wangtz, yehj\}@lamda.nju.edu.cn
}

\hypersetup{
  pdftitle={CausalArena: Benchmarking Causal Discovery in the Foundation Model Era},
  pdfauthor={Zi-Rong Li, Si-Yang Liu, Tian-Zuo Wang, Han-Jia Ye},
  pdfkeywords={causal discovery, benchmark, structural causal models, foundation models, tabular data},
}

\begin{document}

\maketitle

\input{abstract}

\input{introduction}

\input{related_work}

\input{benchmark_design}

\input{scm_construction}

\input{experiments}

\input{analysis}

\input{discussion}

\input{conclusion}

\bibliographystyle{unsrtnat}
\bibliography{cite}
\appendix
\input{appendix}

\end{document}

%% file: abstract.tex
\begin{abstract}
Causal discovery aims to uncover causal structures from data and is fundamental to scientific reasoning and intervention-based decision making. Its evaluation relies heavily on structural causal models (SCMs), which specify a causal graph together with the mechanisms that generate data, yet existing studies differ substantially in graph families, mechanisms, and evaluation protocols. The emergence of causal discovery foundation models (CDFMs) further complicates evaluation: performance may reflect not only causal discovery ability, but also overlap between pretraining environments and test SCMs, making results on fixed synthetic benchmarks difficult to interpret. We introduce \method{}, a unified and evolvable benchmark for causal discovery under a common protocol. Synthetic SCMs supply controlled breadth over structures and mechanisms; semantic operational SCMs provide human-auditable, semantically grounded environments beyond standard synthetic generators; and formula-grounded SCMs test discovery under explicit scientific mechanisms. Public real-world datasets provide an additional external-validity check. Experiments across classical, neural, and pretrained methods reveal substantial ranking shifts across SCM families and protocols, showing that strong performance in one benchmark regime does not reliably transfer to others. These results highlight benchmark diversity and pretraining--evaluation overlap as central challenges for evaluating causal discovery in the foundation model era.

\end{abstract}

%% file: introduction.tex
\section{Introduction}

Causal discovery aims to recover directed causal relationships from observational or interventional data and is fundamental to scientific reasoning and intervention-based decision making~\cite{books/spirtes2000causation,Peters2017}. Because the true causal graph is rarely known in real systems, empirical evaluation relies heavily on \emph{structural causal models} (SCMs), which specify a causal graph together with the mechanisms that generate data. SCMs make it possible to generate observational and interventional samples while retaining known ground-truth structure, and have therefore become a standard basis for comparing classical, neural, and pretrained causal discovery methods.

Yet causal discovery remains difficult to evaluate consistently. Different studies use different graph families, structural mechanisms, noise distributions, dimensions, sample sizes, intervention protocols, thresholding rules, and aggregation conventions. Performance can also depend strongly on the synthetic generator itself, as illustrated by generator artifacts such as varsortability~\cite{NEURIPS2021_e987eff4}. Consequently, a strong result on one benchmark does not necessarily indicate robust causal discovery across other causal environments.

\begin{figure}[!t]
    \centering
    \includegraphics[width=0.98\textwidth]{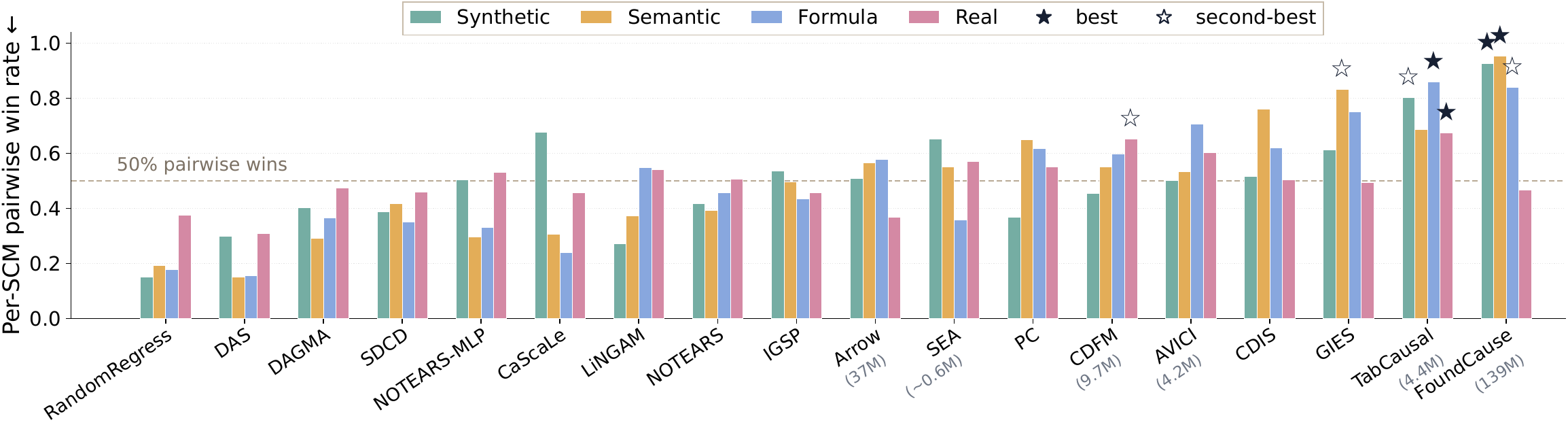}
    \caption{\textbf{Observation-only pairwise win profile across benchmark families.} For each SCM/dataset, method scores are first averaged over repeated runs. On each unit, every method pair is compared separately on F1 and SHD; higher F1 and lower SHD each count as a win, ties count as $0.5$, and the reported win rate pools wins from both metrics. Each bar reports this pooled pairwise win fraction within one benchmark family. Methods are ordered by their mean family-wise win rate from left to right; stars mark the best and second-best method within each family. Values such as \texttt{139M} under pretrained method names denote parameter counts in millions for the evaluated checkpoints.}
    \label{fig:intro-pairwise-win-profile}
\end{figure}

The emergence of causal discovery foundation models (CDFMs) makes this problem more fundamental. Methods such as AVICI~\cite{lorch2022amortizedinferencecausalstructure}, SEA~\cite{wu2025sampleestimateaggregaterecipe}, Arrow~\cite{thompson2026arrowfoundationmodelcausal}, CauScale~\cite{peng2026causcaleneuralcausaldiscovery}, CDFM~\cite{qiao2026cdfm}, FoundCause~\cite{bloebaum2026foundcause}, and TabCausal~\cite{li2026tabcausal} are pretrained over large collections of synthetic or mixed causal environments. Their benchmark performance can therefore reflect not only causal structure-learning ability, but also overlap between pretraining and evaluation---from similar graph families and mechanisms to the same or closely related SCM generators. Moreover, once a fixed SCM benchmark becomes public, it can itself become future pretraining data. Thus, in the foundation-model era, the test distribution is no longer merely an evaluation environment; it may also become part of the training distribution.

Existing resources address different parts of this problem. Synthetic frameworks such as Benchpress~\cite{rios2025benchpress} and CausalProfiler~\cite{panayiotou2025causalprofiler} support controlled and scalable evaluation, while grounded resources such as Sachs~\cite{sachs}, CausalDynamics~\cite{herdeanu2025causaldynamics}, CausalBench~\cite{chevalley2025causalbench}, and Causal Chambers~\cite{gamella2024causalchambers} provide semantic, scientific, or physical context. However, these resources typically emphasize different graph distributions, domains, tasks, or protocols, making scores difficult to compare directly across methods and especially across pretrained models.

These limitations suggest four requirements for evaluating causal discovery in the foundation-model era. \textbf{Breadth} is needed because performance should not be determined by a single synthetic prior. \textbf{Freshness} is increasingly important because no fixed public SCM suite can remain unseen indefinitely; an evaluation framework should therefore support newly constructed causal environments as pretrained models evolve. \textbf{Grounding} complements arbitrary synthetic generators with causal systems whose variables, edges, and interventions have meaningful operational or scientific interpretations. Finally, \textbf{diagnosability} requires going beyond a single leaderboard score to identify which structures, mechanisms, and environments cause methods to succeed or fail.

Motivated by these requirements, we introduce \method{}\footnote{Hugging Face dataset: \url{https://huggingface.co/datasets/LAMDA-Tabular/CausalArena}; leaderboard: \url{https://huggingface.co/spaces/LAMDA-Tabular/CausalArena-leaderboard}.}, a unified and evolvable benchmark for causal discovery. Its three SCM families play complementary roles. \emph{Synthetic SCMs} provide controlled breadth over graph structures, mechanisms, noise, dimensions, and intervention settings. \emph{Semantic operational SCMs} ground causal structures in auditable real-world processes and provide a natural basis for introducing newly authored environments in future benchmark versions. \emph{Formula-grounded scientific SCMs} anchor causal mechanisms in explicit scientific equations, making mechanism-level successes and failures easier to diagnose. All three families compile to a common executable SCM specification and share the same observational and interventional evaluation protocol; public real-world tables with published graphs serve as an additional external-validity check.

The current arena contains 1,200 executable SCM specifications---1,000 synthetic, 100 semantic operational, and 100 formula-grounded scientific SCMs---and evaluates representative classical, neural, and pretrained causal discovery methods under common data and scoring interfaces. The initial public release exposes a balanced half of each generated SCM family, while the remaining audited SCMs are reserved for held-out leaderboard evaluation and will be released progressively through versioned updates. Beyond pooled scores, we analyze performance by SCM family, graph and mechanism properties, sample size, intervention protocol, and semantic or scientific setting. The shared interface and protocol are fixed and standardized so that later rounds can introduce new SCMs, domains, and evaluation settings without rewriting how methods are run or scored.

Our experiments yield several consistent conclusions. Rankings of classical, neural, and pretrained methods shift substantially across Synthetic, Semantic, Formula, and real-data settings: no method dominates every slice, and strong performance on synthetic SCMs does not reliably transfer to semantic, scientific, or real tables. Extra samples and interventions help methods unevenly, so a single pooled score is an incomplete summary. For pretrained models especially, these patterns caution against interpreting leaderboard numbers without considering possible pretraining--evaluation overlap. Beyond ranking methods, \method{} aims to make such failures diagnosable and, through a shared executable interface, to support successive evaluation rounds that can guide and re-test progress as new algorithms and foundation models appear.

\Cref{fig:intro-pairwise-win-profile} summarizes this ranking instability under a shared observation-only protocol. Methods are ordered by mean pairwise win rate across families, yet the family-wise bars show that the leader and second place often change from Synthetic to Semantic, Formula, and real data; a method that wins broadly on one family can sit mid-pack on another. The figure therefore motivates evaluating causal discovery under multiple SCM regimes, with rankings read across families as well as in pooled form.
Our main contributions are:
\begin{itemize}[leftmargin=*,itemsep=0.15em,topsep=0.25em,parsep=0pt,partopsep=0pt]
\item We formulate causal discovery evaluation in the foundation-model era around four requirements---\textbf{breadth, freshness, grounding, and diagnosability}---highlighting the additional challenge introduced by pretraining--evaluation overlap.
\item We introduce \method{}, which combines broad synthetic SCMs, semantically grounded operational SCMs, and formula-grounded scientific SCMs under a shared observational and interventional protocol whose interface can evolve by admitting new environments without changing how methods are compared.
\item We conduct a broad evaluation of classical, neural, and pretrained causal discovery methods and provide family-, mechanism-, sample-size-, intervention-, and real-data analyses, revealing substantial ranking shifts and distinct failure profiles across evaluation environments.
\end{itemize}
\FloatBarrier

%% file: related_work.tex
\section{Related Work}
\label{sec:related_work}

\subsection{Causal Discovery Methods}
Causal discovery aims to recover directed causal structure from observational or interventional data, typically represented by a directed acyclic graph (DAG) associated with an underlying SCM~\cite{books/spirtes2000causation,Peters2017}. Classical approaches differ substantially in their assumptions and learning principles. Constraint- and search-based methods infer structure from conditional-independence relations or graph scores, including PC~\cite{books/spirtes2000causation,kalisch2005estimatinghighdimensionaldirectedacyclic,colombo2013orderindependentconstraintbasedcausalstructure}, GIES~\cite{hauser2012characterizationgreedylearninginterventional}, IGSP~\cite{wang2017permutationbasedcausalinferencealgorithms}, and CDIS~\cite{dai2025selectionmeetsinterventionadditional}. Functional approaches exploit assumptions on causal mechanisms or noise, such as LiNGAM~\cite{lingam} and DAS~\cite{pmlr-v213-montagna23b}, which builds on additive-model structure learning~\cite{buhlmann2014cam,conf/nips/HoyerJMPS08}. Continuous-optimization methods instead formulate structure learning as differentiable optimization, including NOTEARS~\cite{zheng2018dagstearscontinuousoptimization}, NOTEARS-MLP~\cite{zheng2020learning}, DAGMA~\cite{NEURIPS2022_36e2967f}, and SDCD~\cite{nazaret2024stabledifferentiablecausaldiscovery}. RandomRegress serves as a simple random-order regression control for sanity-checking scores against trivial baselines. Since these approaches rely on different structural, functional, and distributional assumptions, their relative performance can change substantially across data-generating conditions.

\subsection{Causal Discovery Foundation Models}
More recently, amortized and pretrained structure learners have shifted causal discovery toward foundation-model-style inference. AVICI~\cite{lorch2022amortizedinferencecausalstructure} amortizes structure prediction over large collections of synthetic SCMs; SEA~\cite{wu2025sampleestimateaggregaterecipe} aggregates classical discovery estimates from sampled variable subsets; Arrow~\cite{thompson2026arrowfoundationmodelcausal} factorizes graphs into skeletons and topological orders with an acyclicity guarantee; CauScale~\cite{peng2026causcaleneuralcausaldiscovery} targets neural discovery at large graph scales; CDFM~\cite{qiao2026cdfm} treats unknown mechanisms as latents in a variational foundation-model recipe; FoundCause~\cite{bloebaum2026foundcause} injects pairwise causal statistics and models latent confounding; and TabCausal~\cite{li2026tabcausal} pretrains across diverse tabular causal environments, including interventional settings. Unlike classical algorithms that are typically fitted or executed independently on each dataset, these models can encode priors learned from large synthetic or mixed causal environments. This changes the interpretation of benchmark performance: a test score may reflect not only causal structure-learning ability, but also similarity between the model's pretraining environments and the evaluation graphs, mechanisms, or generators. Comparisons among pretrained methods are therefore particularly sensitive to differences in their training distributions and evaluation protocols.

\subsection{Evaluation of Causal Discovery}

Synthetic SCMs are the dominant tool for scalable causal discovery evaluation because they generate samples together with known ground-truth graphs. Standard evaluations sample a DAG, assign structural mechanisms and noise distributions, and generate observational or interventional data~\cite{books/spirtes2000causation,Peters2017}. Many method papers consequently introduce their own synthetic evaluation suites, including those accompanying NOTEARS~\cite{zheng2018dagstearscontinuousoptimization}, DAGMA~\cite{NEURIPS2022_36e2967f}, SDCD~\cite{nazaret2024stabledifferentiablecausaldiscovery}, DCDI~\cite{brouillard2020differentiablecausaldiscoveryinterventional}, and NODAGS-Flow~\cite{sethuraman2023nodagsflownonlinearcycliccausal}. More general frameworks such as Benchpress~\cite{rios2025benchpress}, CausalProfiler~\cite{panayiotou2025causalprofiler}, and DECI's CSuite~\cite{geffner2022deci} systematize generation and comparison across broader configurations. However, synthetic evaluation is itself sensitive to generator design: varsortability and related artifacts~\cite{NEURIPS2021_e987eff4}, for example, demonstrate that methods may exploit properties of a particular data generator rather than recover causal structure in a more general sense. Complementary generator designs such as unitless unrestricted Markov-consistent (UUMC) SCM sampling~\cite{herman2025uumc} aim to reduce such nonphysical sortability patterns when constructing synthetic benchmarks.

A complementary line of work evaluates causal discovery on systems with semantic, scientific, or physical grounding. Classic resources include the Sachs protein-signaling data~\cite{sachs} and reference Bayesian networks in bnlearn~\cite{scutari2010bnlearn}, while CauseEffectPairs~\cite{mooij2016causeeffectpairs} remains a standard benchmark for bivariate cause--effect orientation, a narrower task than full multivariate DAG recovery. OCDB~\cite{zhou2024ocdb} develops real-data evaluation with graph metrics designed for more comparable structure assessment. Domain-oriented resources provide richer generative mechanisms: CausalTime~\cite{cheng2024causaltime} introduces realistically generated temporal data, CausalRivers~\cite{stein2025causalrivers} scales real-world hydrology time-series evaluation with constructed ground-truth graphs, CausalDynamics~\cite{herdeanu2025causaldynamics} targets dynamical systems derived from scientific equations, CausalBench~\cite{chevalley2025causalbench} focuses on perturbational biological networks, causalAssembly~\cite{gobler2024causalassembly} models industrial production processes, and Causal Chambers~\cite{gamella2024causalchambers} provides physical systems with controlled interventions. TimeGraph~\cite{ferdous2025timegraph} further stresses nonstationarity, irregular sampling, missingness, and latent confounding in synthetic temporal settings. These settings improve realism and mechanistic grounding, but are typically specialized to particular domains, system classes, or data modalities.

Semantic information has also become increasingly relevant as language models are introduced into causal reasoning and graph construction. CausalGraphBench~\cite{babakov2025causalgraphbench} evaluates language-model-based graph discovery against curated causal structures, while PromptBN/ReActBN~\cite{zhang2025promptbn} studies language-assisted Bayesian-network construction in low- or no-data regimes; related work surveys the use of LLMs for causal extraction, prior injection, and graph refinement~\cite{wan2025llmcdsurvey}. Such resources evaluate a different source of causal information from purely numerical SCM benchmarks, since predictions can depend on semantic priors as well as statistical evidence.

\begin{table}[t]
    \centering
    \tabcolsep 2.5pt
    \small
    \caption{
    Comparison of representative causal discovery evaluation resources and \method{}.
    Symbols summarize typical properties of the cited resources:
    \checkmark{} indicates general support, $\circ$ indicates setting-dependent support,
    and -- indicates that the property is typically absent.
    }
    \label{tab:benchmark_scope}
    \resizebox{\linewidth}{!}{%
    \begin{tabular}{lccccccc}
    \toprule
    \textbf{Resource / benchmark type} &
    \textbf{Regenerable SCM} &
    \textbf{Synthetic breadth} &
    \textbf{Grounded semantics} &
    \textbf{Scientific mechanisms} &
    \textbf{Shared protocol} &
    \textbf{Obs.+Int.} &
    \textbf{Tabular DAG} \\
    \midrule
    Real / expert graphs \cite{sachs,scutari2010bnlearn,zhou2024ocdb} &
    -- & -- & \checkmark & $\circ$ & -- & $\circ$ & \checkmark \\
    
    Method-specific synthetic \cite{zheng2018dagstearscontinuousoptimization,NEURIPS2022_36e2967f,nazaret2024stabledifferentiablecausaldiscovery,brouillard2020differentiablecausaldiscoveryinterventional,sethuraman2023nodagsflownonlinearcycliccausal} &
    \checkmark & -- & -- & -- & -- & $\circ$ & \checkmark \\
    
    Synthetic frameworks \cite{rios2025benchpress,panayiotou2025causalprofiler,geffner2022deci} &
    \checkmark & \checkmark & -- & -- & \checkmark & $\circ$ & \checkmark \\
    
    Domain simulators \cite{herdeanu2025causaldynamics,chevalley2025causalbench,cheng2024causaltime,gobler2024causalassembly} &
    \checkmark & -- & \checkmark & \checkmark & -- & $\circ$ & $\circ$ \\
    
    Time-series CD benchmarks \cite{cheng2024causaltime,stein2025causalrivers,ferdous2025timegraph} &
    $\circ$ & -- & \checkmark & \checkmark & -- & $\circ$ & -- \\
    
    Physical real-world testbeds \cite{gamella2024causalchambers} &
    -- & -- & \checkmark & \checkmark & -- & \checkmark & \checkmark \\
    
    Semantic / LLM graph resources \cite{babakov2025causalgraphbench,zhang2025promptbn} &
    -- & -- & \checkmark & -- & -- & -- & -- \\
    
    Foundation-model evaluation \cite{lorch2022amortizedinferencecausalstructure,wu2025sampleestimateaggregaterecipe,qiao2026cdfm,bloebaum2026foundcause,li2026tabcausal} &
    \checkmark & \checkmark & -- & -- & -- & $\circ$ & \checkmark \\
    
    \textbf{\method{}} &
    \checkmark & \checkmark & \checkmark & \checkmark & \checkmark & \checkmark & \checkmark \\
    \bottomrule
    \end{tabular}%
    }
\end{table}

\Cref{tab:benchmark_scope} summarizes this landscape: existing resources typically cover only a subset of regenerable SCMs, synthetic breadth, semantic or scientific grounding, a shared protocol, and joint observational--interventional tabular DAG evaluation. Synthetic frameworks emphasize breadth and protocol, but not grounding; domain simulators and physical testbeds supply meaning, but rarely a shared cross-domain protocol; foundation-model evaluations often stay within regenerable synthetic settings without operational or scientific grounding. As a result, strengths remain distributed across separate resources, and scores are hard to compare---especially for pretrained models whose training distributions may overlap with the benchmark environments. This motivates a common evaluation framework that combines these axes while retaining enough structure for detailed analysis.

%% file: benchmark_design.tex
\section{Benchmark Formulation and Design}
\label{sec:benchmark_design}

\subsection{Task and Evaluation Formulation}
\label{sec:task_formulation}

We consider causal discovery over $d$ observed variables
$\mathbf{X}=(X_1,\ldots,X_d)$ generated by an SCM $\mathcal{S}$.
An SCM specifies a DAG $G=(V,E)$ together with structural assignments
\begin{equation}
    X_j := f_j\!\left(\mathbf{X}_{\mathrm{Pa}_G(j)}, \epsilon_j\right),
    \qquad j=1,\ldots,d,
    \label{eq:scm}
\end{equation}
where $\mathrm{Pa}_G(j)$ denotes the parents of $X_j$, $f_j$ is its causal
mechanism, and $\epsilon_j$ is an exogenous noise variable
\cite{books/spirtes2000causation,Peters2017}.
Let $A_G\in\{0,1\}^{d\times d}$ denote the directed adjacency matrix of $G$.
The task is to recover $A_G$ from samples generated by $\mathcal{S}$.
Here $\mathcal{S}$ supplies the controlled evaluation environment together with
known ground-truth structure.

Each executable SCM induces an observational distribution
$P_{\mathcal{S}}(\mathbf{X})$ and, after a do-style perturbation on intervention
targets $I$, an interventional distribution
$P_{\mathcal{S}}(\mathbf{X}\mid \mathrm{do}(I))$.
We therefore construct two evaluation settings from the same SCM:
an observation-only dataset $\mathcal{D}^{\mathrm{obs}}$ and an
observation-plus-intervention dataset $\mathcal{D}^{\mathrm{obs+int}}$.
A method $\mathcal{M}$ receives one of these datasets and returns
\begin{equation}
    \widehat{A}_G = \mathcal{M}(\mathcal{D}),
    \label{eq:cd_prediction}
\end{equation}
which is scored against $A_G$ with graph-recovery metrics such as F1 and SHD.

Classical and foundation-model pipelines differ in how $\mathcal{M}$ is obtained.
A classical (or per-dataset neural) method is applied independently to each
evaluation instance,
\begin{equation}
    \widehat{A}_i = \mathcal{M}(\mathcal{D}_i),
    \label{eq:classical_prediction}
\end{equation}
so the learner is fitted or executed only on $\mathcal{D}_i$.
A CDFM instead first learns reusable parameters from a collection of
pretraining environments $\mathcal{E}_{\mathrm{pre}}$,
\begin{equation}
    \theta =
    \mathrm{Pretrain}(\mathcal{E}_{\mathrm{pre}}),
    \qquad
    \widehat{A}_i =
    \mathcal{M}_{\theta}(\mathcal{D}_i),
    \label{eq:cdfm_prediction}
\end{equation}
as in amortized and pretrained causal discovery
\cite{lorch2022amortizedinferencecausalstructure,qiao2026cdfm,li2026tabcausal}.
For a benchmark
$\mathcal{B}=\{(\mathcal{D}_i,A_i)\}_{i=1}^{N}$, a conventional aggregate score
is
\begin{equation}
    \mathrm{Score}(\mathcal{M};\mathcal{B})
    =
    \mathrm{Agg}_{i}
    \left[
        m\!\left(\widehat{A}_i,A_i\right)
    \right],
    \label{eq:benchmark_score}
\end{equation}
where $m$ is a graph-recovery metric and $\mathrm{Agg}$ aggregates over
instances and replicates.

\subsection{Design Requirements}
\label{sec:benchmark_requirements}

Eqs.~\eqref{eq:classical_prediction}--\eqref{eq:benchmark_score} make the
Introduction's difficulties precise, and they constrain what a useful
$\mathrm{Score}(\mathcal{M};\mathcal{B})$ can mean.
First, the score is defined only relative to a chosen $\mathcal{B}$: different
papers use different graph families, mechanisms, noise processes, sample sizes,
and intervention protocols, so a high score on one $\mathcal{B}$ need not
transfer to another.
Second, for a CDFM the reported quantity is
$\mathrm{Score}(\mathcal{M}_{\theta};\mathcal{B})$ with
$\theta=\mathrm{Pretrain}(\mathcal{E}_{\mathrm{pre}})$, and similarity between
$\mathcal{E}_{\mathrm{pre}}$ and $\mathcal{B}$ may occur from an exact SCM or
graph to a shared family or generator, so a high score may mix causal discovery
ability with pretraining--evaluation affinity.
Third, once a fixed public $\mathcal{B}$ is released, its SCMs can later enter
$\mathcal{E}_{\mathrm{pre}}$, and the test distribution can overlap with later
training.
If $\mathcal{B}$ is narrow, the score tracks one generator prior; if it is
fixed while $\mathcal{E}_{\mathrm{pre}}$ grows, the score can drift toward
overlap; if it contains only anonymous tables, it says little about named
processes or equations; if only a pooled value is reported, it hides where
methods succeed or fail.
\method{} is therefore organized as a single evaluation arena whose benchmark
$\mathcal{B}$ is built to answer these needs together.

\begin{figure}[!htbp]
    \centering
    \includegraphics[
        width=0.98\textwidth
    ]{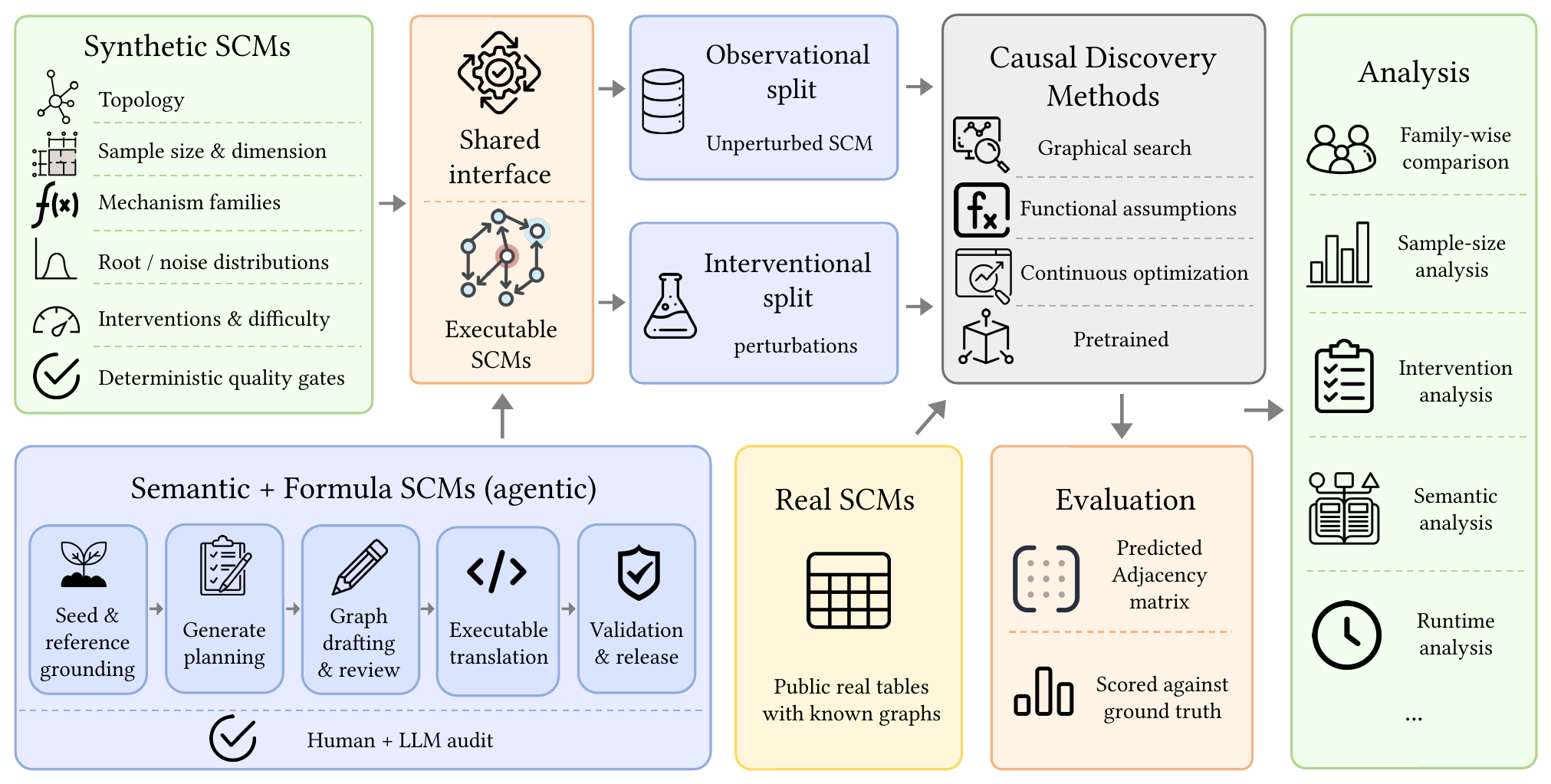}
    \caption{\textbf{Overview of the \method{} framework.}
    Synthetic, semantic operational, and formula-grounded scientific SCMs are
    compiled into executable specifications under a shared observational and
    interventional interface. Classical, neural, and pretrained causal
    discovery methods recover directed adjacency matrices, which are evaluated
    and analyzed across SCM families, mechanisms, sample sizes, intervention
    protocols, and domains. Public real-world tables with published graphs
    provide an additional external-validity check.}
    \label{fig:arena-pipeline-overview}
\end{figure}

Concretely, three SCM families---synthetic, semantic operational, and
formula-grounded scientific---are compiled into one executable specification
and exported under the same observational and interventional interfaces.
Classical, neural, and pretrained methods return directed adjacency estimates
under those interfaces; scores are reported as a pooled
$\mathrm{Score}(\mathcal{M};\mathcal{B})$ and broken down by family, mechanism,
protocol, and domain factors.
Public real-world tables with published graphs serve as an additional
external-validity check alongside the three families.
\Cref{fig:arena-pipeline-overview} summarizes this organization.

Let $\phi(\mathcal{S})$ denote construction factors of an SCM (graph family,
mechanism and noise tags, dimension, domain, intervention protocol, and related
metadata), and write slices of the arena as $\{\mathcal{B}_c\}$ indexed by
factors in $\phi$.
The constraints above motivate the following four design criteria,
stated in terms of $\phi$ and $\{\mathcal{B}_c\}$.

\paragraph{Breadth.}
Because $\mathrm{Score}(\mathcal{M};\mathcal{B})$ depends on the chosen
$\mathcal{B}$, the arena must support a diverse collection of slices
$\{\mathcal{B}_c\}$---graph families, mechanisms, noise, dimensions, sample
sizes, interventions, and difficulty---so that performance is assessed across
many causal environments.

\paragraph{Freshness.}
Because a CDFM score is $\mathrm{Score}(\mathcal{M}_{\theta};\mathcal{B})$ with
$\theta=\mathrm{Pretrain}(\mathcal{E}_{\mathrm{pre}})$, a fixed public
$\mathcal{B}$ cannot remain outside $\mathcal{E}_{\mathrm{pre}}$ indefinitely.
Freshness requires that the evaluation pool can grow: if
$\mathcal{E}_{\mathrm{pre}}$ expands, the arena should admit
$\mathcal{S}_{\mathrm{new}}$ under a shared executable interface and common
scoring rules.
\method{} therefore keeps the interface and protocol fixed and standardized, so
that newly authored SCMs can be added as the arena evolves.

\paragraph{Grounding.}
Because anonymous variation of $\phi$ alone leaves open whether
$\widehat{A}_G$ remains meaningful outside synthetic generators, some
environments must attach operational or scientific meaning to variables, edges,
and interventions.
A recovered edge is then a binary entry in $\widehat{A}_G$ and an interpretable
claim about a named process or equation.

\paragraph{Diagnosability.}
Because a pooled $\mathrm{Score}(\mathcal{M};\mathcal{B})$ leaves success and
failure unexplained, evaluation should retain $\phi(\mathcal{S})$ and report
slice scores
\begin{equation}
    \mathrm{Score}_{c}(\mathcal{M})
    =
    \mathrm{Agg}_{i:\,c(i)=c}
    \left[
        m(\widehat{A}_i,A_i)
    \right],
    \label{eq:stratified_score}
\end{equation}
so that rankings can be read as failure profiles across environments, together
with a global order.

These four requirements pull in different directions at once: breadth favors
large regenerable grids; grounding favors named processes and equations;
freshness favors newly authored environments over time; diagnosability favors
explicit factors and slice reports.
A single SCM family is unlikely to carry all four equally, which leads to the
three complementary families next.

\subsection{Three Complementary SCM Families}
\label{sec:scm_families}

\method{} assigns the requirements to three SCM families that do different
jobs, then compiles them to one executable representation so that sampling,
interventions, wrappers, and scoring stay comparable.
The point of complementarity is coverage of roles that conflict if forced into
one generator: scale and factor control on one side, operational and scientific
meaning on the other, and an authoring path for new environments as models
evolve.

\paragraph{Synthetic SCMs (breadth, and factor-level diagnosis).}
Breadth needs a regenerable factor grid: topology, mechanism and noise tags,
dimension, interventions, and difficulty can be varied systematically and at
large $d$.
Anonymous synthetic SCMs are the practical carrier of that grid.
Because $\phi(\mathcal{S})$ is explicit, the same family also supports
diagnosability through slice scores in Eq.~\eqref{eq:stratified_score}.
What they leave open is meaning: an edge is a matrix entry without an
operational or scientific story.

\paragraph{Semantic operational SCMs (operational grounding and freshness).}
Grounding needs variables and edges that mean something in a process---a policy
setting, a queue length, a sensor reading, a counted outcome.
Semantic SCMs supply that operational meaning and keep construction
human-auditable.
Because new scenarios can be authored through the same interface, they also
provide a practical route to freshness as pretrained models evolve: new
$\mathcal{S}_{\mathrm{new}}$ can be added without changing the protocol.
They do not replace the synthetic grid: operational scenarios are harder to
enumerate at the same factor resolution and scale.

\paragraph{Formula-grounded scientific SCMs (mechanistic grounding and diagnosis).}
Grounding also needs scientific or engineering equations with units and
validity ranges.
Formula SCMs keep named identities exact given their parents and place noise on
instruments and readouts, so an error is an optical, chemical, or mechanical
claim about a named mechanism.
That structure strengthens mechanism-level diagnosability on equation-backed
systems.
Like semantic SCMs, they complement the synthetic family by supplying meaning
that anonymous generators omit, at a cost in how densely $\phi$ can be swept.

Taken together, the synthetic family mainly carries \emph{breadth} and
factor-level \emph{diagnosability}; the semantic family carries operational
\emph{grounding} and a path to \emph{freshness}; the formula family strengthens
mechanistic \emph{grounding} and equation-level \emph{diagnosability}.
The shared protocol is what makes these roles jointly readable as one arena.
Public real tables remain an external wrapper check.
Construction and validation details are in \cref{sec:scm_construction}.
The shared evaluation protocol and compared methods are specified there as
well (\cref{sec:shared_protocol}).

\subsection{Evolvable Evaluation for Foundation Models}
\label{sec:cdfm_governance}

The dependence of Eq.~\eqref{eq:cdfm_prediction} on
$\mathcal{E}_{\mathrm{pre}}$ makes benchmark release fundamentally different
for CDFMs. Publishing executable SCMs is valuable for reproducibility,
inspection, and reuse, while those same SCMs can later enter pretraining
corpora.
A permanently hidden test would make construction and evaluation hard to
inspect and reproduce, and public in-distribution / out-of-distribution splits
alone leave the tension unresolved once SCMs have been released.

\method{} therefore treats pretraining overlap as part of how a result is
interpreted.
Training disclosure and labeled splits clarify what a reported score uses; the
shared executable interface and common protocol allow newly constructed graphs,
mechanisms, and scenarios to be added as the arena evolves.
In this sense, \emph{freshness} is enabled by evolving the evaluation pool under
a fixed interface.
Release policy, contamination risks, and their limitations are discussed
further in \cref{sec:discussion}.

%% file: scm_construction.tex
\section{Benchmark Implementation}
\label{sec:scm_construction}

This section turns the design in \cref{sec:benchmark_design} into an executable arena: how the three SCM families are built and validated, what the current pool covers, which method families are compared, and how they share one evaluation protocol. Concretely, each SCM specification provides a DAG $G$, structural mechanisms $\{f_j\}$, exogenous or root distributions, intervention definitions, and the metadata required for generation and analysis. Once compiled, all families use the shared sampling and scoring interfaces in \cref{sec:shared_protocol}, with wrapper-level details in \cref{sec:evaluation_protocol}.

Their construction procedures reflect their different roles in \method{}. Synthetic SCMs are generated programmatically to provide controlled breadth over causal structures and data-generating mechanisms. Semantic operational SCMs are constructed around meaningful real-world processes to provide grounding and an extensible source of newly authored environments. Formula-grounded SCMs anchor parts of the causal system in explicit scientific equations, providing mechanistic grounding and enabling more interpretable diagnosis. All SCMs pass family-specific validation and diversity checks before entering the evaluation pool.

The current arena contains 1{,}200 executable SCM specifications: 1{,}000 synthetic configurations, 100 semantic operational scenarios, and 100 formula-grounded scientific scenarios. The public Hugging Face package currently releases half of each family. The following sections describe how each family is constructed and validated, then summarize the shared protocol and compared methods.

\subsection{Synthetic SCMs: Controlled Breadth}
\label{sec:synthetic_scm}

The synthetic family is designed to systematically cover variations in the components of an SCM. We programmatically vary graph topology and dimension, structural mechanisms, root and noise distributions, dependencies among root variables, intervention settings, and difficulty factors. These factors yield different combinations of $G$, $\{f_j\}$, and noise distributions, while retaining exact ground-truth structure for evaluation.

The current pool contains 1{,}000 configurations over dimensions
$\{10,20,30,50,100\}$, spanning 10 graph families, 16 mechanism families,
14 root-distribution families, 15 noise families, 4 root-dependency families,
and 9 difficulty settings. The design includes high-dimensional configurations
for stress testing and deliberately varies factors that can otherwise become
implicit generator priors. Detailed family definitions and sampling rules are
given in \cref{sec:appendix-synthetic-pool}.

Before release, deterministic quality gates check DAG validity, mechanism
executability, numerical ranges, degenerate variables, and the intended
coverage factors. These checks ensure that controlled variation in the
generator corresponds to valid evaluation instances rather than numerical or
implementation artifacts. Programmatic audits are detailed in
\cref{sec:appendix-synthetic-audit}.

\subsection{Semantic Operational SCMs: Grounded and Extensible Environments}
\label{sec:semantic_scm}

The semantic family moves beyond unnamed synthetic variables by constructing
SCMs around operational processes with interpretable variables, causal links,
measurements, and interventions.
Nodes are labeled by what they mean in that process, and directed edges carry
reasons for why changing a parent can affect a child.
Interventions change settable fields in the same executable SCM, and
descendants are recomputed accordingly.
Each specification also stores variable definitions, measurement notes,
intervention meanings, and edge reasons, so the causal structure is
inspectable in ways that unnamed synthetic graphs are not, while retaining the
same tabular input and directed adjacency target used by the other families.

The current semantic pool contains 100 scenarios from 10 operational domains:
cybersecurity and IT operations, education, finance and credit, government
services, healthcare delivery, housing and real estate, manufacturing, public
health, urban transportation, and water and sanitation. Each domain contains
10 scenarios, with 17--25 variables and 25--67 directed edges per scenario.

Because semantic SCMs are authored through a reusable construction pipeline,
new operational scenarios can later be added without changing the evaluation
interface. This property provides a practical route toward the evolvability
and future freshness discussed in \cref{sec:benchmark_requirements}; the
current release itself does not assume that semantic knowledge is unseen by
all pretrained models.

\begin{figure}[!htbp]
    \centering
    \includegraphics[width=0.98\textwidth]{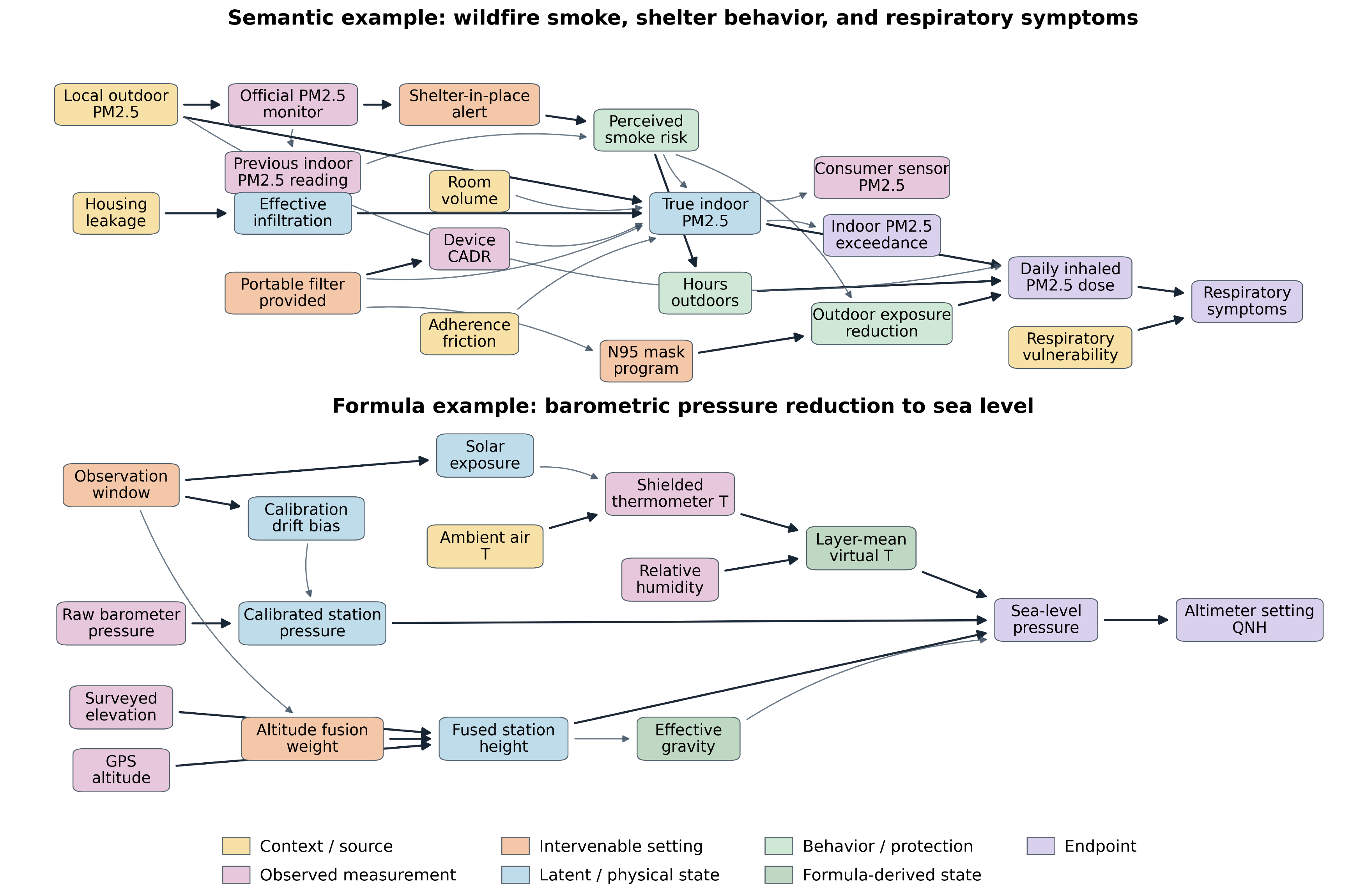}
    \caption{\textbf{Example Semantic and Formula SCMs (full released cards).}
    Node colors follow the legend, and arrows indicate directed causal edges.
    Top: wildfire smoke, shelter behavior, and respiratory symptoms
    (public health; 20 variables, 26 edges). Bottom: barometric pressure
    reduction to sea level (earth systems; 16 variables, 18 edges).}
    \label{fig:semantic-formula-examples}
\end{figure}

\paragraph{Example.}
\Cref{fig:semantic-formula-examples}~(top) is the public-health scenario
\emph{Wildfire smoke and shelter behavior}: 20 variables and 26 edges.
Think of one smoke day at one home, and read the panel left to right.
Local outdoor PM2.5 is the true air near the house.
The official PM2.5 monitor sees the same plume, but not perfectly, so its
reading can differ from the local value.
If that monitor reading is high enough, a shelter-in-place alert may be
issued (this alert can be changed).
The alert and the previous indoor PM2.5 reading together shape perceived
smoke risk.
Higher perceived risk then shortens hours outdoors and increases outdoor
exposure reduction (for example through the N95 mask program).
A second path is about the building.
Housing leakage sets effective infiltration.
Providing a portable filter raises device CADR (clean-air delivery rate).
Room volume and adherence friction also matter.
Local outdoor PM2.5, infiltration, filtration, and behavior then determine
true indoor PM2.5.
From that true indoor level, a consumer sensor gives a noisy reading, and an
indoor exceedance flag is set.
Daily inhaled PM2.5 dose combines hours outdoors, outdoor exposure reduction,
true indoor PM2.5, and some direct outdoor contribution.
Respiratory vulnerability then turns that dose into respiratory symptoms.
Edges therefore have a plain reading: removing
``portable filter \(\rightarrow\) true indoor PM2.5'' drops a real control
people can change, and treating the consumer-sensor reading as the true
indoor level mixes a measurement with the physical state.
The executable SCM keeps the variable definitions, interventions,
measurements, and edge justifications.

\subsection{Formula-Grounded Scientific SCMs: Mechanistic Grounding}
\label{sec:formula_scm}

The formula-grounded family builds SCMs in which some mechanisms are named
scientific or engineering equations, with units and valid ranges written down.
Each scenario starts from one or more such equations together with the
surrounding measurement and correction steps.
Those equations define selected structural mechanisms $f_j$.
Formula nodes are computed exactly from their parents. Randomness sits on
inputs, instrument readings, or other non-formula quantities, not on
arbitrarily jittering the equation itself.
After sampling, residual checks verify that rows that were not intervened still
match the scientific relation.

The current pool contains 100 scenarios spanning 10 scientific and engineering
domains: biology and ecology, astronomy, chemistry, mechanics and fluids,
earth systems, electromagnetism, energy systems, materials and structures,
optics and waves, and thermodynamics. Each domain contributes 10 scenarios,
with 16--25 variables and 18--50 directed edges per scenario.

Explicit equations do more than make the setting look scientific: they write
down known mechanism structure that later analysis can check against.
The formula family therefore supports both scientific grounding and the
mechanism-level diagnosability emphasized in
\cref{sec:benchmark_requirements}. Family-specific validation checks formula
orientation, units, numerical validity ranges, and equation residuals.

\paragraph{Example.}
\Cref{fig:semantic-formula-examples}~(bottom) is the earth-systems scenario
\emph{Barometric pressure profile}: 16 variables and 18 edges.
The panel has three short chains that meet in one sea-level pressure formula.
First, pressure.
A raw barometer pressure reading is corrected by a calibration drift bias.
That drift depends on the observation window (day vs.\ night, changeable),
because temperature cycling shifts the instrument offset.
Raw pressure and drift together yield calibrated station pressure.
Second, height.
Surveyed elevation and GPS altitude are combined with an altitude fusion
weight (also changeable) into fused station height, which then sets
effective gravity.
Third, temperature.
The same observation window sets solar exposure.
Ambient air temperature, biased by solar exposure, becomes the shielded
thermometer reading.
Relative humidity and that shielded temperature then determine layer-mean
virtual temperature.
Finally, calibrated station pressure \(p_{\mathrm{stn}}\), fused height
\(z\), effective gravity \(g_{\mathrm{eff}}\), and layer-mean virtual
temperature \(T_v\) enter the hypsometric sea-level reduction
\[
\mathrm{SLP}
=
p_{\mathrm{stn}}
\exp\!\left(\frac{g_{\mathrm{eff}}\,z}{R_d\,T_v}\right),
\]
with dry-air gas constant \(R_d\).
Altimeter setting QNH is that sea-level pressure converted to hPa and
rounded.
The card therefore mixes exact named equations with ordinary instrument and
calibration steps: changing the fusion weight or observation window
recomputes later nodes in the same executable SCM, and residual checks keep
non-intervened formula rows consistent with the reduction equation.

\subsection{Construction Workflow, Validation, and Diversity Control}
\label{sec:agentic_pipeline}
\label{sec:scm_quality}

Semantic and formula-grounded SCMs share a staged agentic construction
workflow consisting of seed and reference grounding, scenario planning, graph
drafting, domain review, executable translation, adversarial review,
deterministic validation, and release gating. Human and LLM-assisted review
focuses on variable necessity, causal-edge rationale, intervention meaning,
measurement realism, and potential template or alias artifacts. The
formula-grounded family additionally verifies equation orientation, units,
valid ranges, and residual consistency.

The synthetic family does not require the same semantic authoring process;
instead, deterministic programmatic checks validate graph structure,
mechanisms, distributions, numerical behavior, and coverage constraints.
Across all three families, diversity control removes redundant graphs,
near-duplicate scenarios, repeated templates, unsupported edges, and
low-information variables before inclusion in the evaluation pool.

These family-specific procedures ultimately produce the same executable SCM
interface. Consequently, differences across the three families arise from
their causal environments, while sampling, interventions, method wrappers,
adjacency targets, and scoring remain shared. Construction ablations in \cref{sec:experiments} examine the effects of
reference grounding, planning, and graph review, while the complete staged
workflow and release gates are provided in
\cref{sec:appendix-agentic-workflow}.

\subsection{Shared Protocol and Compared Methods}
\label{sec:shared_protocol}
\label{sec:implementation_methods}

All families are scored under one shared protocol.
For every executable SCM $\mathcal{S}_i$, \method{} stores the ground-truth
adjacency $A_i$ and exports two matched tables from the same graph: an
observation-only split $\mathcal{D}^{\mathrm{obs}}$ from the unperturbed SCM,
and an observation-plus-intervention split $\mathcal{D}^{\mathrm{obs+int}}$
obtained by do-style perturbations with descendant recomputation.
Intervention targets are recorded in masks shipped with the data.
On the main leaderboard, the observation-only split uses
$n_{\mathrm{obs}}=1000$ rows; the mixed split uses
$n_{\mathrm{obs}}=800$ observational and $n_{\mathrm{int}}=200$ interventional
rows.
Methods that do not consume intervention indicators are run only on supported
splits, and unsupported cells are recorded as missing.

Every method wrapper receives standardized tabular inputs and returns a
directed adjacency estimate.
Each SCM is evaluated with multiple independent replicates (five in the current
main protocol); replicate scores are averaged before aggregation over families,
domains, dimensions, or the full arena.
Failed, timed-out, or invalid runs are recorded explicitly.
Primary metrics are directed-edge precision, recall, F1, and SHD relative to
$A_i$; slice scores follow Eq.~\eqref{eq:stratified_score}, so rankings can be
read by family, mechanism, domain, sample size, or intervention protocol as
well as in pooled form.
Thresholding, CPDAG/PAG conversion rules, runtime budgets, and wrapper defaults
are specified in \cref{sec:evaluation_protocol}.

Under this protocol the arena compares 18 methods with common wrappers: a
random-order regression control (RandomRegress); graphical search (CDIS, GIES,
IGSP, and PC); functional-assumption methods (DAS and LiNGAM); continuous
optimization (DAGMA, NOTEARS, NOTEARS-MLP, and SDCD); and pretrained or
amortized methods (Arrow, AVICI, CauScale, CDFM, FoundCause, SEA, and
TabCausal).
The next section reports what these methods do under the current release.

%% file: experiments.tex
\section{Results and Observations}
\label{sec:experiments}

We evaluate the methods and SCM pool from \cref{sec:scm_construction} under the shared protocol in \cref{sec:evaluation_protocol}. The focus here is on observations: how rankings move across families and splits, how real tables and protocol choices change those rankings, and what that implies for foundation-model evaluation. Factor-level and Semantic/Formula analyses follow in \cref{sec:analysis}.

\subsection{Main Results}

\Cref{fig:intro-pairwise-win-profile} previews cross-family ranking instability under the shared observation-only protocol. For each SCM or real dataset, we first average F1 and SHD \emph{separately} over repeated runs, yielding one score per metric per unit. The intro figure then compares every method pair on F1 and on SHD independently (higher F1 wins; lower SHD wins; ties count as $0.5$) and reports the pooled win fraction over both metric-wise comparisons within each family (full matrix in \cref{fig:appendix-pairwise-win-heatmap}). The tables and plots below use the same per-unit, replicate-averaged scores, summarized as family-level means.

\begin{figure}[!htbp]
\centering
\includegraphics[width=0.82\textwidth]{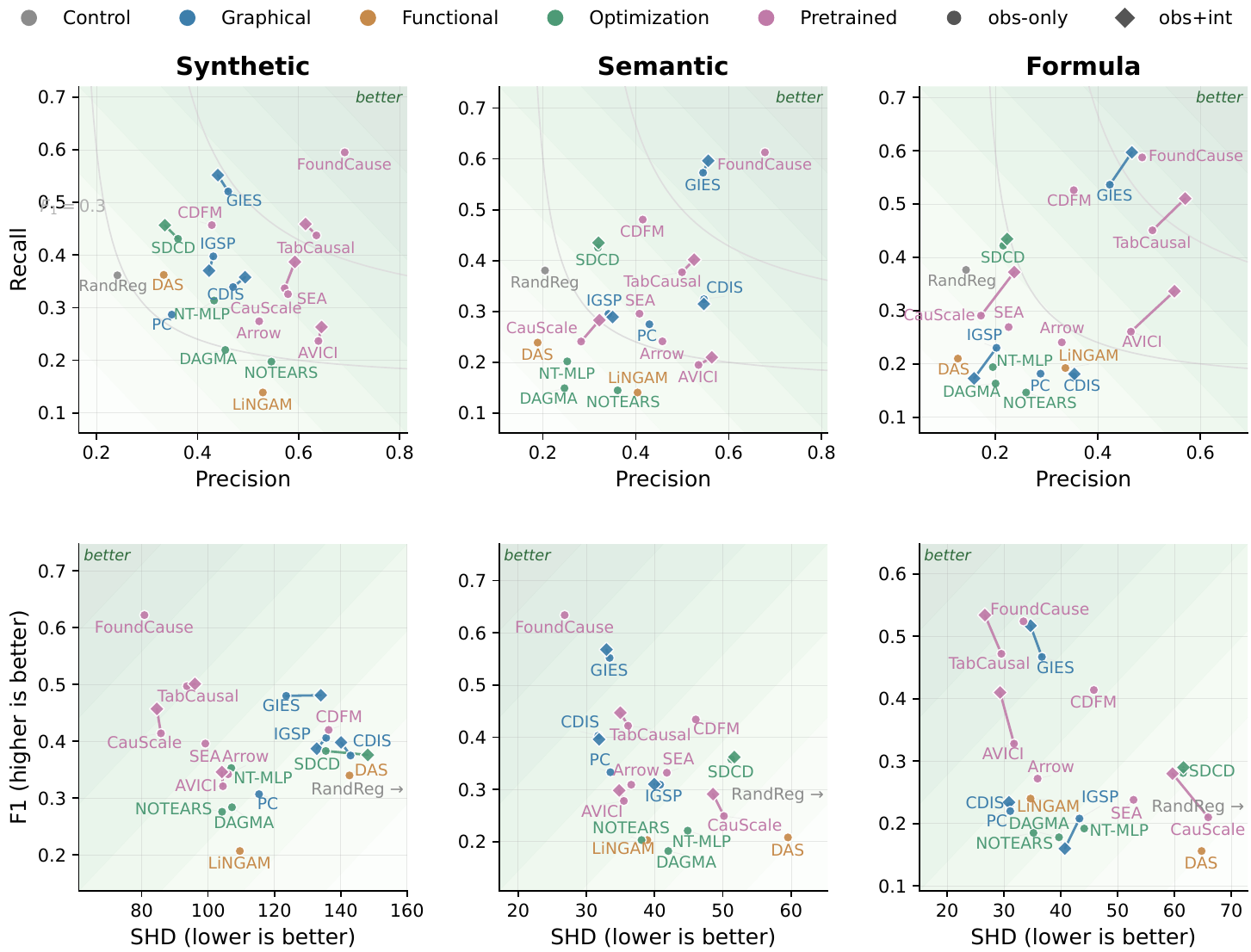}
\caption{\textbf{Precision--recall and F1--SHD on the main benchmark.} Top row: directed-edge precision versus recall. Bottom row: F1 versus SHD. Circles are obs-only, diamonds are obs+int, and a line joins the two when both exist. Shaded corners and the italic \emph{better} marker indicate the preferred region (higher precision and recall; higher F1 and lower SHD). Axes are cropped to the plotted methods; RandomRegress is marked at the right edge when its SHD lies off-scale. Each method is labeled once; \emph{RandReg} is RandomRegress and \emph{NT-MLP} is NOTEARS-MLP. Gray curves are constant F1. SHD axes are independent across families.}
\label{fig:main-leaderboard-precision-recall}
\end{figure}

\cref{tab:main-compact-f1-shd} compares all evaluated methods on Synthetic, Semantic, and Formula benchmarks. \Cref{fig:main-leaderboard-precision-recall} plots directed-edge precision against recall and F1 against SHD, with obs-only and obs+int shown together. Complete method-wise metrics are in \cref{sec:appendix-complete-synthetic} for Synthetic, \cref{sec:appendix-complete-semantic} for Semantic, and \cref{sec:appendix-complete-formula} for Formula.

\input{tables/generated_results/paper_compact/main_compact_f1_shd}

The main observation is that rankings depend strongly on the evaluation environment. FoundCause is strongest among the supported observation-only slices (F1 $0.62$/$0.63$/$0.52$ on Synthetic/Semantic/Formula), GIES is competitive on several intervention-aware slices (best Semantic obs+int F1 $0.57$), and TabCausal is among the stronger pretrained methods, especially with interventional evidence (Formula obs+int F1 $0.53$). No method wins every slice or metric: F1 and SHD sometimes disagree, and the order changes across Synthetic, Semantic, and Formula. A single pooled score would hide these moves.

\Cref{fig:main-leaderboard-precision-recall} makes the same pattern geometric. Most methods occupy a mid-precision, mid-recall band; FoundCause sits farther toward the high-F1 corner on observation-only slices, while RandomRegress is the high-SHD outlier (off-scale on the bottom row). The obs+int setting is a fixed-budget mixed-evidence protocol: it replaces 200 observational rows with interventional rows instead of adding interventions on top of the same observational sample. Thus, moving from obs-only to obs+int changes both the information type and the sample composition, producing modest gains or trade-offs between F1 and SHD depending on the method. Semantic is more spread out than Formula, where several methods collapse toward similar precision--recall and F1--SHD values. Factor-level and semantic analyses in \cref{sec:analysis} explain these shifts.

\subsection{Real-data Check}
\label{sec:real_data_check}

\begin{figure}[!htbp]
\centering
\begin{minipage}[t]{0.40\textwidth}
\vspace{0pt}
\centering
\begin{minipage}[c][0.245\textheight][c]{\linewidth}
\centering
\resizebox{\linewidth}{!}{%
\input{tables/generated_results/paper_compact/real_data_compact_side}}
\end{minipage}
\end{minipage}%
\hspace{0.8em}%
\begin{minipage}[t]{0.57\textwidth}
\vspace{0pt}
\centering
\begin{minipage}[c][0.245\textheight][c]{\linewidth}
\centering
\includegraphics[width=\linewidth]{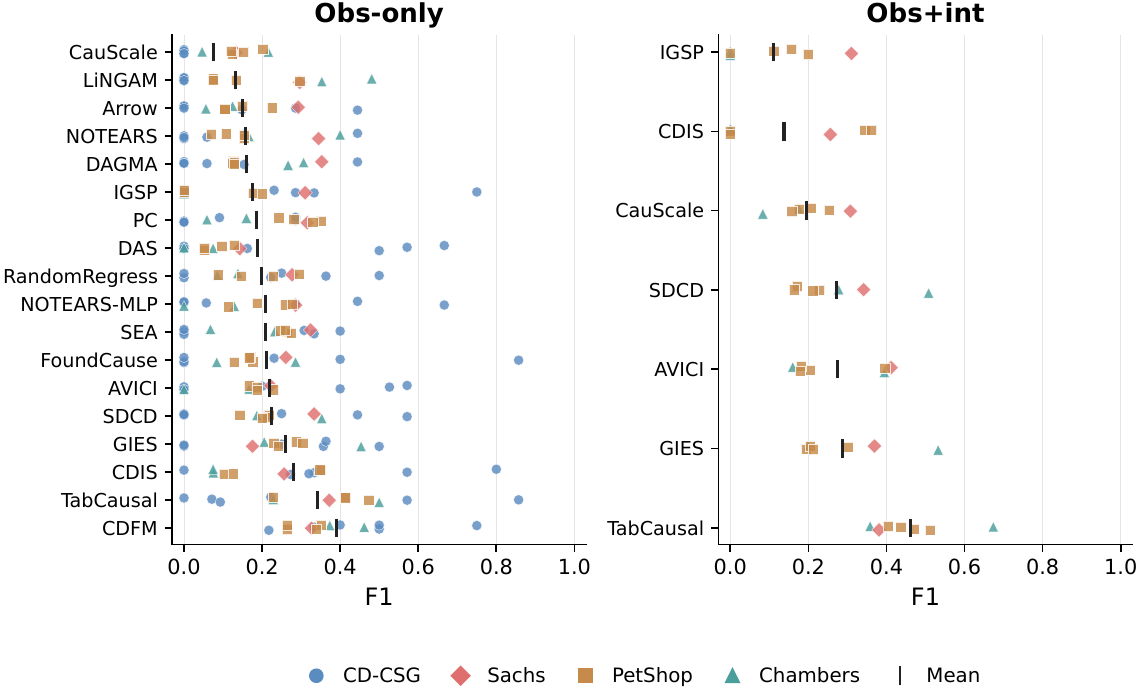}
\end{minipage}
\end{minipage}
\caption{\textbf{Real-data check.} Left: compact F1 / SHD results, with cells reporting mean\scorestd{std} over datasets (obs-only: 13; obs+int: 7) and SHD shown to one decimal for space. Right: per-dataset real-data F1; each marker is one scored table, and the tick is the method mean.}
\label{fig:real-data-check}
\end{figure}

\Cref{fig:real-data-check} evaluates the same wrappers on real scored tabular datasets with public samples and published directed graphs. The observation-only sources are six CD-CSG datasets~\cite{zhao2023causalstar}: Abalone, Auto MPG, Cardiac Arrhythmia, Concrete Compressive Strength, Deutscher Wetterdienst, and Ozone. The interventional sources are Sachs flow-cytometry~\cite{sachs}; PetShop high traffic, low traffic, temporal traffic 1, and temporal traffic 2~\cite{hardt2024petshop}; and the Causal Chambers Light Tunnel and Wind Tunnel experiments~\cite{gamella2024causalchambers}. The observation-only aggregate combines these six CD-CSG datasets with observation-only conversions of the seven interventional datasets, obtained by dropping intervention rows; the interventional aggregate uses the seven datasets with intervention rows.

On these tables the synthetic ranking does not carry over. CDFM leads observation-only F1 ($0.39$), TabCausal is second ($0.34$) and then leads observation-plus-intervention ($0.46$), while FoundCause falls to $0.21$. \Cref{fig:real-data-check} shows why the means are only a coarse summary: within a method the dataset markers span a wide F1 range, so a few tables dominate both the average and the standard deviation. Because obs-only averages include six additional CD-CSG datasets, intervention effects should be read on the seven paired interventional sources: TabCausal, AVICI, CauScale, and SDCD improve there, GIES and IGSP are nearly flat, and CDIS drops. This track checks whether the same wrappers still recover published graphs on public real tables, next to the executable SCM arena. Those graphs are expert-specified or experimentally motivated in the source papers; they are not independently verified unique true DAGs. Complete method-wise and per-dataset metrics are in \cref{sec:appendix-complete-real}.
\subsection{Sample Size and Intervention Protocol}
\label{sec:protocol_sensitivity_experiments}

\begin{figure}[!htbp]
\centering
\begin{subfigure}[b]{0.485\textwidth}
\centering
\includegraphics[width=\linewidth]{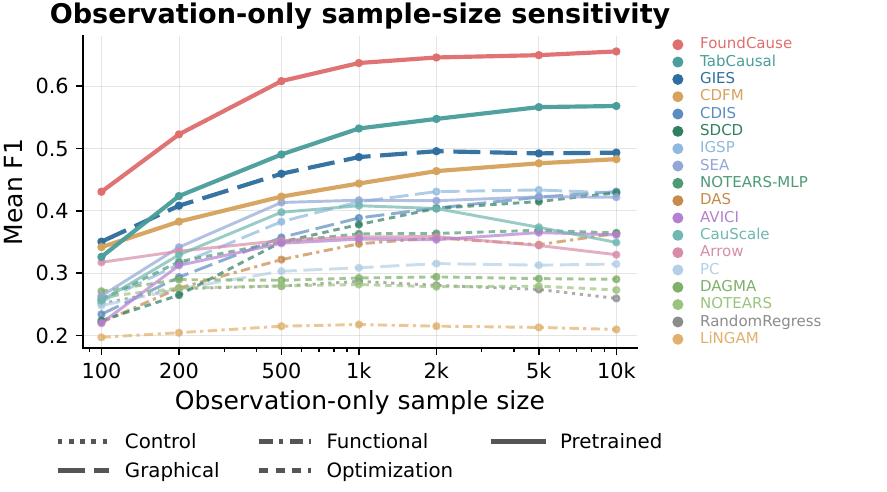}
\phantomsubcaption\label{fig:sample-scaling-f1}
\end{subfigure}
\hfill
\begin{subfigure}[b]{0.485\textwidth}
\centering
\includegraphics[width=\linewidth]{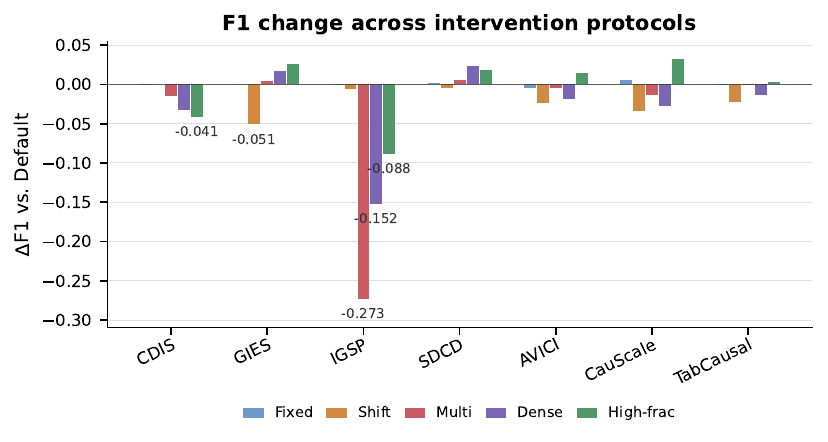}
\phantomsubcaption\label{fig:intervention-protocol-profiles}
\end{subfigure}
\caption{\textbf{Sample-size and intervention-protocol sensitivity.} (a) Mean F1 on the synthetic $d{=}30$ observation-only suite; line styles encode method class, and the right column lists full method names ordered by final F1. (b) F1 change relative to the default mixed protocol for each method that supports the intervention-protocol suite; negative values are drops.}
\label{fig:sample-and-intervention-sensitivity}
\end{figure}

These experiments keep the synthetic SCM collection fixed and change sample size or intervention design. The sample-size suite fixes $d=30$, uses 100 synthetic SCMs, and varies the observation-only sample size $n\in\{100,200,500,1000,2000,5000,10000\}$. \Cref{fig:sample-scaling-f1} plots F1 against $n$ for all evaluated methods. The corresponding F1/SHD table is in \cref{sec:appendix-complete-sample}.

Most of the F1 gain occurs between $n=100$ and $n=1{,}000$; beyond $n=2{,}000$ the curves flatten. FoundCause remains highest throughout ($0.43$ at $n=100$, $0.64$ at $n=1{,}000$, $0.66$ at $n=10{,}000$), and TabCausal has the next largest rise ($0.33$ to $0.57$). These absolute levels should be read with the suite's fixed $d=30$ SCM mix in mind: the curves isolate how F1 changes with $n$, while a method's height can also reflect affinity to that held-fixed graph, mechanism, and noise composition. Search and optimization methods such as GIES, IGSP, CDIS, SDCD, and NOTEARS-MLP improve through $n=1{,}000$ and then stall. NOTEARS, DAGMA, LiNGAM, and RandomRegress barely move. Arrow and CauScale are not monotone: both peak near $n=1{,}000$--$2{,}000$ and then drop at $n=10{,}000$ (Arrow $0.36\to 0.33$, CauScale $0.41\to 0.35$), so extra observational samples do not uniformly help pretrained wrappers.

The intervention-protocol suite also fixes $d=30$ with 100 SCMs and compares the default resampling intervention with fixed-setpoint, parameter-shift, high-intervention-fraction, multi-target-per-sample, and dense mixed-intervention settings. \Cref{tab:intervention-protocol-compact} reports F1 and SHD; \cref{fig:intervention-protocol-profiles} plots the F1 change relative to Default.

\input{tables/generated_results/paper_compact/intervention_protocol_compact_f1_shd}

The F1-variation column further separates methods whose scores stay similar from those that move: TabCausal and SDCD vary least across these protocols (F1 SD $0.009$ and $0.010$), while IGSP is most sensitive (F1 SD $0.100$). \Cref{fig:intervention-protocol-profiles} localizes that drop. Fixed-setpoint stays near Default for every method; the bars that stand out are IGSP under multi-target, dense mixed, and high-intervention-fraction interventions ($\Delta$F1 $-0.27$, $-0.15$, and $-0.09$). This is consistent with IGSP's reliance on conditional-independence and invariance tests, which need sufficient per-intervention samples to be reliable; prior benchmark discussions have likewise noted that IGSP can underperform when only a small number of intervention samples are available per node. GIES, which uses the same interventional tables, rises under dense mixed and high-intervention-fraction ($\Delta$F1 $+0.02$ and $+0.03$) and drops mainly under parameter-shift. Complete protocol-wise metrics are in \cref{sec:appendix-complete-intervention}.

\subsection{Ablation Studies}

\begin{figure}[!htbp]
\centering
\includegraphics[width=0.78\textwidth]{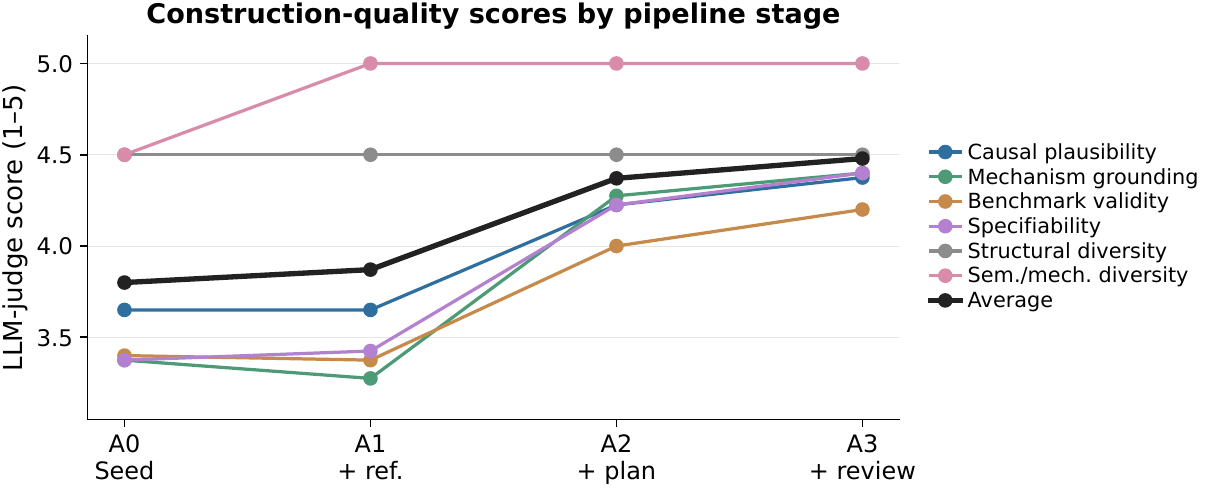}
\caption{\textbf{Construction-pipeline ablation.}
LLM-judge stage curves over 20 scenarios (10 Semantic, 10 Formula); the thicker line is the reported average.
Criterion-level means with standard deviations are in \cref{tab:construction-ablation-quality}.}
\label{fig:construction-ablation-stages}
\end{figure}

\paragraph{Construction pipeline.}
\Cref{fig:construction-ablation-stages} compares successive stages of the agentic construction pipeline using an LLM-judge quality rubric on 20 sampled scenarios, 10 from Semantic and 10 from Formula. The full construction process is described in \cref{sec:benchmark-data-details}. The reference pack adds external factual context; planning organizes variables and causal chains before graph drafting; graph review removes aliases, unsupported shortcuts, and low-information support nodes. In \cref{fig:construction-ablation-stages} the average rises from $3.80$ (seed only) to $3.87$ after the reference pack, $4.37$ after planning, and $4.48$ after graph review. The reference pack mainly lifts semantic/mechanism diversity; planning is the largest step, raising causal plausibility, mechanism grounding, benchmark validity, and specifiability together; graph review adds a smaller further gain on the same criteria. Structural diversity is already high at the seed stage and stays flat. The score pattern supports keeping these stages.

\subsection{Runtime and Resource}

Pretrained and amortized wrappers pay a one-time load before they score a graph. On a short evaluation shard that load can dwarf the inference itself. We therefore time nested runs that score $k$ graphs after one load, using a synthetic sample balanced across dimensions and graph families, and fit
\[
T(k)=T_{\mathrm{fixed}}+kT_{\mathrm{graph}},
\]
where $T_{\mathrm{fixed}}$ is the load and $T_{\mathrm{graph}}$ is the extra seconds for one more graph once the model is already in memory.
The nested $k$ grid is larger at low $d$, where each extra graph is cheap and a short run would make the load/inference split unstable, and smaller at high $d$, where each extra graph is expensive: $k\in\{100,200,500,1000\}$ at $d=10$, $k\in\{50,100,200,500\}$ at $d=20,30$, and $k\in\{20,50,100,200\}$ at $d=50,100$.

\input{tables/generated_results/paper_compact/runtime_pretrained_fit}

\begin{figure}[!htbp]
\centering
\includegraphics[width=0.92\textwidth]{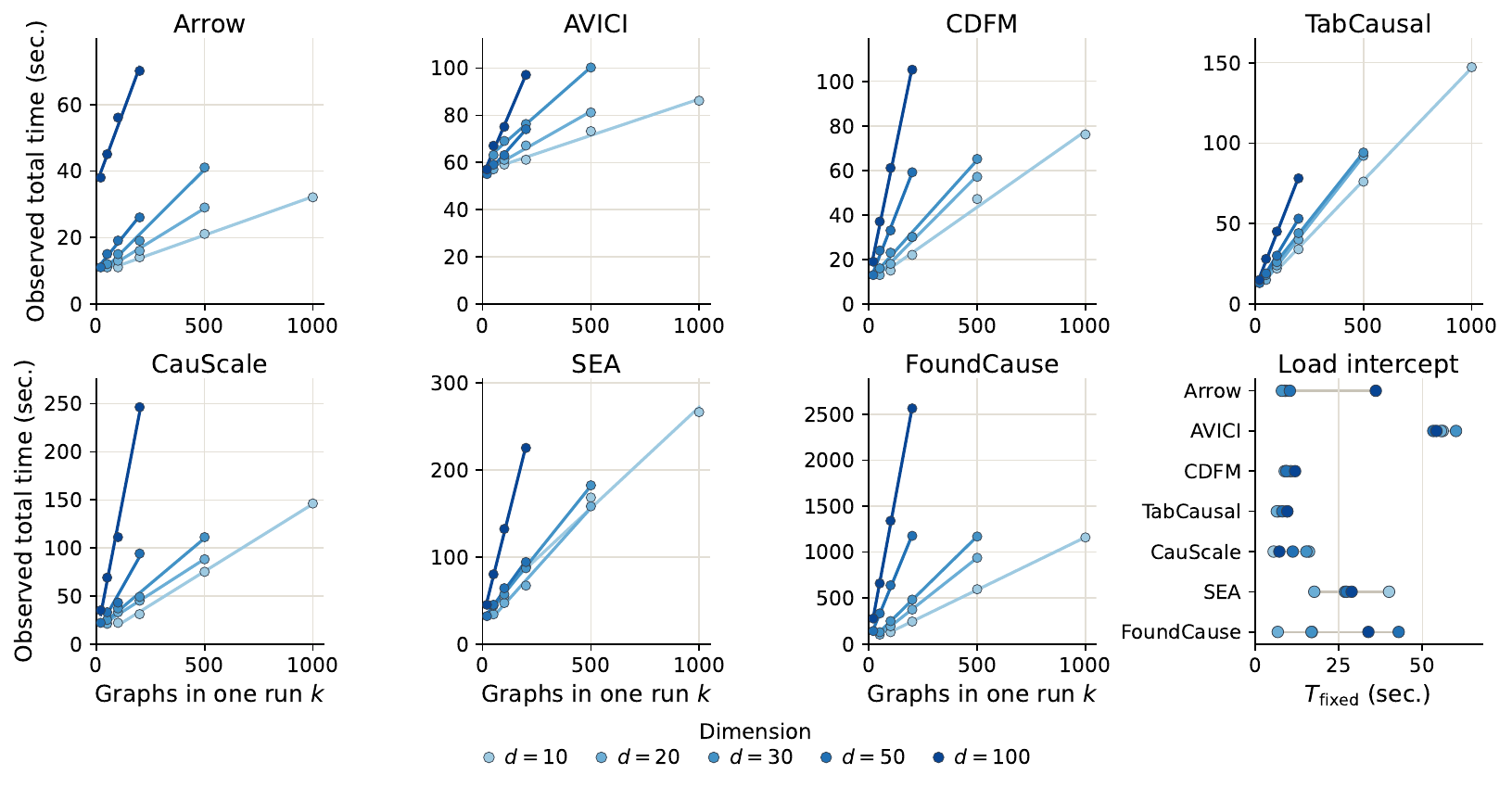}
\caption{\textbf{Pretrained runtime fit quality.} Each line panel is one pretrained/amortized wrapper on its own vertical scale. Points are observed total wrapper time on a synthetic sample balanced across dimensions and graph families; a line is the fitted $T(k)=T_{\mathrm{fixed}}+kT_{\mathrm{graph}}$ within dimension $d$. Dots on the line mean the split is stable; a steeper line, on that panel's scale, costs more per extra graph. The last panel shows the fitted load $T_{\mathrm{fixed}}$.}
\label{fig:pretrained-runtime-fit-quality}
\end{figure}

\Cref{tab:pretrained-runtime-fit,fig:pretrained-runtime-fit-quality} split that cost. All method--$d$ fits have $R^2>0.97$, and all but CauScale at $d=50$ exceed $0.98$. After load, Arrow and AVICI are the cheapest extra graphs, CDFM and TabCausal are next, CauScale and SEA are slower, and FoundCause sits in a higher band. TabCausal also has the shortest mean load ($8.13$s); AVICI has the longest ($55.84$s). The later comparison still drops $T_{\mathrm{fixed}}$. A pretrained wrapper is usually loaded once and then applied to many graphs---a leaderboard pool, a dimension or family shard, or a batch of related systems---so the one-time cost is \(T_{\mathrm{fixed}}/k\) per graph and becomes small once $k$ is large. The comparable quantity that remains is therefore $T_{\mathrm{graph}}$.

\Cref{tab:runtime-compact} reports one comparable per-graph number for every method. At each $d$, the nested-$k$ sample draws equally many graphs from each graph family, matching the synthetic-main mix, so the ranking is comparable; $T_{\mathrm{fixed}}$ is omitted because a large-$k$ run amortizes the load.

\input{tables/generated_results/paper_compact/runtime_compact}

\begin{figure}[!htbp]
\centering
\includegraphics[width=0.66\textwidth]{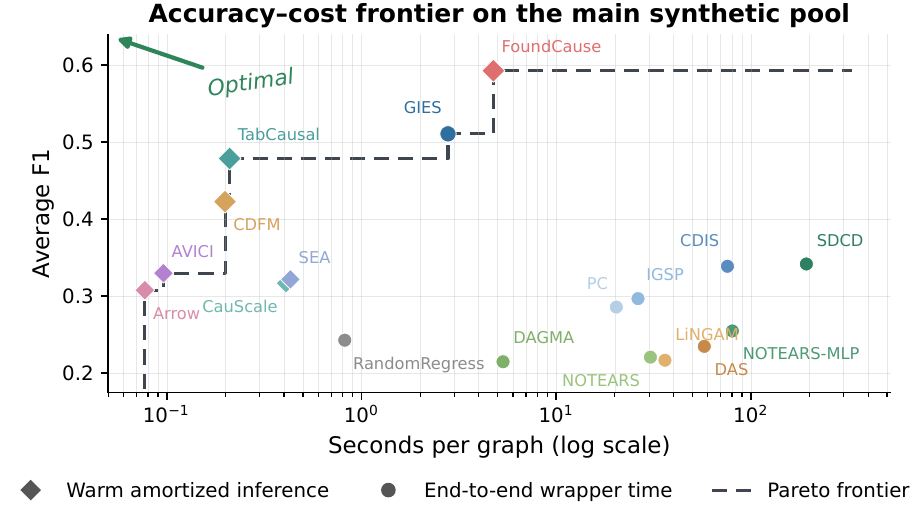}
\caption{\textbf{Accuracy--cost frontier.} Average F1 from \cref{tab:main-compact-f1-shd} against the comparable per-graph times in \cref{tab:runtime-compact}. Diamonds use pretrained $T_{\mathrm{graph}}$ from a sample balanced across synthetic dimensions and graph families, omitting $T_{\mathrm{fixed}}$; circles use mean end-to-end time on the synthetic main benchmark. Average F1 follows the table Avg.\ column, which averages only the family--split cells available for each method. The dashed line is the Pareto frontier; \emph{Optimal} points toward higher F1 and lower time.}
\label{fig:method-cost-frontier}
\end{figure}

\Cref{fig:method-cost-frontier} uses the same comparable per-graph times. For pretrained methods the $x$-coordinate is $T_{\mathrm{graph}}$ with $T_{\mathrm{fixed}}$ omitted, because a large-scale causal-discovery run amortizes the load. The useful frontier is not simply the set of fastest methods. Arrow and AVICI are very cheap but have lower average F1; CDFM and TabCausal move the frontier upward at nearly the same cost scale; GIES remains the strongest low-cost graphical-search baseline; and FoundCause reaches the highest average F1 among its supported observation-only cells at a larger $T_{\mathrm{graph}}$. Several classical or optimization methods are dominated in this view: they are slower than the frontier methods while also having lower average F1. Documented pretraining priors do not cover this benchmark equally, so part of a pretrained score could sit on documented support. A later observation-only check in \cref{sec:ood-score-analysis} splits categories by each method's own audited prior and finds only modest ID-to-OOD shrinks in relative lead for that reference group; the coarse order is stable in a coverage-mix reading, with the limits of that diagnostic stated there.

%% file: tables/generated_results/paper_compact/main_compact_f1_shd.tex
\begin{table}[!htbp]
\centering
\caption{\textbf{Main benchmark results (F1 / SHD).} Cells report mean\scorestd{standard deviation}. Columns labeled obs+int are observational-plus-interventional, not intervention-only. Higher F1 and lower SHD are better; bold and underline mark the best and second-best F1 and SHD separately in each column. Avg.\ is the unweighted mean over the available family--split cells for that method, so methods without intervention support are averaged over three obs-only cells, not six. SHD is not commensurate across families of different graph size.}
\label{tab:main-compact-f1-shd}
\scriptsize
\setlength{\tabcolsep}{2.4pt}
\renewcommand{\arraystretch}{1.12}
\resizebox{\textwidth}{!}{%
\begin{tabular}{@{}llccccccc@{}}
\toprule
\textbf{Type} & \textbf{Method} &
\textbf{Syn. obs} & \textbf{Syn. obs+int} & \textbf{Sem. obs} & \textbf{Sem. obs+int} & \textbf{Form. obs} & \textbf{Form. obs+int} & \textbf{Avg.} \\
\midrule
Control & RandomRegress & 0.28\scorestd{0.07} / 218.0\scorestd{308.7} & -- & 0.25\scorestd{0.07} / 93.2\scorestd{69.0} & -- & 0.20\scorestd{0.06} / 108.3\scorestd{69.9} & -- & 0.24\scorestd{0.07} / 139.8\scorestd{149.2} \\
\addlinespace[0.4em]
\multirow{4}{*}{\shortstack[l]{Graphical\\search}} & CDIS & 0.38\scorestd{0.14} / 143.0\scorestd{351.2} & 0.40\scorestd{0.13} / 140.1\scorestd{352.9} & 0.40\scorestd{0.17} / \underline{31.7\scorestd{9.0}} & 0.40\scorestd{0.16} / \textbf{31.9\scorestd{8.9}} & 0.23\scorestd{0.18} / \underline{31.0\scorestd{6.8}} & 0.23\scorestd{0.18} / 30.9\scorestd{6.9} & 0.34\scorestd{0.16} / 68.1\scorestd{122.6} \\
 & GIES & 0.48\scorestd{0.13} / 123.6\scorestd{174.1} & \underline{0.48\scorestd{0.11}} / 134.0\scorestd{180.7} & \underline{0.55\scorestd{0.13}} / 33.4\scorestd{15.2} & \textbf{0.57\scorestd{0.13}} / \underline{32.9\scorestd{15.4}} & 0.47\scorestd{0.12} / 36.7\scorestd{14.1} & \underline{0.52\scorestd{0.13}} / 34.7\scorestd{14.0} & \underline{0.51\scorestd{0.13}} / 65.9\scorestd{68.9} \\
 & IGSP & 0.41\scorestd{0.12} / 135.6\scorestd{191.9} & 0.39\scorestd{0.12} / 132.8\scorestd{185.7} & 0.31\scorestd{0.16} / 40.8\scorestd{13.8} & 0.31\scorestd{0.16} / 39.9\scorestd{12.9} & 0.21\scorestd{0.15} / 43.3\scorestd{16.9} & 0.16\scorestd{0.16} / 40.7\scorestd{15.9} & 0.30\scorestd{0.15} / 72.2\scorestd{72.8} \\
 & PC & 0.31\scorestd{0.12} / 115.4\scorestd{148.9} & -- & 0.33\scorestd{0.15} / 33.5\scorestd{9.4} & -- & 0.22\scorestd{0.16} / 31.1\scorestd{6.8} & -- & 0.29\scorestd{0.14} / 60.0\scorestd{55.0} \\
\addlinespace[0.4em]
\multirow{2}{*}{\shortstack[l]{Functional\\assumptions}} & DAS & 0.34\scorestd{0.11} / 142.6\scorestd{179.8} & -- & 0.21\scorestd{0.06} / 59.5\scorestd{15.4} & -- & 0.16\scorestd{0.06} / 64.8\scorestd{15.5} & -- & 0.23\scorestd{0.08} / 89.0\scorestd{70.2} \\
 & LiNGAM & 0.21\scorestd{0.12} / 109.6\scorestd{142.2} & -- & 0.20\scorestd{0.08} / 38.9\scorestd{9.3} & -- & 0.24\scorestd{0.08} / 34.7\scorestd{8.0} & -- & 0.22\scorestd{0.10} / 61.1\scorestd{53.2} \\
\addlinespace[0.4em]
\multirow{4}{*}{\shortstack[l]{Continuous\\optimization}} & DAGMA & 0.28\scorestd{0.12} / 107.2\scorestd{138.6} & -- & 0.18\scorestd{0.07} / 42.0\scorestd{9.1} & -- & 0.18\scorestd{0.07} / 39.7\scorestd{8.6} & -- & 0.21\scorestd{0.09} / 63.0\scorestd{52.1} \\
 & NOTEARS & 0.28\scorestd{0.11} / 104.2\scorestd{136.0} & -- & 0.20\scorestd{0.07} / 38.1\scorestd{8.5} & -- & 0.18\scorestd{0.08} / 35.2\scorestd{7.3} & -- & 0.22\scorestd{0.09} / 59.2\scorestd{50.6} \\
 & NOTEARS-MLP & 0.35\scorestd{0.10} / 107.0\scorestd{136.3} & -- & 0.22\scorestd{0.08} / 44.9\scorestd{10.7} & -- & 0.19\scorestd{0.06} / 44.1\scorestd{8.9} & -- & 0.26\scorestd{0.08} / 65.3\scorestd{52.0} \\
 & SDCD & 0.38\scorestd{0.09} / 135.5\scorestd{165.2} & 0.38\scorestd{0.09} / 148.1\scorestd{174.1} & 0.36\scorestd{0.09} / 51.2\scorestd{16.0} & 0.36\scorestd{0.09} / 51.7\scorestd{15.8} & 0.28\scorestd{0.08} / 61.5\scorestd{16.0} & 0.29\scorestd{0.09} / 61.6\scorestd{16.3} & 0.34\scorestd{0.09} / 84.9\scorestd{67.2} \\
\addlinespace[0.4em]
\multirow{7}{*}{Pretrained} & Arrow & 0.34\scorestd{0.13} / 106.0\scorestd{140.6} & -- & 0.31\scorestd{0.09} / 36.6\scorestd{9.5} & -- & 0.27\scorestd{0.08} / 35.9\scorestd{8.6} & -- & 0.31\scorestd{0.10} / 59.5\scorestd{52.9} \\
 & AVICI & 0.32\scorestd{0.15} / 104.5\scorestd{139.8} & 0.35\scorestd{0.16} / 104.2\scorestd{139.8} & 0.28\scorestd{0.13} / 35.5\scorestd{9.3} & 0.30\scorestd{0.13} / 34.8\scorestd{9.3} & 0.33\scorestd{0.12} / 31.8\scorestd{8.2} & 0.41\scorestd{0.12} / \underline{29.3\scorestd{8.0}} & 0.33\scorestd{0.14} / 56.7\scorestd{52.4} \\
 & CauScale & 0.41\scorestd{0.11} / \underline{85.8\scorestd{107.1}} & 0.46\scorestd{0.10} / \textbf{84.6\scorestd{106.6}} & 0.25\scorestd{0.07} / 50.2\scorestd{19.2} & 0.29\scorestd{0.08} / 48.6\scorestd{18.8} & 0.21\scorestd{0.07} / 65.9\scorestd{23.4} & 0.28\scorestd{0.09} / 59.6\scorestd{23.1} & 0.32\scorestd{0.09} / 65.8\scorestd{49.7} \\
 & CDFM & 0.42\scorestd{0.14} / 136.4\scorestd{176.5} & -- & 0.43\scorestd{0.07} / 46.0\scorestd{12.4} & -- & 0.41\scorestd{0.08} / 45.8\scorestd{13.9} & -- & 0.42\scorestd{0.10} / 76.1\scorestd{67.6} \\
 & FoundCause & \textbf{0.62\scorestd{0.12}} / \textbf{80.8\scorestd{111.8}} & -- & \textbf{0.63\scorestd{0.11}} / \textbf{26.8\scorestd{10.5}} & -- & \textbf{0.52\scorestd{0.12}} / 33.4\scorestd{12.3} & -- & \textbf{0.59\scorestd{0.12}} / \textbf{47.0\scorestd{44.9}} \\
 & SEA & 0.40\scorestd{0.13} / 99.2\scorestd{129.1} & -- & 0.33\scorestd{0.11} / 41.8\scorestd{17.9} & -- & 0.24\scorestd{0.09} / 52.8\scorestd{19.7} & -- & 0.32\scorestd{0.11} / 64.6\scorestd{55.6} \\
 & TabCausal & \underline{0.50\scorestd{0.16}} / 93.6\scorestd{127.3} & \textbf{0.50\scorestd{0.15}} / \underline{96.0\scorestd{128.4}} & 0.42\scorestd{0.13} / 36.1\scorestd{10.1} & \underline{0.45\scorestd{0.13}} / 35.0\scorestd{10.2} & \underline{0.47\scorestd{0.13}} / \textbf{29.5\scorestd{9.5}} & \textbf{0.53\scorestd{0.14}} / \textbf{26.6\scorestd{9.2}} & 0.48\scorestd{0.14} / \underline{52.8\scorestd{49.1}} \\
\bottomrule
\end{tabular}%
}
\end{table}

%% file: tables/generated_results/paper_compact/real_data_compact_side.tex
\scriptsize
\setlength{\tabcolsep}{2.0pt}
\renewcommand{\arraystretch}{0.92}
\begin{tabular}{@{}llcc@{}}
\toprule
\textbf{Type} & \textbf{Method} & \textbf{Obs-only} & \textbf{Obs+int} \\
\midrule
Control & RandomRegress & 0.20\scorestd{0.14} / 59.9 & -- \\
\addlinespace[0.15em]
\multirow{4}{*}{Graphical} & CDIS & 0.28\scorestd{0.22} / 99.5 & 0.14\scorestd{0.17} / 45.4 \\
 & GIES & 0.26\scorestd{0.15} / 48.0 & \underline{0.29\scorestd{0.13}} / 118.7 \\
 & IGSP & 0.18\scorestd{0.22} / 31.1 & 0.11\scorestd{0.12} / 59.1 \\
 & PC & 0.18\scorestd{0.14} / 28.9 & -- \\
\addlinespace[0.15em]
\multirow{2}{*}{Functional} & DAS & 0.19\scorestd{0.23} / 61.9 & -- \\
 & LiNGAM & 0.13\scorestd{0.17} / \textbf{24.3} & -- \\
\addlinespace[0.15em]
\multirow{4}{*}{Optim.} & DAGMA & 0.16\scorestd{0.14} / 31.8 & -- \\
 & NOTEARS & 0.16\scorestd{0.15} / 29.1 & -- \\
 & NOTEARS-MLP & 0.21\scorestd{0.19} / 55.0 & -- \\
 & SDCD & 0.22\scorestd{0.17} / 44.5 & 0.27\scorestd{0.12} / 103.4 \\
\addlinespace[0.15em]
\multirow{7}{*}{Pretrained} & Arrow & 0.15\scorestd{0.13} / 70.5 & -- \\
 & AVICI & 0.22\scorestd{0.18} / \underline{26.0} & 0.28\scorestd{0.12} / \underline{42.6} \\
 & CauScale & 0.08\scorestd{0.08} / 36.2 & 0.19\scorestd{0.07} / 62.1 \\
 & CDFM & \textbf{0.39\scorestd{0.14}} / 38.5 & -- \\
 & FoundCause & 0.21\scorestd{0.23} / 98.3 & -- \\
 & SEA & 0.21\scorestd{0.14} / 31.3 & -- \\
 & TabCausal & \underline{0.34\scorestd{0.24}} / 28.4 & \textbf{0.46\scorestd{0.11}} / \textbf{38.4} \\
\bottomrule
\end{tabular}

%% file: tables/generated_results/paper_compact/intervention_protocol_compact_f1_shd.tex
\begin{table}[!htbp]
\centering
\scriptsize
\setlength{\tabcolsep}{2.2pt}
\renewcommand{\arraystretch}{1.18}
\begin{threeparttable}
\newcommand{\protocolcell}[3]{\begin{tabular}{@{}c@{}}#1 / #3\\[-0.2em]\textcolor{black!55}{\tiny $\Delta$F1 #2}\end{tabular}}
\caption{\textbf{Intervention-protocol sensitivity.} Cells report F1 / SHD; non-default protocols put the gray F1 change relative to Default on the second line. The last column is the standard deviation of each method's F1 across intervention protocols (lower is better). Bold and underline mark the best and second-best F1, SHD, and F1 SD separately.}
\label{tab:intervention-protocol-compact}
\begin{tabularx}{\linewidth}{@{}l*{7}{>{\centering\arraybackslash}X}@{}}
\toprule
\textbf{Method} & \textbf{Default} & \textbf{Fixed} & \textbf{Shift} & \textbf{Multi} & \textbf{Dense} & \textbf{High-frac} & \textbf{F1 SD} \\
\midrule
CDIS & 0.397 / 48.1 & \protocolcell{0.396}{-0.001}{48.3} & \protocolcell{0.397}{+0.000}{48.3} & \protocolcell{0.382}{-0.015}{48.9} & \protocolcell{0.365}{-0.032}{47.2} & \protocolcell{0.355}{-0.041}{47.6} & 0.017 \\
GIES & \underline{0.480} / 59.7 & \protocolcell{\underline{0.480}}{+0.000}{59.7} & \protocolcell{\underline{0.429}}{-0.051}{64.1} & \protocolcell{\underline{0.483}}{+0.004}{59.3} & \protocolcell{\underline{0.497}}{+0.017}{56.3} & \protocolcell{\underline{0.506}}{+0.026}{56.7} & 0.024 \\
IGSP & 0.400 / 53.7 & \protocolcell{0.398}{-0.001}{54.2} & \protocolcell{0.394}{-0.006}{54.4} & \protocolcell{0.127}{-0.273}{54.6} & \protocolcell{0.248}{-0.152}{51.8} & \protocolcell{0.312}{-0.088}{52.0} & 0.100 \\
SDCD & 0.380 / 73.0 & \protocolcell{0.381}{+0.002}{72.7} & \protocolcell{0.375}{-0.004}{73.2} & \protocolcell{0.385}{+0.005}{72.4} & \protocolcell{0.404}{+0.024}{68.3} & \protocolcell{0.398}{+0.018}{70.3} & \underline{0.010} \\
AVICI & 0.390 / \textbf{43.4} & \protocolcell{0.385}{-0.005}{\underline{43.7}} & \protocolcell{0.365}{-0.024}{\textbf{44.6}} & \protocolcell{0.385}{-0.005}{\textbf{43.7}} & \protocolcell{0.371}{-0.019}{\textbf{44.3}} & \protocolcell{0.404}{+0.014}{\underline{42.9}} & 0.013 \\
CauScale & 0.451 / \underline{43.6} & \protocolcell{0.458}{+0.006}{\textbf{43.0}} & \protocolcell{0.418}{-0.034}{\underline{45.2}} & \protocolcell{0.438}{-0.014}{45.7} & \protocolcell{0.424}{-0.027}{46.7} & \protocolcell{0.484}{+0.032}{\textbf{40.8}} & 0.022 \\
TabCausal & \textbf{0.539} / 45.6 & \protocolcell{\textbf{0.538}}{-0.001}{45.8} & \protocolcell{\textbf{0.517}}{-0.022}{47.6} & \protocolcell{\textbf{0.538}}{-0.001}{\underline{45.7}} & \protocolcell{\textbf{0.526}}{-0.013}{\underline{46.1}} & \protocolcell{\textbf{0.542}}{+0.003}{44.8} & \textbf{0.009} \\
\bottomrule
\end{tabularx}
\end{threeparttable}
\end{table}

%% file: tables/generated_results/paper_compact/runtime_pretrained_fit.tex
\begin{table}[!htbp]
\centering
\scriptsize
\setlength{\tabcolsep}{3.2pt}
\renewcommand{\arraystretch}{1.10}
\begin{threeparttable}
\caption{\textbf{Load and per-graph time for pretrained/amortized methods.} Fits use a synthetic sample balanced across dimensions $d\in\{10,20,30,50,100\}$ and graph families. $T_{\mathrm{fixed}}$ is the fitted one-time load, averaged over $d$. $T_{\mathrm{graph}}$ is the extra seconds for one more graph after the model is loaded, reported by $d$ and as an unweighted mean over those $d$. Nested $k$ is $\{100,200,500,1000\}$ at $d{=}10$, $\{50,100,200,500\}$ at $d{=}20,30$, and $\{20,50,100,200\}$ at $d{=}50,100$: larger $k$ at low $d$ for a stable split, smaller $k$ at high $d$ because extra graphs cost more. Lower time is better. Bold is lowest in a time column and underline is second; $R^2$ columns are not ranked. All method--$d$ fits have $R^2>0.97$, and all but one exceed $0.98$.}
\label{tab:pretrained-runtime-fit}
\begin{tabular*}{\textwidth}{@{\extracolsep{\fill}}l r r r r r r r r r}
\toprule
 & & \multicolumn{6}{c}{\textbf{Per-graph seconds $T_{\mathrm{graph}}$}} & & \\
\cmidrule(lr){3-8}
\textbf{Method} & \textbf{$T_{\mathrm{fixed}}$} & \textbf{$d{=}10$} & \textbf{$d{=}20$} & \textbf{$d{=}30$} & \textbf{$d{=}50$} & \textbf{$d{=}100$} & \textbf{Mean} & \textbf{Mean $R^2$} & \textbf{Min $R^2$} \\
\midrule
Arrow & 14.47 & \textbf{0.023} & \textbf{0.040} & \textbf{0.065} & \textbf{0.081} & \textbf{0.176} & \textbf{0.077} & 0.9905 & 0.9837 \\
AVICI & 55.84 & \underline{0.031} & \underline{0.052} & \underline{0.081} & \underline{0.104} & \underline{0.215} & \underline{0.096} & 0.9933 & 0.9897 \\
CDFM & \underline{10.02} & 0.068 & 0.098 & 0.108 & 0.249 & 0.472 & 0.199 & 0.9947 & 0.9922 \\
TabCausal & \textbf{8.13} & 0.140 & 0.171 & 0.170 & 0.224 & 0.346 & 0.210 & 0.9993 & 0.9979 \\
CauScale & 11.07 & 0.140 & 0.145 & 0.190 & 0.398 & 1.170 & 0.409 & 0.9902 & 0.9718 \\
SEA & 28.14 & 0.232 & 0.278 & 0.310 & 0.341 & 0.993 & 0.431 & 0.9951 & 0.9898 \\
FoundCause & 23.46 & 1.146 & 1.861 & 2.310 & 5.729 & 12.725 & 4.754 & 0.9995 & 0.9984 \\
\bottomrule
\end{tabular*}
\end{threeparttable}
\end{table}

%% file: tables/generated_results/paper_compact/runtime_compact.tex
\begin{table}[!htbp]
\centering
\scriptsize
\setlength{\tabcolsep}{3.2pt}
\renewcommand{\arraystretch}{1.08}
\begin{threeparttable}
\caption{\textbf{Per-graph runtime on the synthetic mix.} Non-pretrained entries are mean end-to-end wrapper time on the synthetic main benchmark, mean\scorestd{standard deviation}. $^{\dagger}$ marks the unweighted mean of per-$d$ $T_{\mathrm{graph}}$ from nested-$k$ fits. At each $d$, the sample draws equally many graphs from each graph family, matching the synthetic-main mix, so the times are comparable; $T_{\mathrm{fixed}}$ is omitted because it amortizes at large $k$. Lower is better.}
\label{tab:runtime-compact}
\begin{tabular*}{\textwidth}{@{\extracolsep{\fill}}r l c r l c r l c@{}}
\toprule
\textbf{Rank} & \textbf{Method} & \textbf{Sec./graph} & \textbf{Rank} & \textbf{Method} & \textbf{Sec./graph} & \textbf{Rank} & \textbf{Method} & \textbf{Sec./graph} \\
\midrule
1 & Arrow & $0.077^{\dagger}$ & 7 & RandomRegress & $0.82\scorestd{0.53}$ & 13 & NOTEARS & $30.46\scorestd{57.50}$ \\
2 & AVICI & $0.096^{\dagger}$ & 8 & GIES & $2.78\scorestd{1.19}$ & 14 & LiNGAM & $36.17\scorestd{59.43}$ \\
3 & CDFM & $0.199^{\dagger}$ & 9 & FoundCause & $4.754^{\dagger}$ & 15 & DAS & $57.60\scorestd{62.02}$ \\
4 & TabCausal & $0.210^{\dagger}$ & 10 & DAGMA & $5.33\scorestd{3.01}$ & 16 & CDIS & $75.75\scorestd{253.70}$ \\
5 & CauScale & $0.409^{\dagger}$ & 11 & PC & $20.36\scorestd{146.76}$ & 17 & NOTEARS-MLP & $80.27\scorestd{99.35}$ \\
6 & SEA & $0.431^{\dagger}$ & 12 & IGSP & $26.30\scorestd{131.27}$ & 18 & SDCD & $192.64\scorestd{115.28}$ \\
\bottomrule
\end{tabular*}
\end{threeparttable}
\end{table}

%% file: analysis.tex
\section{Analysis}
\label{sec:analysis}

A released suite still needs to be checked for breadth and for whether scores can be traced to specific graph, mechanism, or domain properties. Following the diagnosability requirement in \cref{sec:benchmark_requirements}, this section reports synthetic factor analyses, a check of pretrained methods on categories outside their documented training support, and Semantic/Formula domain breakdowns.

\subsection{Synthetic Benchmark Behavior Analysis}
\label{sec:synthetic-behavior-analysis}

\paragraph{Interaction score.}
Aggregate leaderboards rank methods, but they do not say why the ranks differ. We analyze the synthetic obs-only benchmark with factor-specific interaction scores. For a method $m$ and factor category $c$, let $F_{m,c}$ be the mean F1 on that category. We report
\begin{equation}
    I_{m,c}=F_{m,c}-\bar F_{m,\cdot}-\bar F_{\cdot,c}+\bar F_{\cdot,\cdot}.
\end{equation}
This double-centering removes each method's overall strength and each category's overall difficulty. A positive value means that a method is stronger on that category than expected from both marginal averages. For graph-level factors, categories are assigned per graph. For edge-level factors, mechanisms and noises are assigned to an edge by the child node's structural equation.

\paragraph{Graph structure.}
\begin{figure}[!htbp]
    \centering
    \includegraphics[width=0.88\textwidth]{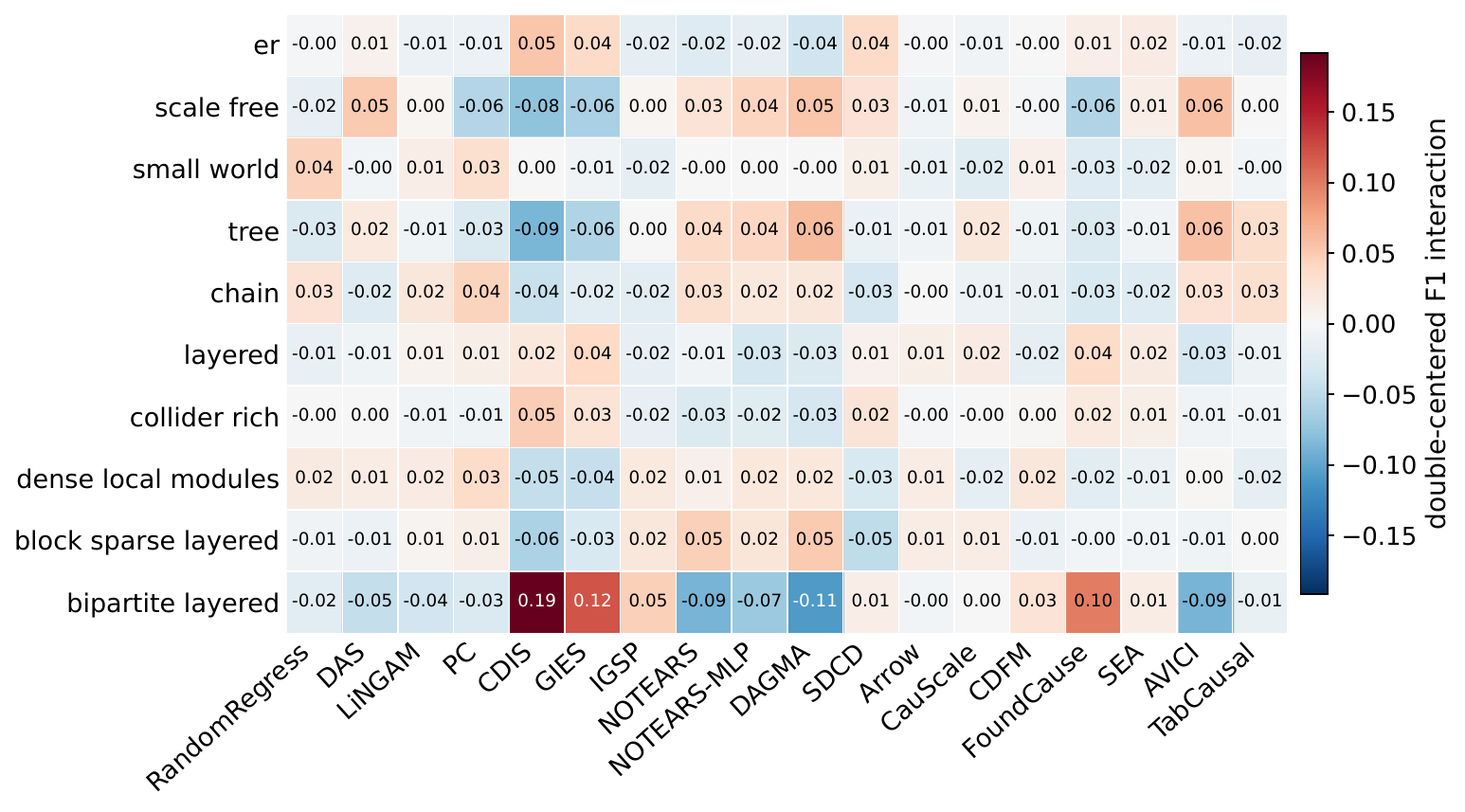}
    \caption{\textbf{Synthetic graph-family factor interactions.} Each cell is a double-centered F1 interaction on the observation-only synthetic benchmark.}
    \label{fig:synthetic-graph-factor-interactions}
\end{figure}

\Cref{fig:synthetic-graph-factor-interactions} shows that methods differ by graph topology even after category difficulty is removed. The clearest contrast is the bipartite-layered family: CDIS ($+0.192$), GIES ($+0.120$), and FoundCause ($+0.098$) are relatively stronger, while DAGMA ($-0.108$), NOTEARS ($-0.089$), and AVICI ($-0.089$) are relatively weaker. For CDIS/GIES, this is consistent with graphical-search behavior: bipartite and collider-rich structures create many separation and v-structure signals that conditional-independence or equivalence-class search can exploit. FoundCause is pretrained, but its pairwise-statistics pathway and triangular refinement module are also designed to reason over higher-order motifs such as chains and colliders, which plausibly explains why it benefits from the same structured signals. By contrast, continuous-optimization methods and AVICI are more favorable on sparse branching families such as trees or scale-free graphs, where recovering a weighted acyclic parent structure is closer to their optimization or amortized parent-selection bias.

\paragraph{Scale effects.}
\begin{figure}[!htbp]
    \centering
    \includegraphics[width=0.88\textwidth]{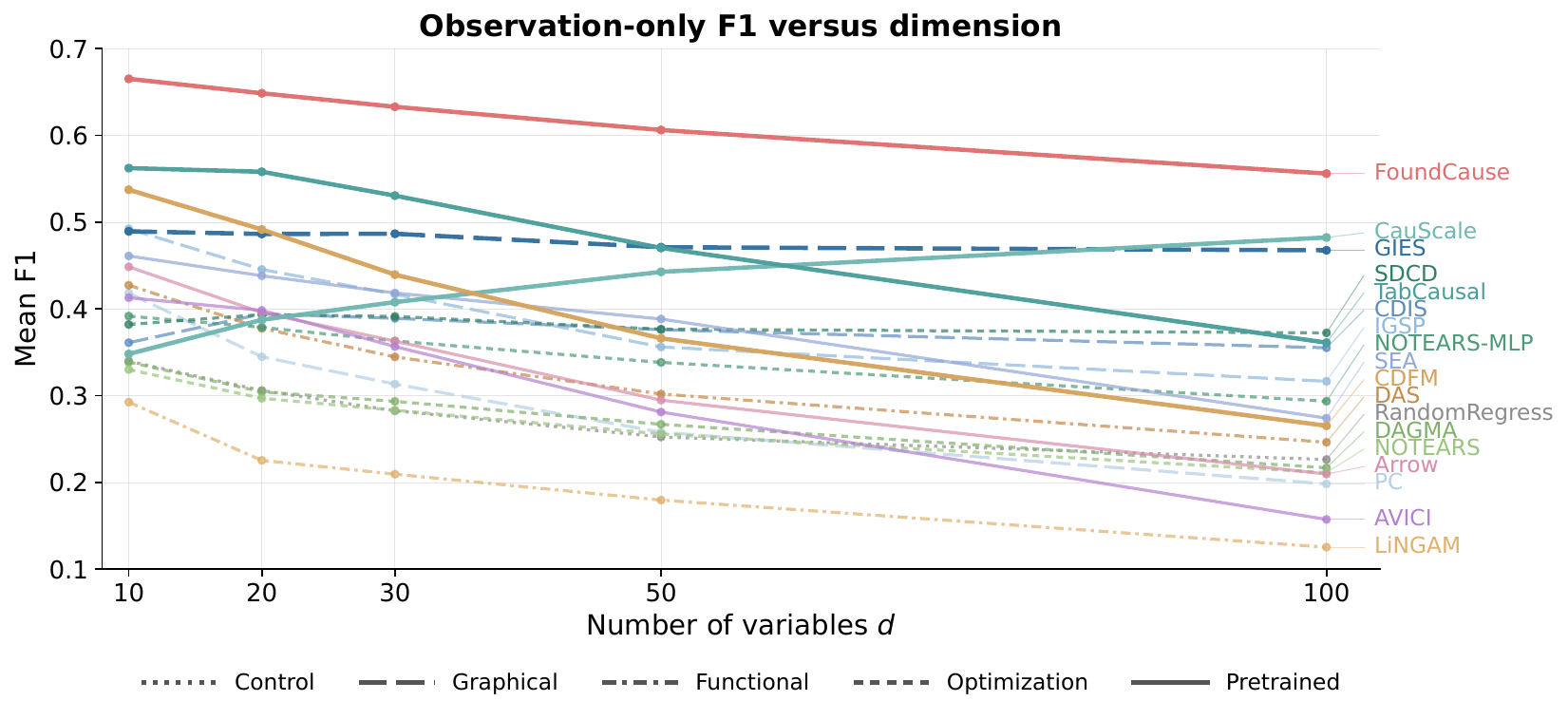}
    \caption{\textbf{Synthetic dimension scaling.} Mean observation-only F1 versus the number of variables $d$. Line styles follow method families, as in the sample-size plot. CauScale is the only method whose F1 rises with $d$.}
    \label{fig:synthetic-dimension-scaling}
\end{figure}

Dimension is ordered, so \cref{fig:synthetic-dimension-scaling} reports it as curves. Almost every method loses F1 as $d$ grows, but the slopes differ. FoundCause stays highest throughout ($0.67$ at $d=10$, $0.56$ at $d=100$) while declining. CauScale is the exception that rises ($0.35$ to $0.48$), matching its design goal and training setup for large graphs: its synthetic training includes graph sizes up to hundreds of nodes, and its reduction/tied-attention design is optimized for scaling. GIES, CDIS, and SDCD are comparatively flat. Arrow, CDFM, and AVICI drop sharply by $d=100$ (to $0.21$, $0.27$, and $0.16$); this likely reflects dense pairwise decoding, skeleton/order or edge-head uncertainty accumulating over many variable pairs, and training distributions whose strongest support is not centered on our high-dimensional heterogeneous synthetic pool. TabCausal also falls ($0.56$ to $0.36$) but remains above those collapsing pretrained methods. Both TabCausal and FoundCause combine broad synthetic task variation with explicit pairwise or motif-level signals, which helps preserve relative performance as the variable set grows.

\begin{figure}[!htbp]
    \centering
    \begin{subfigure}[t]{0.88\textwidth}
        \centering
        \includegraphics[width=\textwidth]{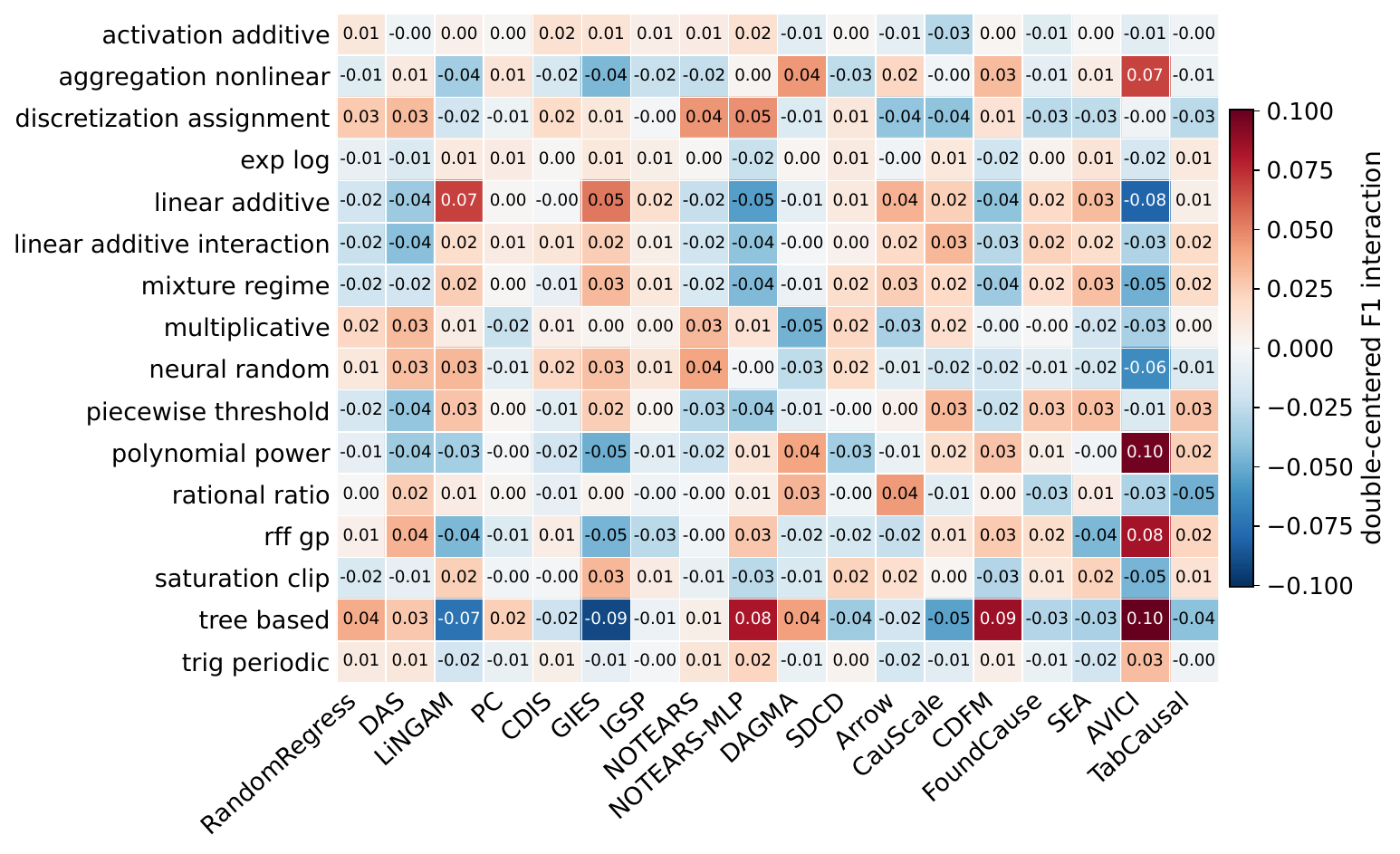}
        \caption{Mechanism interactions (child-node mechanism family).}
        \label{fig:synthetic-mechanism-interactions}
    \end{subfigure}\\[0.6em]
    \begin{subfigure}[t]{0.88\textwidth}
        \centering
        \includegraphics[width=\textwidth]{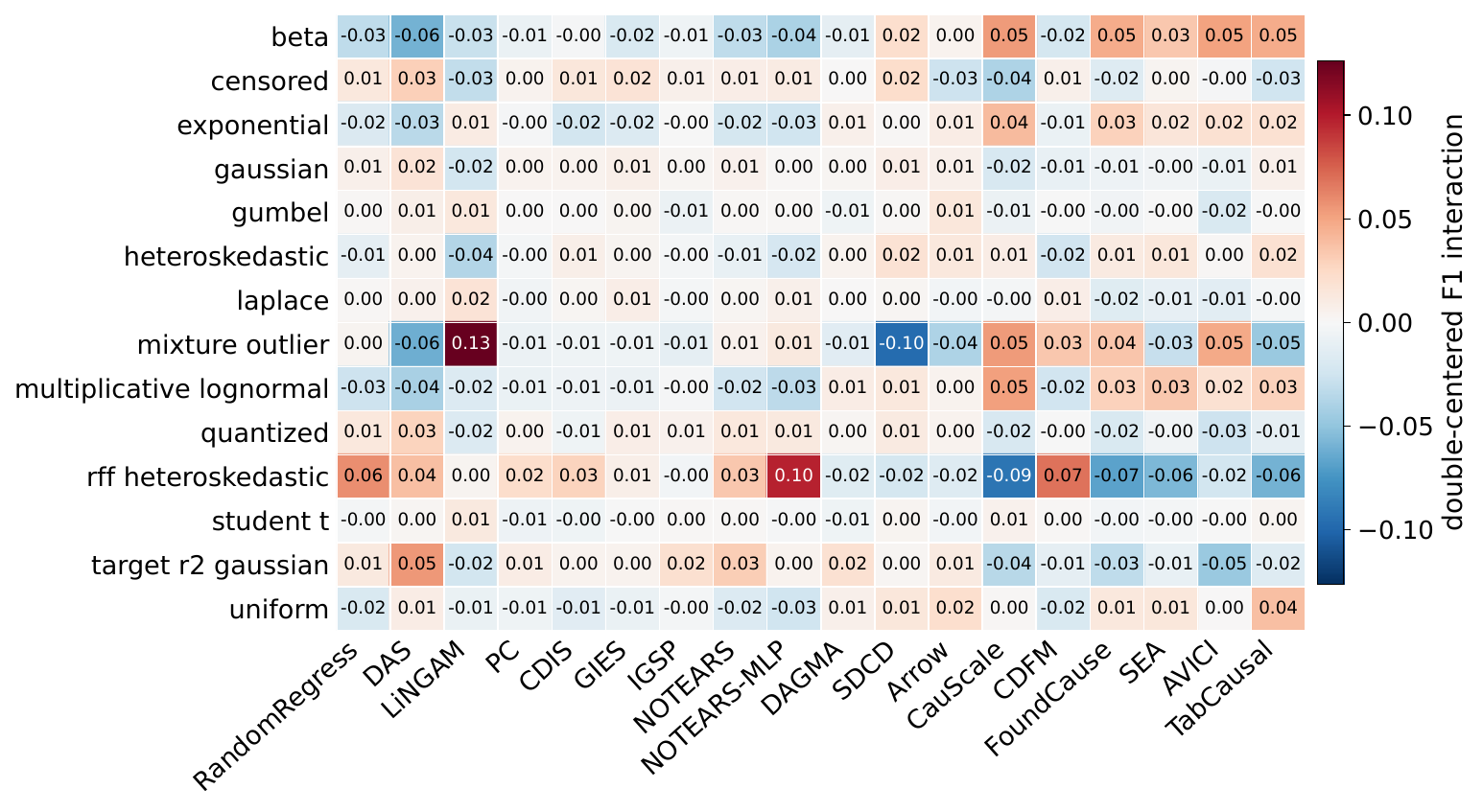}
        \caption{Noise interactions (child-node noise family).}
        \label{fig:synthetic-noise-interactions}
    \end{subfigure}
    \caption{\textbf{Synthetic mechanism and noise interactions.} Each cell is a double-centered F1 interaction on the observation-only synthetic benchmark.}
    \label{fig:synthetic-mechanism-noise-interactions}
\end{figure}

\paragraph{Mechanism and noise effects.}
\Cref{fig:synthetic-mechanism-interactions} shows a clear split between linear-identifiability assumptions and nonlinear mechanism coverage. On the linear-additive row, LiNGAM is the strongest positive method ($+0.069$), which is expected because this setting matches its linear non-Gaussian identification principle. NOTEARS is nominally a linear-SEM method, but its squared-loss objective is closer to a linear-Gaussian score and does not exploit non-Gaussian residual information for orientation; since most linear-additive cases in this pool are not Gaussian-noise linear SEMs, NOTEARS is not helped in the same way ($-0.023$). GIES follows LiNGAM closely ($+0.053$ on linear-additive, and a row-wise mechanism correlation of about $0.95$ with LiNGAM), even though the algorithms are different. The common pattern is that both are most comfortable when causal effects are globally simple and separable, and both lose relative advantage when the edge signal is local, nonlinear, or non-smooth.

The tree-based row creates the strongest mechanism split. AVICI ($+0.100$), CDFM ($+0.086$), and NOTEARS-MLP ($+0.082$) are strongly positive, while GIES ($-0.091$), LiNGAM ($-0.074$), CauScale ($-0.054$), and TabCausal ($-0.041$) are negative after double-centering. This row corresponds to piecewise and threshold-like parent effects: the relevant parent set can be recoverable when a model has enough nonlinear function capacity or has seen similar nonlinear signatures during pretraining, but the same structure is hard for linear/non-Gaussian assumptions or score-based search to express as a stable global dependence pattern. The negative values for some pretrained models mark relative specialization: their overall accuracy may remain strong, but their advantage is less concentrated on this particular piecewise mechanism class. AVICI is especially mechanism-sensitive in this panel: its interaction variance across mechanism rows is the largest among the pretrained methods, with large positive values on tree-based, polynomial-power ($+0.097$), and RFF-GP ($+0.083$) rows but a large negative value on linear-additive ($-0.081$). We therefore read AVICI as less stable across mechanism families than broader or more motif-oriented pretrained methods such as TabCausal and FoundCause, without attributing this pattern to a single mechanism property.

The noise panel separates whether scores move more with the noise family or with the mechanism family. Overall, noise families create weaker method separation than child mechanisms: the average row spread is smaller for noise than for mechanisms, although mixture-outlier and RFF-heteroskedastic noise are strong exceptions. Graphical-search methods are comparatively flat across noise rows, whereas pretrained methods show more visible noise-dependent swings than graphical search and most continuous-optimization methods. This sensitivity is less distinctive in the graph-family and mechanism panels, where variation is shared across several method classes. Noise families therefore show how each method uses residual shape, scale variation, and support constraints.

The largest contrasts come from noises that change more than marginal variance. Mixture-outlier noise injects sparse contamination, so methods that rely on smooth losses, ordering heuristics, or stable low-order associations can be pulled toward spurious supports; this is consistent with negative interactions for SDCD ($-0.097$) and DAS ($-0.061$). LiNGAM is the main exception ($+0.126$): in this slice, outliers amplify non-Gaussian residual cues, which its identification principle can sometimes exploit. RFF-heteroskedastic noise produces a different split because the noise scale changes with parent values, so the edge evidence is a conditional-variance signal. NOTEARS-MLP is strongly positive ($+0.097$), plausibly because its nonlinear regression model can absorb associated nonlinear parent-dependent patterns, while CDFM is also positive ($+0.069$), consistent with explicit heteroskedastic coverage in its pretraining prior. Several pretrained methods are negative on this row: even a broad pretraining prior can remain sensitive to variance-driven evidence. RandomRegress also has visible tendencies across noise rows, for example on RFF-heteroskedastic noise ($+0.059$); because it is a fixed-random-order regression control, these tendencies reflect sensitivity of nonzero regression support to scale/noise artifacts.

\paragraph{Root dependency effects.}
\begin{figure}[!htbp]
    \centering
    \includegraphics[width=0.72\textwidth]{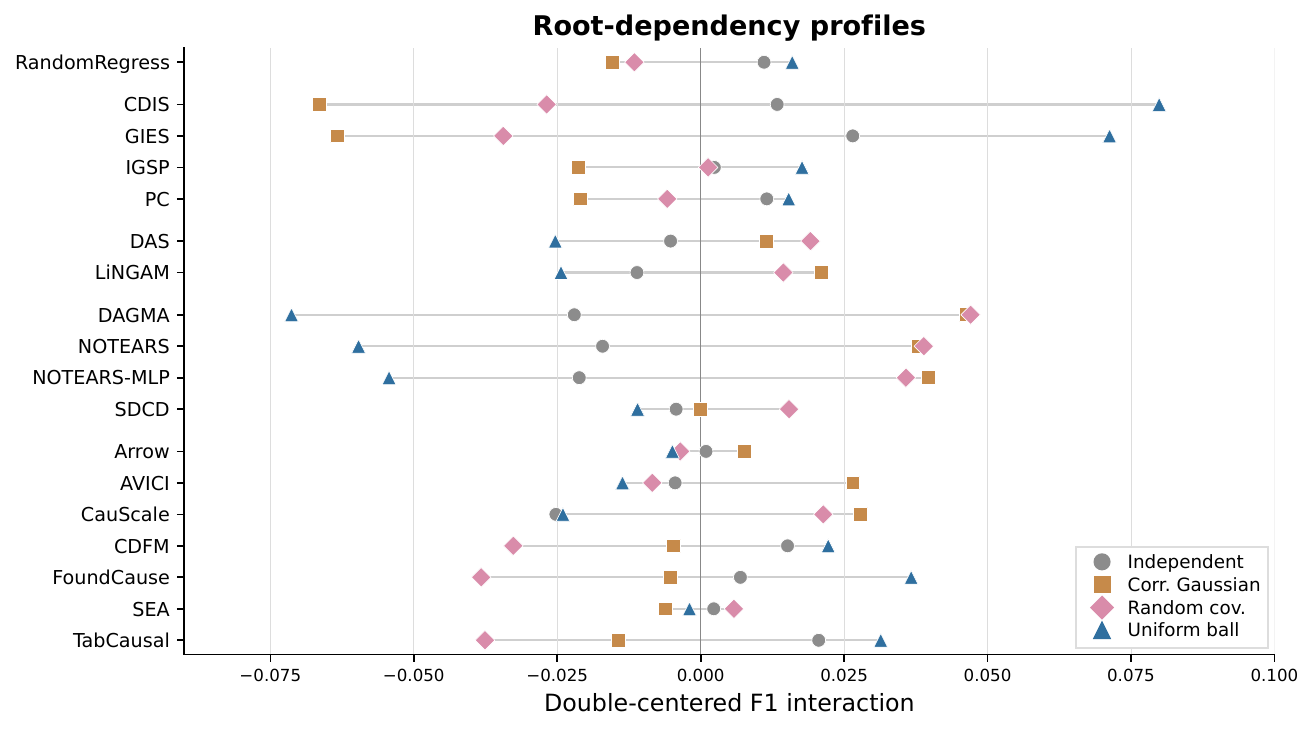}
    \caption{\textbf{Synthetic root-dependency interactions.} Each row is one method; markers are the four root-dependency categories, using double-centered F1. Wider rows are more root-sensitive. The uniform-ball versus common-factor (correlated-normal) reversal is the left--right swap of triangles and squares.}
    \label{fig:synthetic-root-dependency-interactions}
\end{figure}

\Cref{fig:synthetic-root-dependency-interactions} isolates graph-level dependence among root variables. Independent-root markers sit near zero for most methods, as expected for the setting closest to the standard SCM assumption. Search methods such as CDIS and GIES have wide rows: uniform-ball markers (triangles) lie on the positive side and common-factor correlated-normal markers (squares) on the negative side. NOTEARS, NOTEARS-MLP, and DAGMA reverse that order. A plausible explanation is that smooth root covariance violates exogenous independence and can be mistaken by graphical-search procedures as root--root edges or v-structures, whereas global continuous objectives can sometimes absorb this covariance with fewer structural disruptions. Bounded uniform-ball roots favor CDIS/GIES and FoundCause, consistent with a setting where support constraints and sharper conditional changes provide useful search or motif cues but are harder to express with smooth additive objectives. Together, these views show that a synthetic score can be traced to graph structure, local mechanisms, noise processes, and root-dependency patterns.

\subsection{Pretraining-Exposure OOD Score Analysis}
\label{sec:ood-score-analysis}

The synthetic factor analysis above shows where methods are relatively strong or weak. For pretrained and foundation-style methods we add an observation-only check on categories they do not list as pretraining support. For each method and factor axis, we split categories by whether they appear in that method's documented prior, and compare the method to a fixed graphical-search reference (CDIS, GIES, IGSP, and PC) on the same held-out categories:
\begin{equation}
    R_{m,a}^{\mathrm{OOD}}
    = \frac{F_{m,a}^{\mathrm{OOD}}-\bar F_{\mathrm{ref},a}^{\mathrm{OOD}}}
           {\bar F_{\mathrm{ref},a}^{\mathrm{OOD}}}.
    \label{eq:ood-relative-gain}
\end{equation}
The figure reports $100R^{\mathrm{OOD}}$ as a percent of the reference F1 (e.g.\ $+60\%$ means $60\%$ above the reference mean), not an additive F1 increment. We use this reference group because these four methods form a coherent graphical-search family among the evaluated baselines. They still rely on standard causal-discovery assumptions, but they are comparatively less tied to a specific parametric mechanism family, neural pretraining simulator, or fixed variable ordering. Their average is a useful proxy for how hard a category is under common non-pretrained graph-search procedures.

All four axes use the same scenario-level F1 rows as the main benchmark. For graph family, each scenario has one category; for mechanism, noise, and root-source axes, a scenario contributes to every factor category present in its SCM, and category scores are macro-averaged before forming ID and OOD summaries. The overall OOD score is the unweighted mean of $R_{m,a}^{\mathrm{OOD}}$ over these four factor axes. A positive value means that the method stays above the four-method graphical-search reference on held-out categories, and larger values mean a larger relative lead. We also report an ID--OOD gap: the method's relative gain on supported categories minus its relative gain on held-out categories. A larger positive gap means weaker generalization under this split, because the lead over the reference is more concentrated on documented support. Smaller gaps mean more stable generalization: a gap of zero, or a small negative gap, means the OOD relative gain is not below the ID relative gain.

Taking the union of all pretrained methods' documented priors would leave too few held-out categories for a stable factor-level analysis, and that union would keep shrinking as new pretrained models appear. A per-method split asks a narrower question: given what a method documents as pretraining support, does its relative advantage remain outside that support?

\begin{figure}[!htbp]
    \centering
    \begin{minipage}[c]{0.54\textwidth}
        \centering
        \input{tables/generated_results/analysis/ood_prior_coverage}
    \end{minipage}
    \hfill
    \begin{minipage}[c]{0.44\textwidth}
        \centering
        \includegraphics[width=\linewidth,height=0.32\textheight,keepaspectratio]{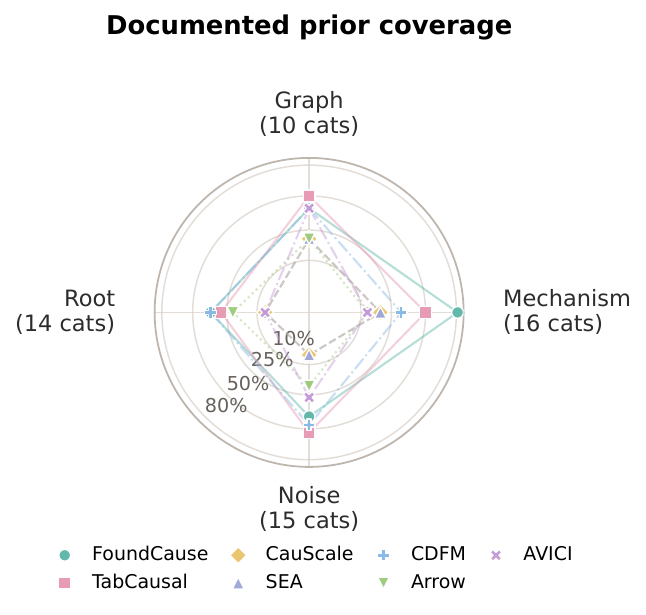}
    \end{minipage}
    \caption{\textbf{Documented prior coverage for pretrained methods.} Source-level priors are mapped to CausalArena graph, mechanism, noise, and root-family tags. The radar shows axis-wise coverage ratios on a square-root radial scale. Table rows are alphabetical. Cells are covered/total tags on that axis; Mean is the unweighted average of the four ratios, reported as a percentage. Bold is the highest value in a column and underline is second; ties share the mark. CauScale and SEA have the same mapped coverage, so their radar markers overlap.}
    \label{fig:ood-prior-coverage-radar}
\end{figure}

The prior labels are audited from the corresponding papers or code. For the plot, source-level priors are mapped conservatively to CausalArena's factor tags and counted only when they have a clear counterpart in the current synthetic pool. \Cref{fig:ood-prior-coverage-radar} visualizes this mapping as a documented prior-coverage profile over the four axes, with covered/total counts beside the radar. The mean column is the unweighted average of the four axis ratios, matching the four-axis averaging used for the OOD scores; we do not use radar-polygon area, which would mix axis order with the square-root radius. The full audited category mapping is released with the benchmark artifacts.

\Cref{fig:ood-prior-coverage-radar} shows that the pretrained methods differ in documented support: FoundCause has the broadest mapped mechanism coverage ($13/16$), TabCausal is more balanced across graph, mechanism, and noise axes ($5/10$, $8/16$, and $8/15$), CDFM covers a relatively broad noise/root set, and AVICI has reasonable graph-family coverage but narrow mechanism and root coverage. CauScale and SEA share the same mapped counts. Those coverage differences are why the check is defined per method.

\begin{figure}[!htbp]
    \centering
    \includegraphics[width=\textwidth]{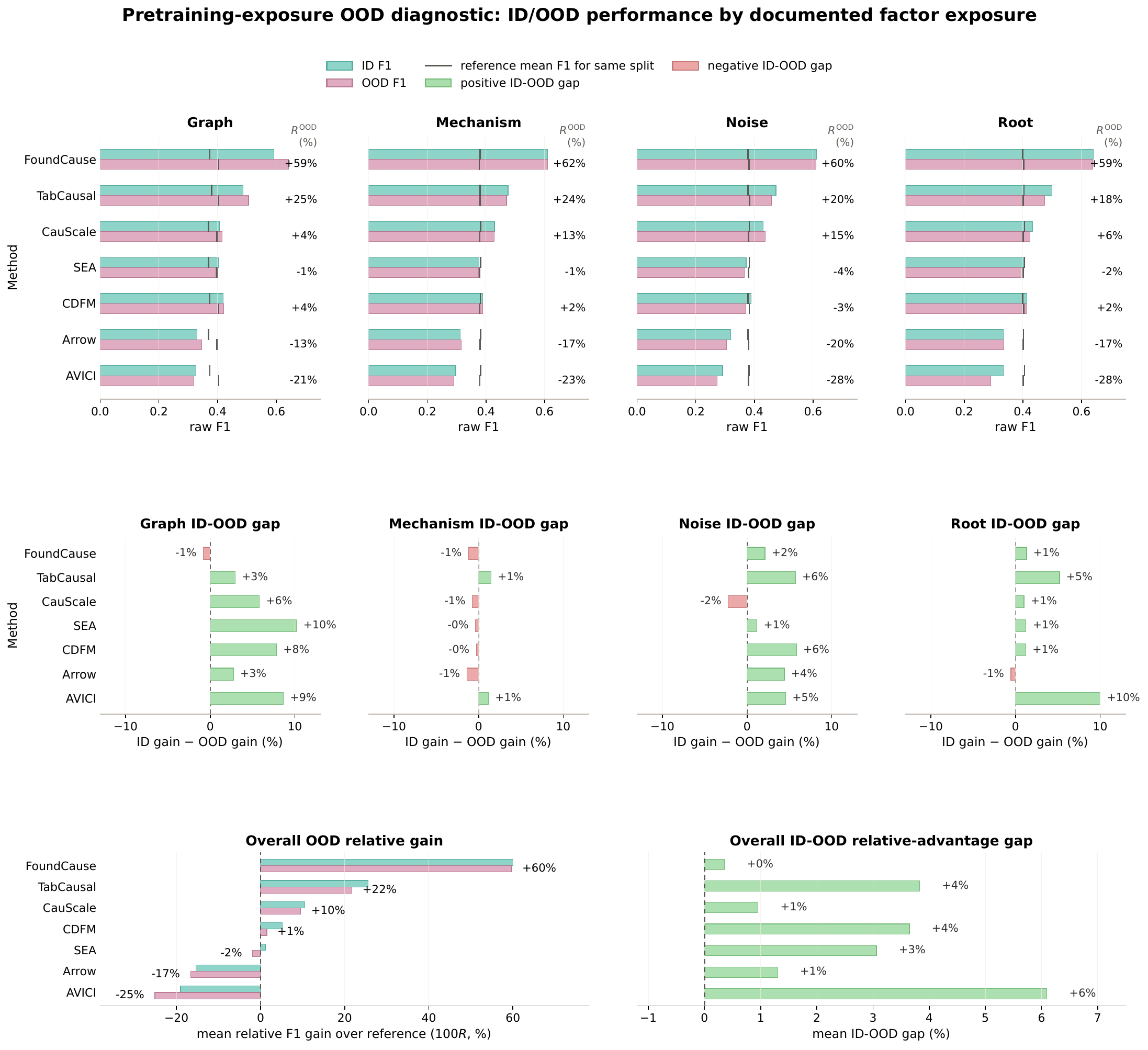}
    \caption{\textbf{Observation-only pretraining-exposure OOD diagnostics.} For each pretrained method we split synthetic obs-only categories into documented support (ID) and held-out categories (OOD), using that method's own audited prior. Columns are the four factor axes. \emph{Top row:} raw F1 on ID (teal) and OOD (pink). The dark tick on each bar is the graphical-search reference mean (CDIS, GIES, IGSP, PC) on the same ID or OOD slice; a bar to the right of its tick is better than the reference. The number at the right of each OOD bar is the relative gain $R^{\mathrm{OOD}}$ from \cref{eq:ood-relative-gain}, shown as a percent of the reference F1. \emph{Middle row:} ID relative gain minus OOD relative gain, in the same percent units. A larger positive value means a larger drop from ID to OOD in that relative lead, so generalization is weaker; smaller values mean more stable generalization, and a gap of zero or a small negative value means OOD is not worse than ID. Green is a positive gap (advantage shrinks on OOD); red is a negative gap (advantage grows on OOD). \emph{Bottom row:} unweighted means over the four axes. Left: overall $R^{\mathrm{OOD}}$, how far the method sits above the four-baseline mean on held-out categories, again as a percent. Right: the mean of the middle-row gaps, how much larger the ID relative gain is than the OOD relative gain; smaller values indicate more stable generalization.}
    \label{fig:ood-coverage-gain-panel}
\end{figure}

\Cref{fig:ood-coverage-gain-panel} then asks whether a method's lead survives on categories it does not document as pretraining support. The top row is easiest to read as a location relative to the reference tick. FoundCause's ID and OOD bars both sit well to the right of the ticks on every axis, with OOD relative gains near $+60\%$. TabCausal remains above the ticks with smaller but still positive gains. CauScale is modestly above the reference, most clearly on mechanism and noise. CDFM is close to the ticks; SEA straddles them. Arrow and AVICI fall to the left of the ticks on several axes, so their OOD relative gains are negative. Holding out undocumented categories therefore changes how we read the scores more than it changes the coarse ranking: FoundCause and TabCausal stay first and second, but the bars show whether that order is confined to documented support.

The bottom-left panel summarizes the same OOD comparison: it is the unweighted mean of $R^{\mathrm{OOD}}$ over the four axes, i.e., how far each method sits above the four-baseline mean on held-out categories. FoundCause remains far above that reference (about $+60\%$). TabCausal is next ($+22\%$), with positive gains on all four axes. CauScale is also positive overall ($+10\%$), especially on mechanism and noise, while CDFM is only slightly positive ($+1\%$) and SEA is near the reference ($-2\%$). Arrow and AVICI have negative overall OOD gains ($-17\%$ and $-25\%$).

The middle row and the bottom-right panel measure generalization as a drop from ID to OOD in relative gain. A larger positive bar means weaker generalization: the method's lead over the reference is more concentrated on documented support. Smaller values mean more stable generalization, and a gap of zero, or a small negative gap, means OOD is not worse than ID. Green is a positive ID--OOD gap. Under this reference group, $20$ of $28$ method--axis gaps are positive, and every overall mean gap is nonnegative or nearly zero. The bottom-right panel is the mean of those four middle-row gaps, so smaller values indicate more stable generalization. FoundCause is the exception that stays flat (overall gap $+0\%$), so its large lead is not concentrated on documented categories. TabCausal's advantage shrinks on held-out categories (overall gap $+4\%$) but remains substantial. CauScale's gap is small ($+1\%$). AVICI has the largest overall gap ($+6\%$), driven by graph ($+9\%$) and especially held-out root-family categories ($+10\%$), consistent with its narrow mapped root support ($1/14$). Arrow sits below the reference on most held-out axes; CDFM's positive OOD gains are smaller than those of FoundCause, TabCausal, and CauScale despite broader noise/root coverage. Broad documented pretraining support helps, but architecture, how coverage is balanced across factors, and how the model uses nonparametric graph signals still matter.

Those four-axis means also let us back out a crude full-ID / full-OOD F1 for the same average that \cref{fig:method-cost-frontier} reports. Let $F$ be that average F1, $c$ the unweighted mean documented coverage over the four axes, and
\begin{equation}
    \delta
    = \frac{1+\bar R^{\mathrm{ID}}}{1+\bar R^{\mathrm{OOD}}}-1
    \label{eq:ood-id-ood-ratio}
\end{equation}
the relative gap between the mean ID and mean OOD leads in \cref{fig:ood-coverage-gain-panel}. Treating the reported score as a coverage mix,
\begin{equation}
    F
    = c\,\hat F^{\mathrm{ID}} + (1-c)\,\hat F^{\mathrm{OOD}},
    \qquad
    \hat F^{\mathrm{ID}}=(1+\delta)\,\hat F^{\mathrm{OOD}},
    \label{eq:ood-coverage-mix}
\end{equation}
gives $\hat F^{\mathrm{OOD}}=F/(1+c\delta)$ and $\hat F^{\mathrm{ID}}=F(1+\delta)/(1+c\delta)$. This is only an estimate: $c$ and $\delta$ are four-axis means, and $F$ is the family--split average already used in \cref{fig:method-cost-frontier}.

\begin{figure}[!htbp]
    \centering
    \includegraphics[width=0.88\textwidth]{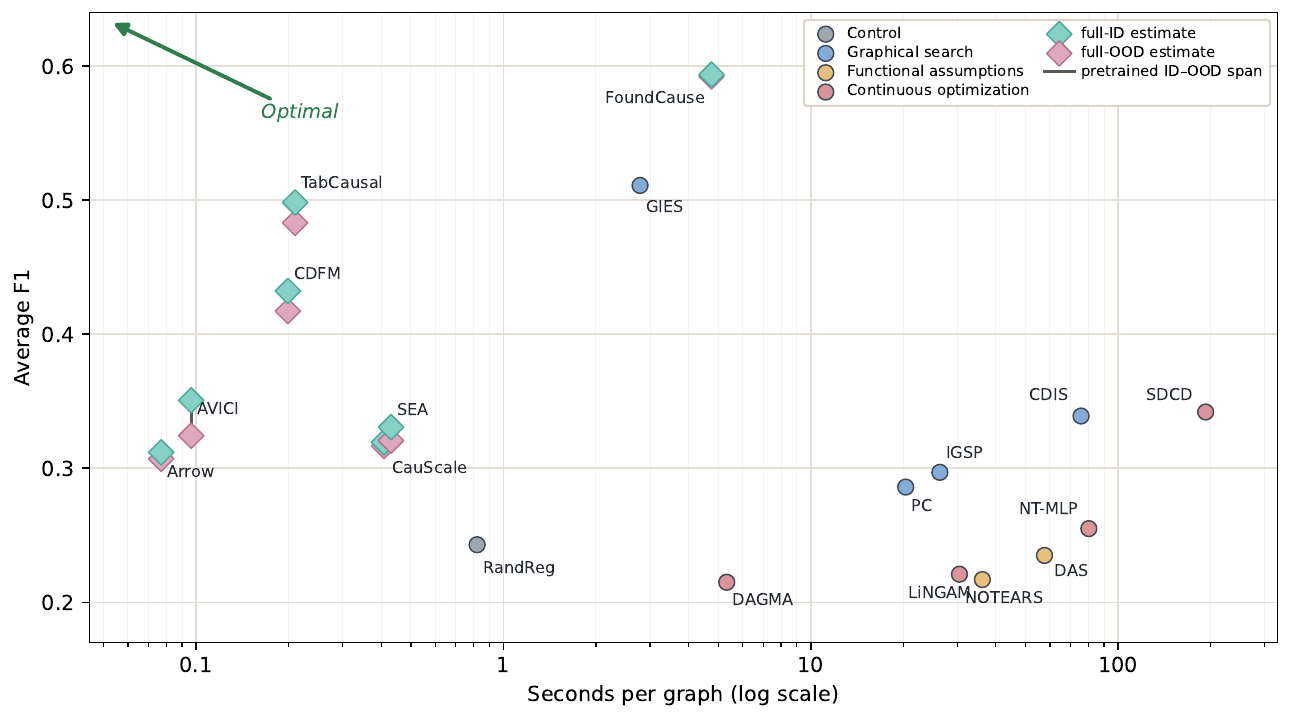}
    \caption{\textbf{Accuracy--cost view with full-ID / full-OOD estimates.} Same axes as \cref{fig:method-cost-frontier}: average F1 against comparable per-graph time, with pretrained $x$ equal to $T_{\mathrm{graph}}$. There is no Pareto curve. Each pretrained method is a vertical pair: the teal diamond is $\hat F^{\mathrm{ID}}$ and the pink diamond is $\hat F^{\mathrm{OOD}}$ from \cref{eq:ood-coverage-mix}; the observed $F$ lies on that segment. Circles are non-pretrained methods at their reported F1. \emph{Optimal} points toward higher F1 and lower time.}
    \label{fig:ood-method-cost-span}
\end{figure}

\Cref{fig:ood-method-cost-span} plots the same ranking as a vertical F1 interval. The segments are short. FoundCause stays near $0.59$ at either endpoint. TabCausal moves from about $0.48$ (full OOD) to $0.50$ (full ID) around its reported $0.49$. CDFM is $0.42$--$0.43$; AVICI has the longest relative span ($0.32$--$0.35$), consistent with its larger ID--OOD gap and narrower mapped support. CauScale, SEA, and Arrow barely move. The coarse pretrained order is therefore not an artifact of documented coverage: replacing the observed mix with either a fully ID or a fully OOD reading does not reorder FoundCause, TabCausal, and the cheaper lower-F1 group.

\subsection{Semantic and Formula Benchmark Behavior Analysis}
\label{sec:semantic-formula-behavior-analysis}

\paragraph{Semantic analysis protocol.}
The previous subsections use anonymous synthetic factors. Semantic and Formula SCMs let us look at operational processes and scientific formula families. We report self-centered F1 shifts: a method's F1 on a domain minus its own family average. Positive values mark domains where the method scores higher than its own family average; negative values mark domains where it scores lower. The plots show which semantic or scientific settings each method handles better or worse.

\paragraph{Example scenarios.}
\label{sec:example-scenarios}
\Cref{fig:semantic-formula-examples} and the examples in
\cref{sec:semantic_scm,sec:formula_scm} introduce two released SCMs: a wildfire
smoke / shelter process and a barometric pressure reduction to sea level.
We reuse those scenarios here when interpreting domain-level Semantic and
Formula behavior: a missing or extra edge is then a claim about that smoke day
(e.g., treating a consumer indoor reading as true indoor PM2.5, or dropping
filter provision as a parent of indoor concentration) or a claim about the
pressure reduction (e.g., writing the sea-level formula without fused height,
or treating a thermometer reading as if it were already layer-mean virtual
temperature).
Released Semantic and Formula SCMs, including these two, pass several review
and validation passes before they leave the construction pipeline
(\cref{sec:appendix-agentic-workflow}).

\begin{figure}[!htbp]
    \centering
    \includegraphics[width=0.95\textwidth]{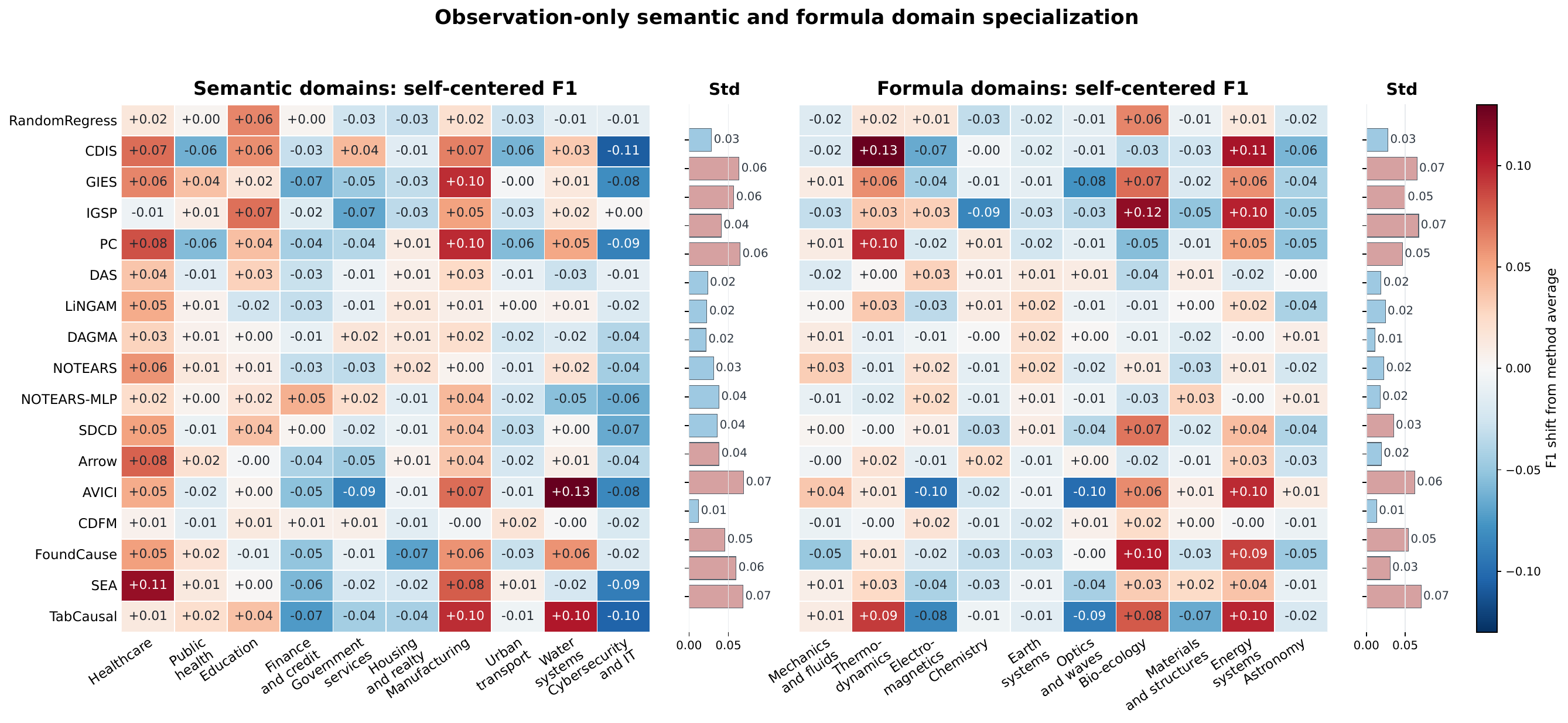}
    \caption{\textbf{Observation-only Semantic and Formula domain specialization.}}
    \label{fig:semantic-formula-domain-obs-profile}
\end{figure}

We focus on observation-only behavior, restricting the comparison to methods evaluated on the same split. In \cref{fig:semantic-formula-domain-obs-profile}, the heatmaps show method-domain self-centered F1 shifts, and the side bars report each method's standard deviation across domains. A larger standard deviation means that the method is more domain-sensitive after subtracting its own average performance.

\paragraph{Domain-level patterns.}
After each method's own average is removed, some domains still sit consistently above or below that baseline. On Semantic, cybersecurity and IT operations is usually weak: many methods score lower there than on their own Semantic average. Education, healthcare delivery, manufacturing, and water and sanitation are the usual strong ones, though not for every method. The side bars repeat this by method. DAS, LiNGAM, and DAGMA hardly move across Semantic domains; PC, CDIS, AVICI, and TabCausal move more. On Formula, energy systems and thermodynamics tend to be easier. Astronomy, electromagnetism, and optics and waves more often invert a law, pass through a geometric identity, or put a sensor chain around the equation, and that is where several methods lose relative F1.\looseness=-1

\paragraph{Method-level summaries.}
A domain heatmap only locates the gain or drop. It does not describe the processes or equations behind that domain. To read the Semantic and Formula benchmarks at that grain, we used an LLM agent to summarize each method's characteristic strengths and weaknesses from the observation-only results together with the scenario descriptions. \Cref{tab:semantic-formula-method-profiles} reports those summaries: how the method tends to behave on Semantic scenarios, how it tends to behave on Formula scenarios, and a short overall sketch.

\input{tables/generated_results/analysis/semantic_formula_pattern_tables}

\paragraph{Operational-semantic failure modes.}
The following two paragraphs paraphrase \cref{tab:semantic-formula-method-profiles}; they are LLM-assisted qualitative readings, not a separate human coding of errors. On Semantic cases, the summaries look better on a short visible chain and worse once backlog, drift, or a cascade takes over. RandomRegress looks strong only when the column order happens to match a simple sequence; it fails when the case needs several paths, coordination, or a behavioral response. PC, CDIS, IGSP, and GIES do better when a visible input goes into screening, inspection, or a staged workflow, and worse when demand, backlog, capacity, or an adaptive response moves through the rest of the system. IGSP also drops when rare events or hidden eligibility set the graph. DAS has only a mild edge on anomaly monitoring and escalation, and loses it on outreach and queues. LiNGAM prefers a one-way trigger-to-action chain, and drops when capacity, adherence, or saturation sets the outcome. DAGMA, NOTEARS, NOTEARS-MLP, and SDCD like one bottleneck, risk signal, triage step, or early warning that moves a measurable next action; they struggle with coordination, overflow, backlog, and long-run accumulation. Arrow fits detect-then-fix better than planning or scheduling. AVICI, CauScale, FoundCause, SEA, and TabCausal likewise prefer a short visible action that moves a downstream outcome, and they weaken on drift, spillover, workload, competing allocation, or cascade. CDFM is too flat in this release to assign a Semantic profile.

\paragraph{Scientific-formula failure modes.}
On Formula cases, they look better on a short equation chain, and worse once extra instruments sit around that equation. RandomRegress can match a few nearby variables by chance, then fail on coupled equations, conservation relations, and measurement chains. PC, CDIS, IGSP, and GIES handle small directed formulas better than shared parameters, extra outputs, or readout chains; IGSP is also weak on thermodynamic, absorption, and exchange relations. DAS prefers a direct physical-quantity chain over a biological, load, or pump-like coupled system. LiNGAM is more plausible on monotone scaling than on nonlinear response, buffering, oscillation, or several inputs. DAGMA and NOTEARS can fit a smooth or near-monotone formula, but not load, saturation, thermal, or interference structure. NOTEARS-MLP helps on a closed-form nonlinear formula, but not on control policies, threshold gates, or process feedback. SDCD likes power-law relations and not measurement or geometric readouts. CauScale is strong when calibrated physical inputs directly set a law output, and much weaker once control, calibration, or instrument loops wrap the law. FoundCause and TabCausal like a clear input-to-derived-quantity chain, and weaken when vessels, instruments, heat exchange, regime switches, or measurement chains sit around the law. Arrow handles some engineering and geometric relations, not biochemical curves or buffered systems. SEA prefers a classic single-law physics case over field, wave, or complex product relations. AVICI covers several common physical-law forms, but the score still moves with the formula type and the readout. CDFM again has no reliable Formula direction.

%% file: tables/generated_results/analysis/ood_prior_coverage.tex
{\footnotesize
\setlength{\tabcolsep}{3.2pt}
\renewcommand{\arraystretch}{1.18}
\begin{tabular}{@{}lccccc@{}}
\toprule
\textbf{Method} & \textbf{Graph} & \textbf{Mech.} & \textbf{Noise} & \textbf{Root} & \textbf{Mean} \\
\midrule
Arrow & 2/10 & 2/16 & 3/15 & 3/14 & 18\% \\
AVICI & \underline{4/10} & 2/16 & 4/15 & 1/14 & 22\% \\
CauScale & 2/10 & 3/16 & 1/15 & 1/14 & 13\% \\
CDFM & \underline{4/10} & 5/16 & \underline{7/15} & \textbf{5/14} & 38\% \\
FoundCause & \underline{4/10} & \textbf{13/16} & 6/15 & \textbf{5/14} & \textbf{49\%} \\
SEA & 2/10 & 3/16 & 1/15 & 1/14 & 13\% \\
TabCausal & \textbf{5/10} & \underline{8/16} & \textbf{8/15} & \underline{4/14} & \underline{45\%} \\
\bottomrule
\end{tabular}}

%% file: tables/generated_results/analysis/semantic_formula_pattern_tables.tex

\begingroup
\scriptsize
\setlength{\tabcolsep}{3.0pt}
\renewcommand{\arraystretch}{1.18}
\setlength{\LTpre}{4pt}
\setlength{\LTpost}{4pt}
\begin{longtable}{p{0.115\textwidth}p{0.255\textwidth}p{0.255\textwidth}p{0.255\textwidth}}
\caption{\textbf{Method-level behavior on the Semantic and Formula benchmarks.}
An LLM agent summarized each method's characteristic strengths and weaknesses from the observation-only scores and the scenario descriptions. The table reports those qualitative directions; it does not list raw scores.}\label{tab:semantic-formula-method-profiles}\\
\toprule
\textbf{Method} & \textbf{Semantic benchmark direction} & \textbf{Formula benchmark direction} & \textbf{Overall profile} \\
\midrule
\endfirsthead
\toprule
\textbf{Method} & \textbf{Semantic benchmark direction} & \textbf{Formula benchmark direction} & \textbf{Overall profile} \\
\midrule
\endhead
RandomRegress & Appears strong only when the variable order accidentally matches a simple operational sequence; collapses when the story requires distinguishing multiple paths, coordination, or behavioral response. & Shows the same order artifact on formula cases: it can match local support by chance, but fails on coupled equations, conservation relations, and measurement chains. & A sanity control, not a semantic reasoner; its large swings expose ordering and sparse-regression artifacts. \\
\specialrule{0.18pt}{1.4pt}{1.4pt}
CDIS & Favors locally clear signal-to-decision structures, where an observed input flows into a screening or triage step and then to an outcome; weakens when demand, policy response, backlog, or containment propagates through the system. & Handles small directed formula chains better than coupled multiplicative systems or shared-parameter equations. & Works best when the causal story is locally separable; hidden demand, shared drivers, and cascading structure blur the conditional signals it relies on. \\
\specialrule{0.18pt}{1.4pt}{1.4pt}
GIES & Prefers stable, sparse, stage-like workflows in which monitoring signals lead to interventions or outcomes; degrades under accumulated pressure, capacity limits, and cross-module propagation. & More reliable on static physical constraints than on readout chains, exponential/periodic transformations, or observation-heavy formula contexts. & Score/search behavior is strongest for stable staged mechanisms and less reliable when the observed variables are indirect traces of a latent process. \\
\specialrule{0.18pt}{1.4pt}{1.4pt}
IGSP & Performs well on observable service workflows such as processing, maintenance, vaccination, or patching; weakens when rare positives, hidden eligibility, or surveillance logic determine the graph. & Handles some direct multiplicative or power-law chains, but struggles with thermodynamic, absorption, and exchange-process relations. & Benefits from explicit workflow/intervention alignment; sparse event evidence and scientific readout layers make its direction tests less decisive. \\
\specialrule{0.18pt}{1.4pt}{1.4pt}
PC & Recovers local screening and inspection cascades better than adaptive, behavioral, or policy-response processes. & Handles low-coupling formula systems better than equations with shared parameters, multiple outputs, or derived readouts. & Conditional-independence search is useful when the semantic story decomposes into local tests, but brittle under adaptation, common drivers, and indirect measurements. \\
\specialrule{0.18pt}{1.4pt}{1.4pt}
DAS & Shows mild advantages on anomaly-monitoring and escalation chains, but loses advantage in outreach, engagement, and priority-queue settings. & Works better on direct physical-quantity chains than on biological, load, or pump-like coupled systems. & Has no sharp semantic specialization; it is mainly helped when the ordering signal is simple and directly observable. \\
\specialrule{0.18pt}{1.4pt}{1.4pt}
LiNGAM & Favors continuous, one-way trigger-to-action stories; weakens when capacity, adherence, follow-up management, or saturation controls the outcome. & More plausible on monotone scaling relations and weaker on nonlinear response, buffering, oscillation, or multi-input scientific systems. & Its linear non-Gaussian orientation prior can help on direct driving chains, but does not match capacity-limited or strongly nonlinear semantic processes. \\
\specialrule{0.18pt}{1.4pt}{1.4pt}
DAGMA & Works comparatively well when a bottleneck or allocation lever explains a measurable downstream outcome; weakens on distributed monitoring, tracing, and coordination chains. & Fits smooth, compressible formula skeletons better than load, saturation, or shared-mechanism systems. & Continuous optimization favors centralized, smooth causal structure; diffuse coordination and non-smooth process logic are harder to represent. \\
\specialrule{0.18pt}{1.4pt}{1.4pt}
NOTEARS & Favors direct risk-or-demand-to-action stories; degrades when the system involves propagation, containment, or multi-stage response. & Can recover some near-monotone physical chains, but loses reliability on complex multiplicative, thermal, and interference relations. & Linear DAG optimization is most useful when a semantic process can be compressed into a direct risk-to-action edge pattern. \\
\specialrule{0.18pt}{1.4pt}{1.4pt}
NOTEARS-MLP & The nonlinear SEM helps on triage and resource-allocation chains, but not enough for saturation, alert fatigue, overflow, or backlog dynamics. & Captures closed-form nonlinear formulas better than linear NOTEARS, yet remains weak on control policies, threshold gates, and process feedback. & Extra function flexibility helps, but system-level semantics such as accumulation, saturation, and feedback remain the limiting factor. \\
\specialrule{0.18pt}{1.4pt}{1.4pt}
SDCD & Performs better when an early warning signal leads to a resource intervention, and worse when capacity constraints or long-run accumulation modulate the effect. & Is relatively favorable on scale-law or power-law relations, and weaker on measurement readouts and geometric observation chains. & Most effective for monotone, observable intervention chains; less robust when a hidden state or measurement layer mediates the causal relation. \\
\specialrule{0.18pt}{1.4pt}{1.4pt}
Arrow & Favors event-response stories: an incident is detected and then remediated or mitigated. It is weaker on planning, scheduling, and resource-prioritization tasks. & Handles some engineering and geometric relations, but weakens on biochemical response curves and buffered systems. & Better aligned with event-response semantics than with strategic allocation or delayed planning. \\
\specialrule{0.18pt}{1.4pt}{1.4pt}
AVICI & Strongest when a concrete trigger or intervention changes a downstream metric; weaker in administrative coordination, workload, and prioritization cascades. & Covers several common physical-law structures, but its performance varies noticeably across formula types and readout contexts. & Its pretrained prior helps with local trigger-to-outcome chains, while multi-actor organization and instrumented scientific contexts remain harder. \\
\specialrule{0.18pt}{1.4pt}{1.4pt}
CauScale & Works best on short, separable signal-to-action stories; weakens on drift, spillover, allocation competition, and equilibrium-like processes. & Very strong when calibrated physical inputs directly determine a law output, but much weaker once control, calibration, or instrument loops wrap the formula. & Especially good at compact causal chains; system-level equilibrium, propagation, and control layers are the main failure mode. \\
\specialrule{0.18pt}{1.4pt}{1.4pt}
CDFM & Does not show a stable semantic specialization in this release; strong and weak residual cases are close after centering. & The formula readout is similarly flat, with no reliable direction that separates favorable and unfavorable scientific contexts. & The residual evidence does not support a clear method-specific semantic profile, so we avoid over-interpreting it. \\
\specialrule{0.18pt}{1.4pt}{1.4pt}
FoundCause & Favors direct readiness, allocation, and response chains; weakens on drift, spillover, cascade, and feedback-like processes. & Handles some clearly structured formula chains, but struggles when shared physical parameters, vessels, instruments, or heat-exchange contexts mediate the relation. & Structural refinement appears helpful for direct response chains, but less sufficient for long-range propagation and shared-mechanism systems. \\
\specialrule{0.18pt}{1.4pt}{1.4pt}
SEA & Favors short detection-to-intervention stories and quality/adherence-to-outcome chains; weakens under retention, allocation, cascade, and spillover dynamics. & Works better on classic single-law physics than on field-strength, wave/readout, or complex multiplicative relations. & Stronger on short protocol-like causal stories; slower accumulation and cross-system propagation remain difficult. \\
\specialrule{0.18pt}{1.4pt}{1.4pt}
TabCausal & Favors clear policy or operational levers whose effects flow to downstream outcomes; weakens on load propagation, backlog, and cascade-heavy systems. & Strong on direct formula pipelines from physical inputs to derived quantities, but weaker when instrumentation, regime switches, or measurement chains surround the base law. & Its strongest semantic profile is clear intervention point plus directed downstream chain; multi-stage propagation and instrumented readout chains remain the main weaknesses. \\
\bottomrule
\end{longtable}
\endgroup

%% file: discussion.tex
\section{Discussion}
\label{sec:discussion}

\subsection{Open and Evolvable Evaluation}

The most direct validation of a causal discovery claim is often feedback in a real system: whether intervening on a recovered parent changes the intended downstream quantity. During method development, however, such closed-loop checks are usually unavailable at scale, so the field relies on benchmarks with known graphs. \method{} is designed for that development setting. Using executable SCMs rather than only fixed sampled tables makes graph structure, mechanisms, interventions, and sampling procedures inspectable and reproducible, and it lets later releases add new SCMs while preserving the interfaces in \cref{sec:benchmark_design}.

For CDFMs, openness also changes how a score should be read. Once an SCM, its graph, or its generator becomes public, related environments may enter later pretraining. This does not make a public benchmark useless, but it does mean that a result should be interpreted together with the model's training context: performance on a familiar generator distribution and performance on newly introduced environments answer different questions. We therefore treat \method{} as an \emph{evolvable} arena. Versioned public releases provide reproducible reference evaluations, while the common SCM interface supports newly authored graphs, mechanisms, and scenarios as pretrained models evolve. This is how the freshness requirement in \cref{sec:benchmark_requirements} is pursued without giving up transparency.

\subsection{Interpreting Foundation-Model Results}
\label{sec:cdfm_release_discussion}

To make pretrained comparisons easier to read, \method{} ties results to explicit benchmark versions and release states. Leaderboard entries are encouraged to disclose whether public \method{} SCMs, generated samples, scenario metadata, or closely related generators were used during development or pretraining. The current package distinguishes public and reserved subsets: 500 synthetic, 50 semantic, and 50 formula-grounded SCMs are public, while the remaining audited SCMs are reserved for held-out leaderboard evaluation and progressive release in later benchmark versions. Results in this paper use the complete 1{,}200-SCM pool and are labeled by this benchmark version.

These practices make pretraining--evaluation overlap easier to discuss; they are not a claim that overlap can be removed once and for all. As earlier versions become public, their SCMs may eventually enter future training corpora. Continued evaluation of new foundation models therefore benefits from periodically extending the SCM pool and reporting scores together with the benchmark version and training disclosure. The semantic construction pipeline is one practical route for adding newly authored operational environments in later releases.

\subsection{Scope and Future Extensions}

The current release focuses on multivariate tabular causal structure recovery with known directed graph ground truth. Synthetic SCMs emphasize controlled breadth, semantic SCMs emphasize operational grounding, and formula-grounded SCMs emphasize explicit scientific mechanisms. Public real-world datasets serve as an external-validity check; their published graphs should be read as reference structures, with the caveat that they are not uniquely verified causal ground truth.

The same executable interface can later support additional directions: new domains and SCM families, larger graphs, latent confounding, and neighboring causal tasks such as intervention-target reasoning, effect estimation, and counterfactual inference. Such extensions can reuse the current versioning and evaluation infrastructure while remaining comparable with earlier releases.

%% file: conclusion.tex
\section{Conclusion}
We introduced \method{}, an evolvable benchmark for causal discovery in the foundation-model era. \method{} combines broad synthetic SCMs, semantically grounded operational SCMs, and formula-grounded scientific SCMs under a shared observational and interventional evaluation protocol. Together, these families support broad coverage, meaningful grounding, detailed diagnosis, and future extension with newly constructed causal environments.
Experiments across classical, neural, and pretrained causal discovery methods show that rankings vary substantially across SCM families, mechanisms, sample sizes, and intervention settings, and that strong performance in one evaluation regime does not necessarily transfer to others. For pretrained models, these results further highlight the importance of interpreting benchmark scores in light of possible pretraining--evaluation overlap. By coupling reproducible public reference suites with a versioned and extensible SCM interface, \method{} provides a foundation for continued evaluation as causal discovery methods and foundation models evolve.

%% file: appendix.tex
\section*{Appendix}
\addcontentsline{toc}{section}{Appendix}
\label{sec:appendix}

This appendix provides benchmark data details, the evaluation protocol, and complete numerical result tables.

\input{benchmark_data_details}

\input{evaluation_protocol}

\input{complete_results}

%% file: benchmark_data_details.tex
\section{Benchmark Data Details}
\label{sec:benchmark-data-details}

This section records the data construction details behind the current release. We use paper-facing names, and separate generator sampling rules from realized release statistics.

\subsection{Synthetic SCM Pool}
\label{sec:appendix-synthetic-pool}

The synthetic main benchmark contains 1{,}000 SCM configurations. It crosses 10 graph families with five dimensionalities $d\in\{10,20,30,50,100\}$ and 20 SCMs per graph-family/dimension pair. Each configuration is generated with both an observation-only split and an observation-plus-intervention split; generated files enter the release only after passing the audit rules in \cref{sec:appendix-synthetic-audit}.

\paragraph{Graph structure.}
\begin{itemize}[leftmargin=*]
    \item \textbf{Erdos--Renyi:} random ordered graphs with edge density sampled from 0.05--0.25.
    \item \textbf{Scale-free:} preferential-attachment-style skeletons oriented by an acyclic order, with density 0.04--0.22.
    \item \textbf{Chain:} sparse path-like DAGs with density 0.03--0.16 and long directed propagation.
    \item \textbf{Tree:} branching sparse DAGs with density 0.03--0.16.
    \item \textbf{Layered:} layer-respecting DAGs with density 0.04--0.20.
    \item \textbf{Bipartite layered:} two-stage or multi-stage bipartite-like DAGs with density 0.04--0.20.
    \item \textbf{Small-world:} locally clustered DAGs with short directed paths, density 0.04--0.20.
    \item \textbf{Block-sparse layered:} block-structured layered DAGs with sparse cross-block edges, density 0.04--0.20.
    \item \textbf{Dense local-module:} modular DAGs with denser within-module connectivity, density 0.06--0.24.
    \item \textbf{Collider-rich:} DAGs that deliberately increase v-structure frequency, density 0.06--0.24.
    \item \textbf{Validity filters:} candidate graphs are resampled until they are DAGs with no isolated variables, sufficient graph depth, and at least $\lceil0.8d\rceil$ directed edges. The optional in-degree cap is sampled from no explicit cap, 4, or 5 for $d\leq20$, from 4, 5, or 6 for $20<d\leq50$, and from 5, 6, or 8 for $d=100$.
\end{itemize}

\paragraph{Family mixing.}
\begin{itemize}[leftmargin=*]
    \item \textbf{Mechanism/noise family count:} the number of active local mechanism and noise families increases with dimension: 2--4 at $d=10$, 3--5 at $d=20$, 4--6 at $d=30$, 5--8 at $d=50$, and 6--10 at $d=100$.
    \item \textbf{Root-family count:} the number of active root families depends on the number of root variables: one family for a single root, 1--2 for at most three roots, 2--4 for at most eight roots, and 3--6 otherwise.
    \item \textbf{Local assignment:} after active families are selected, node-level mechanisms, root families, and noise families are assigned across the graph so that a single SCM can contain heterogeneous local structural equations and noise channels.
\end{itemize}

\paragraph{Mechanisms.}
\begin{itemize}[leftmargin=*]
    \item \textbf{Linear additive:} parent weights are scaled by $0.8/\sqrt{k}$ and biases are sampled in $[-0.5,0.5]$.
    \item \textbf{Polynomial additive:} powers 2--4 are used, with quadratic coefficients in $[-0.5,0.5]$ and cubic coefficients in $[-0.15,0.15]$.
    \item \textbf{Multiplicative parent product:} parent effects are multiplied after weight scaling by $0.8/\sqrt{k}$ and bias sampling in $[-0.4,0.4]$.
    \item \textbf{Random-function product:} transformed parent responses are multiplied and rescaled with output scale in $[0.4,1.6]$. Together with multiplicative parent product, this form is scored as one multiplicative family, so the analyses use 16 mechanism tags.
    \item \textbf{Additive interaction:} up to three parent pairs are sampled, with interaction weights scaled by $0.5/\sqrt{\#\mathrm{pairs}}$.
    \item \textbf{Rational/ratio:} denominator-safe ratio transforms avoid near-zero denominators through bounded offsets and clipping.
    \item \textbf{Exponential/log:} exponential inputs are clipped to $[-3,3]$ and paired with log or signed-log style transforms where needed.
    \item \textbf{Trigonometric/periodic:} periodic parent transforms use sine/cosine-style responses with randomized frequency and phase.
    \item \textbf{Piecewise/threshold:} cutoffs are sampled in $[-0.6,0.6]$, low slopes in $[0.2,1.0]$, high slopes in $[1.0,2.5]$, and offsets in $[-1.5,1.5]$.
    \item \textbf{Saturation/clipping:} saturating nonlinearities use response scales in $[0.8,2.5]$.
    \item \textbf{Mixture/regime switching:} gates use scales in $[1,4]$, cutoffs in $[-0.5,0.5]$, and alternate response scales in $[0.5,1.8]$.
    \item \textbf{Random neural:} one-hidden-layer neural mechanisms use 3--8 hidden units.
    \item \textbf{RFF/GP-style:} random-feature mechanisms use 64--256 features, length scales in $[0.25,3.0]$, and output scales in $[0.4,2.5]$.
    \item \textbf{Activation additive:} activation families include ReLU, leaky-ReLU, softplus, sigmoid, absolute value, sign, step, hard-tanh, ReLU6, ELU, SELU, SiLU, modulo, round, and rank transforms; modulo periods are in $[0.75,3.0]$ and rounding steps in $[0.2,1.0]$.
    \item \textbf{Tree-based:} shallow tree mechanisms use depth 2--4 with thresholds in $[-0.8,0.8]$.
    \item \textbf{Discretization/assignment:} discretized mechanisms use 3--7 bins or clusters.
    \item \textbf{Aggregation nonlinearities:} max or log-sum-exp aggregation uses temperatures in $[0.3,1.5]$.
\end{itemize}

\paragraph{Root distributions.}
\begin{itemize}[leftmargin=*]
    \item \textbf{Uniform:} lower bounds are sampled in $[-2,0]$ and upper bounds in $[0.5,3.0]$.
    \item \textbf{Gaussian:} the mean is drawn from $\mathcal{N}(0,1)$ and the standard deviation from $[0.5,2.0]$.
    \item \textbf{Truncated Gaussian:} the mean is drawn from $\mathcal{N}(0,0.8^2)$, the standard deviation from $[0.4,1.4]$, and samples are clipped to mean $\pm2.5$ standard deviations.
    \item \textbf{Lognormal:} the log mean is sampled in $[-0.3,0.7]$ and the log standard deviation in $[0.25,0.9]$.
    \item \textbf{Gamma:} concentration is sampled in $[1,5]$ and rate in $[0.5,2]$.
    \item \textbf{Scaled beta:} shape parameters are sampled in $[0.8,5]$, then values are mapped to a sampled interval with lower endpoint in $[-1,0]$ and upper endpoint in $[0.8,3.0]$.
    \item \textbf{Exponential:} the rate is sampled in $[0.4,2.5]$.
    \item \textbf{Gumbel:} the location is sampled in $[-1,1]$ and the scale in $[0.3,1.8]$.
    \item \textbf{Bernoulli:} probabilities are sampled in $[0.15,0.85]$.
    \item \textbf{Poisson:} rates are sampled in $[1,12]$.
    \item \textbf{Ordinal categorical:} the number of ordered levels is sampled from $\{3,\ldots,6\}$; category probabilities are produced by a softmax over i.i.d. normal logits, and samples take ordered values $0,\ldots,k-1$.
    \item \textbf{Mixture Gaussian:} the mixture probability is sampled in $[0.2,0.8]$, the two means in $[-2,-0.2]$ and $[0.2,2]$, and component standard deviations in $[0.3,1.0]$.
    \item \textbf{Multinomial discrete:} the number of unordered categories is sampled from $\{2,\ldots,10\}$; logits are sampled in $[-1.5,1.5]$, converted to probabilities by softmax, and the sampled category index is standardized.
    \item \textbf{Zipf/power-law:} support size is sampled from $\{8,\ldots,20\}$, probabilities are proportional to $v^{-\alpha}$ with $\alpha\in[2,4]$, sampled values are capped at 10, and the result is centered.
    \item \textbf{Root dependency:} 800 of the 1{,}000 synthetic SCM configurations use independent standardized Gaussian root blocks. The remaining 200 use multi-root dependent blocks: a common-factor Gaussian with $\rho\in[0.25,0.75]$ and random signs; a random-covariance Gaussian whose latent dimension is sampled from $1,\ldots,r$ for $r$ root variables with covariance $LL^\top+0.15I$; or a uniform-ball sampler that draws a random direction, radius $U^{1/r}$, scale in $[0.75,2.0]$, and center coordinates in $[-0.5,0.5]$ before column standardization.
\end{itemize}

\paragraph{Structural noise.}
\begin{itemize}[leftmargin=*]
    \item \textbf{Gaussian:} additive Gaussian noise uses scale $[0.05,0.45]$.
    \item \textbf{Uniform:} additive bounded noise uses scale $[0.05,0.45]$.
    \item \textbf{Laplace:} heavy-tailed additive noise uses scale $[0.03,0.35]$.
    \item \textbf{Student-$t$:} degrees of freedom are sampled in 2.5--8 and scale in $[0.03,0.30]$.
    \item \textbf{Multiplicative lognormal:} multiplicative noise uses $\sigma\in[0.03,0.25]$.
    \item \textbf{Heteroskedastic:} the scale is $\mathrm{base}+\mathrm{slope}\cdot\sigma(|s|)$ with base in $[0.02,0.15]$ and slope in $[0.02,0.20]$.
    \item \textbf{RFF-heteroskedastic:} parent values pass through 64--256 random Fourier features with length scale $[0.25,3.0]$ and output scale $[0.3,1.8]$; a softplus transform converts this latent function into a sample-specific noise scale.
    \item \textbf{Target-$R^2$ Gaussian:} the noise variance is chosen from the empirical signal variance to target $R^2\in[0.1,0.9]$.
    \item \textbf{Mixture-outlier:} a base scale in $[0.02,0.15]$ is mixed with rare outliers whose probability is $[0.01,0.08]$ and scale is $[0.8,2.5]$.
    \item \textbf{Quantized:} Gaussian-noisy values are rounded to a grid with step size $[0.05,0.35]$.
    \item \textbf{Censored:} Gaussian-noisy values are clipped at their empirical 3rd and 97th percentiles.
    \item \textbf{Beta:} centered beta noise uses shape parameters in $[1,10]$ and scale in $[0.05,0.45]$.
    \item \textbf{Exponential:} centered one-sided disturbances use rate $[0.8,4.0]$ and scale $[0.03,0.35]$.
    \item \textbf{Gumbel:} centered extreme-value disturbances use scale $[0.03,0.35]$.
    \item \textbf{No-noise option:} deterministic local equations are allowed, but the generator limits no-noise assignments so that at most 10\% of nodes in a graph use this option.
\end{itemize}

\paragraph{Difficulty variants.}
\begin{itemize}[leftmargin=*]
    \item \textbf{Schedule:} within each graph-family/dimension block, the generator cycles through one clean baseline case, two single-stressor cases, and one paired-stressor case.
    \item \textbf{Signal and noise stressors:} weak-signal cases multiply structural mechanism strength by 0.45; high-noise cases add extra Gaussian structural noise with scale 0.35 after the selected node noise.
    \item \textbf{Observation stressors:} feature warping applies signed-log, tanh, or signed-square-root transforms to about 55\% of columns; outlier contamination perturbs about 2.5\% of entries with heavy-tailed jumps; missingness/imputation affects about 5\% of entries and fills them with column means plus small jitter; feature corruption permutes about 3\% of entries within the same column.
    \item \textbf{Intervention stressors:} interventional distribution shift applies shift/scale perturbations to selected interventional columns; soft interventions blend the original structural value with the intervened value using softness 0.45.
\end{itemize}

\paragraph{Interventions.}
\begin{itemize}[leftmargin=*]
    \item \textbf{Main split:} the obs+int split uses $800$ observational rows and $200$ interventional rows.
    \item \textbf{Target-pool size:} each SCM samples an intervention target pool whose size grows with dimension: 2--4 targets for $d=10$, 3--6 for $d=20$, 4--8 for $d=30$, 5--12 for $d=50$, and 8--20 for $d=100$.
    \item \textbf{Row assignment:} every interventional row selects one target from this pool using balanced assignment, applies a hard do-resampling intervention, records the binary intervention mask, and recomputes all descendants.
    \item \textbf{Intervention value:} the do-resampling value is drawn uniformly from the target's empirical baseline 10th--90th percentile range. Intervention targets may include root or structural variables unless a sensitivity suite explicitly restricts targets to structural nodes.
\end{itemize}

\subsection{Synthetic Audit and Protocol-sensitivity Suites}
\label{sec:appendix-synthetic-audit}

Synthetic SCMs are audited before release. Graph-level checks require an acyclic graph, no isolated variables, minimum depth 3 for chain/tree/small-world/scale-free families and 2 otherwise, and at least $\lceil0.8d\rceil$ edges. Dataset-level checks require finite values, no constant or near-constant continuous variables, no discrete variable with a single category occupying more than 99\% of rows, boundary mass at explicit clipping boundaries at most 10\%, no near-duplicate variable pair above correlation 0.99, and intervention masks consistent with the target pool. For obs+int files, every recorded target must receive at least one affected row, target values must change relative to their observational mechanism, and descendants are regenerated under the intervened values.

The sample-size suite uses a fixed $d=30$ representative subset of the synthetic pool, with 10 SCMs per graph family and $n\in\{100,200,500,1000,2000,5000,10000\}$. In the main text we report the observation-only sample-size results to isolate observational sample count. The intervention-protocol suite also uses 100 SCMs at $d=30$, but changes the intervention generator while keeping the same graph/mechanism pool:
\begin{itemize}[leftmargin=*]
    \item \textbf{Default resampling} uses the main protocol: $800$ observational rows, $200$ interventional rows, one target per interventional row, all sampled targets eligible, and hard do-resampling from the target's 10th--90th percentile range.
    \item \textbf{Fixed setpoint} keeps the $800/200$ split and one target per row, but sets targets to one of their 20th, 50th, or 80th percentile baseline values with jitter equal to 1\% of the interquartile range.
    \item \textbf{Parameter shift} keeps the $800/200$ split, restricts targets to structural non-root nodes, and changes the target mechanism by scaling parent effects, shifting a bias/threshold by 0.3--0.8 interquartile ranges, or changing nonlinear strength by a factor sampled from 0.6--0.85 or 1.15--1.6.
    \item \textbf{High intervention fraction} uses $200$ observational rows and $800$ interventional rows with single-target hard resampling.
    \item \textbf{Multi-target per sample} keeps the $800/200$ split but samples 2--4 targets per interventional row and applies hard resampling to each selected target.
    \item \textbf{Dense mixed intervention} combines the harder settings: $200$ observational rows, $800$ interventional rows, 2--4 structural targets per interventional row, and a balanced mixture of hard resampling, fixed-setpoint interventions, and parameter shifts.
\end{itemize}

\subsection{Agentic Construction Workflow and Quality Gates}
\label{sec:appendix-agentic-workflow}

Semantic and Formula SCMs are built with staged agentic workflows. We describe the paper-facing stages and quality gates without exposing implementation filenames. Each stage leaves an auditable record, and failed checks are routed back to graph design or executable specification before release.

\paragraph{Stage 1: seed and reference grounding.}
The benchmark starts from human-curated seed banks. Semantic seeds cover a fixed 10-domain plan with 10 scenarios per domain and specify the setting, causal question, measurement context, plausible actions or interventions, and the operational ambiguity to expose. Formula seeds specify a scientific task family, possible formula anchors or formula-selection policy, response-variable intent, units, validity conditions, and experimental context. The first agentic stage collects compact grounding evidence: Semantic scenarios record concrete variables, records, sensors, actions, queues, delays, capacities, and outcomes; Formula scenarios record equations, constants, units, validity domains, instruments, calibration/readout variables, and causal-orientation constraints.

\paragraph{Stage 2: structure and mechanism planning.}
Before any full graph is written, the workflow diagnoses several structurally distinct options and selects a candidate. The graph-design target is 15--25 observed variables with batch-average size close to 20; the Formula validator keeps a wider 10--30-node safety envelope for justified deviations, although the released Formula pool remains within 16--25 variables. Edge count is controlled by the selected per-scenario structural plan, with no single global quota; each plan must state expected edge-count, depth, root-count, and sink-count ranges, and the later graph must report its realized node, edge, root, sink, depth, and motif alignment. A mechanism-execution plan then binds each planned node to a root sampler, mechanism family, noise family, and range policy. This plan must include at least three scenario-justified executable mechanism families and normally keeps the allowed fraction of generic bounded-sigmoid children at or below 0.25.

\paragraph{Stage 3: graph drafting, review, and repair.}
The graph stage turns the selected plan into variables, directed edges, intervention handles, and local mechanisms. Every variable must define its operational meaning, measurement scale, unit, numerical range, data source or sampling story, observation process, intervenability status, and child-level structural mechanism over exactly its graph parents. Every edge must have a local mechanism explanation. Semantic reviews check causal plausibility, measurement realism, no vague latent placeholders, no hidden parents, realistic intervention semantics, and no alias, unit-conversion, or tiny-noise near-copy variables. Measurement, score, status, and readout nodes are kept only when they change the causal mechanism, observation process, intervention semantics, or diagnostic logic. Formula reviews additionally check formula correctness, unit compatibility, validity ranges, solve-for and response-variable orientation, invalid reverse orientations, and justified calibration/readout nodes. Blocking findings trigger graph revision and re-review.

\paragraph{Stage 4: executable translation.}
After graph approval, a translation table maps graph variables and edges into executable root distributions, structural operations, node-level noise, range policies, intervention resamplers, saved adjacency entries, and metadata. The executable specification must use legal structured operations over graph parents. It cannot introduce hidden parents, omit graph parents from the child operation, use batch-level row statistics, or rely on prose-only mechanisms. Formula specifications also store equation identifiers, inputs, outputs, constants, validity constraints, and residual checks; formula outputs are recomputed from parents and are never clipped merely to pass a range check. Interventional rows resample the flagged targets, record the binary intervention channel, and recompute descendants.

\paragraph{Stage 5: deterministic validation and release gate.}
Final validation checks schema validity, acyclicity, unique node identifiers, valid edge references, finite generated values, nonconstant columns, compatible observation-only and mixed-interventional files, target changes under intervention, descendant recomputation, and agreement among saved adjacency, generated columns, masks, graph metadata, and executable specification. The data gates require continuous variables to avoid more than 10\% exact mass at a valid-range boundary unless the graph declares a real boundary mechanism; continuous/non-discrete near-copy pairs are revised when absolute correlation reaches 0.90 without scientific justification; binary, categorical, ordinal, and grid-valued nodes are revised when one category dominates without a rare-event or pass-heavy justification. Formula SCMs must additionally pass residual checks for non-intervened formula rows within numerical tolerance. Anti-template checks block duplicate support topologies, repeated coarse role topologies, slot-filled mechanism text, generic readout/flag/calibration padding, and mechanism/root/noise-family dominance. Only scenarios that pass review, executable validation, adversarial audit, and the final release gate enter the benchmark package.

\subsection{Semantic SCM Pool}
\label{sec:appendix-semantic-pool}

Semantic scenario cards target operational DAGs with observable variables, actionable intervention handles, and scenario-specific measurement channels. The released semantic pool contains 100 operational scenarios, with 10 scenarios in each of 10 domains: cybersecurity and IT operations, education, finance and credit, government services, healthcare delivery, housing and real estate, manufacturing, public health, urban transportation, and water and sanitation. The realized release contains 17--25 variables (mean 22.89), 25--67 directed edges (mean 40.28), and 1--12 intervention targets (mean 3.82). Core metadata fields include variable definitions, units/ranges, observability, intervenability, parent lists, sampling policies, edge rationales, mechanism descriptions, monotonicity/sign tags, intervention resamplers, raw-value summaries, graph artifact paths, and generator-spec artifact paths. For example, the school-start-time scenario records that delaying the bell changes planned pickup schedules and circadian alignment, while grace-window policy affects whether an arrival is counted as tardy.

\begin{figure}[!htbp]
    \centering
    \includegraphics[width=\textwidth]{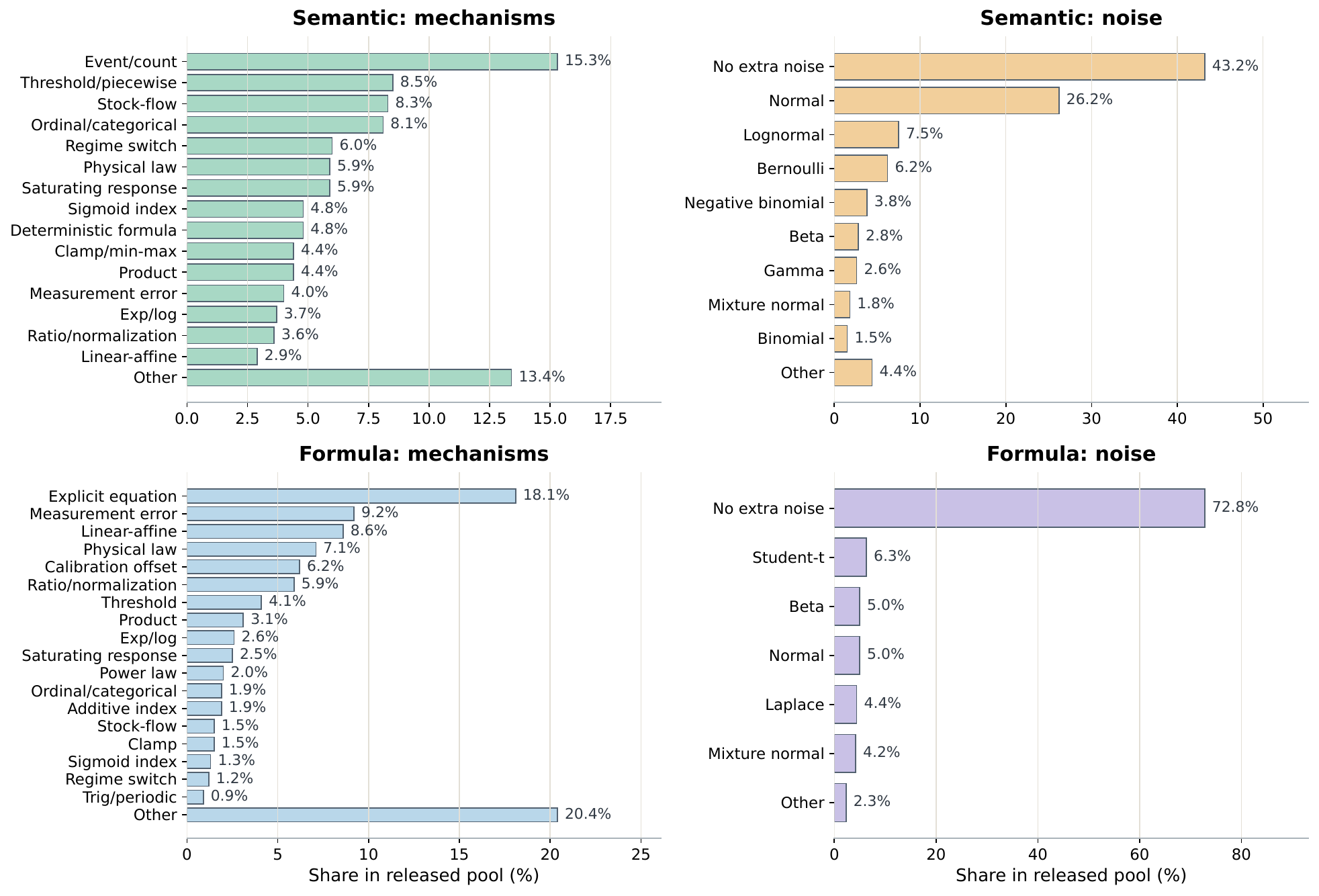}
    \caption{\textbf{Composition of the released Semantic and Formula pools.} Mechanism bars are edge shares; noise bars are node shares, from one representative file per released SCM. Remaining unlabeled families are grouped as Other. Formula nodes themselves have no extra noise, so that bar is large.}
    \label{fig:semantic-formula-composition}
\end{figure}

\Cref{fig:semantic-formula-composition} reports the released Semantic mix by mechanism and by noise. The largest mechanism shares are event/count processes ($15.3\%$), cutoffs ($8.5\%$), stock-flow accumulation ($8.3\%$), and ordinal rules ($8.1\%$). On the noise side, $43.2\%$ of nodes have no extra noise; the rest are led by normal ($26.2\%$), then lognormal, Bernoulli, and count-style families.

\subsection{Formula-grounded SCM Pool}
\label{sec:appendix-formula-pool}

Formula scenario cards target scientific DAGs whose key dependencies are grounded in explicit equations, causal-orientation contracts, units, validity ranges, and measurement/readout semantics. The released formula pool contains 100 equation-grounded scenarios, with 10 scenarios in each of 10 domains: biology and ecology, astronomy, chemistry, mechanics and fluids, earth systems, electromagnetism, energy systems, materials and structures, optics and waves, and thermodynamics. The realized release contains 16--25 variables (mean 22.68), 18--50 directed edges (mean 32.97), 1--17 intervention targets (mean 5.78), and explicit equation/residual metadata for formula-derived variables. Core metadata fields include variable definitions, units/ranges, constants, formula identifiers, equation inputs/outputs, solve-for variables, causal-orientation contracts, invalid reverse-orientation notes, measurement readouts, intervention resamplers, raw summaries, and residual checks on non-intervened rows. For example, the thin-lens scenario records the lens equation for target image distance, the Airy-disk diffraction relation, and the magnification equation, then adds actuator backlash, exposure, saturation, and quality-control readouts downstream.

\Cref{fig:semantic-formula-composition} also shows the Formula mix. The largest mechanism shares are explicit equations ($18.1\%$), measurement-error edges ($9.2\%$), linear-affine relations ($8.6\%$), physical laws ($7.1\%$), and calibration offsets ($6.2\%$). Formula nodes are computed exactly from their parents, so they carry no extra noise; that is why ``No extra noise'' is $72.8\%$ of nodes. The remaining noisy nodes are mostly measurement and readout terms: Student-$t$, beta, normal, Laplace, or mixture-normal. Both families use do-style intervention records with descendant recomputation. The difference is what the edges write down: a running process in Semantic, a named equation plus instruments in Formula.

%% file: evaluation_protocol.tex
\section{Detailed Evaluation Protocol}
\label{sec:evaluation_protocol}

This section records the evaluation conventions needed to interpret the reported results. File schemas, wrapper I/O details, and runnable commands are released with the code; here we summarize the protocol choices affecting comparability.

\subsection{Evaluation Settings}

Each executable SCM is evaluated under two exported splits from the same underlying graph. The \emph{obs-only} split contains only unperturbed samples and uses $n_{\mathrm{obs}}=1000$ in the main leaderboard. The \emph{obs+int} split contains observational rows plus do-style interventional rows with descendant recomputation and uses $n_{\mathrm{obs}}=800,n_{\mathrm{int}}=200$ in the main leaderboard. Intervention targets are recorded in the exported masks. Methods that do not consume intervention indicators are evaluated only on supported splits and are marked missing otherwise.

Each generated SCM benchmark scenario or synthetic configuration is evaluated with 5 random replicates, a fixed compromise between wall-clock cost and stable aggregate estimates. For a fixed method and split, we run the method on each replicate, score against the shared ground-truth adjacency, average replicate scores to a scenario/configuration-level value, and then aggregate over families, domains, dimensions, or the full arena. Failed, timed-out, or invalid runs are recorded explicitly and never silently removed.

\subsection{Metrics}

Given a predicted directed graph $\hat{G}$ and ground-truth adjacency $G$, we report directed-edge precision, recall, F1, and structural Hamming distance (SHD). Higher F1 and lower SHD are better. Compact tables report F1/SHD as mean\scorestd{standard deviation}; full F1 and SHD tables appear in \cref{sec:appendix-complete-results}. For PC and GIES, remaining undirected CPDAG edges are randomly oriented with a fixed per-file seed before these DAG metrics are computed, so the scores are directed-graph scores, not equivalence-class scores. Appendix complete tables additionally report structural intervention distance (SID), the number of ordered pairs whose intervention distributions would be misidentified from $\hat{G}$; normalized SHD $\mathrm{nSHD}=\mathrm{SHD}/(d(d-1))$; and, when a method returns edge scores, AUROC and average precision (AP). Hard-graph methods without edge scores therefore have missing AUROC/AP entries. Runtime and failure status are reported separately from accuracy metrics.

\subsection{Runtime, Hardware, and Timeout Policy}

GPU-based wrappers were evaluated on NVIDIA RTX 4090 GPUs when GPU inference was required. CPU baselines ran on AMD EPYC 7763 CPU cores with bounded process-level parallel workers. The current audited protocol uses a one-hour wall-time budget per graph row. Runs are executed in shards, typically with five graph/replicate rows per shard, so the scheduler assigns a shard-level budget proportional to the number of rows plus a small startup guard. The scheduler records completed rows, partial rows, empty failures, timeout statuses, and elapsed time in status files.
The main-text runtime ranking uses two protocols over the same synthetic mix: mean end-to-end wrapper time on the synthetic main benchmark for non-pretrained methods, and nested-$k$ load/inference fits on a sample balanced across dimensions and graph families for pretrained methods.

Reduced configurations are used only when the full configuration has been verified to be impractical under the benchmark budget, and these cases are recorded. The current real-data table is complete after these recorded repairs. In the real-data obs-int-as-obs conversion, PC timed out on four larger datasets (Wind Tunnel, Light Tunnel, PetShop temporal traffic 1, and PetShop temporal traffic 2) after a long best-effort budget; the reduced repair uses $\alpha=0.001$, at most 200 sampled rows, constant-column dropping, and fixed-seed random orientation, with predictions mapped back to the original variable dimension. CDIS timed out or failed to complete on the same four datasets; its reduced repair caps the conditioning-set size at 1 and preserves the native PAG output. SEA required a preprocessing repair on the two Causal Chamber datasets because constant columns produced a smaller returned matrix; the repaired wrapper drops constant columns internally and maps predictions back to the original dimension.

\subsection{Method Settings}

Unless noted below, methods are run once for the leaderboard with wrapper defaults and no hyperparameter search on benchmark labels. Library defaults not listed below are left unchanged by the wrapper. For pretrained or amortized methods, the checkpoint/source named below is fixed across generated-SCM and real-data runs. Methods that output equivalence-class objects preserve the native output when available, and directed-edge F1/SHD are computed after the deterministic conversion rules below.

\begin{itemize}[leftmargin=*]
    \item \textbf{RandomRegress.} A fixed-seed random-order regression baseline. It outputs a directed graph from nonzero regression support.
    \item \textbf{CDIS.} Uses observational and intervention groups under the wrapper default conditioning search. For directed DAG metrics, PAG endpoint uncertainty is converted to a directed graph using the wrapper's liberal orientation rule. The reduced repair caps conditioning-set size at 1 only for cases verified to exceed the budget.
    \item \textbf{GIES.} Uses the interventional score with the BIC-style penalty left at the wrapper default. Undirected edges in the returned essential-graph/CPDAG-like object are randomly oriented with a fixed per-file seed for DAG metrics.
    \item \textbf{IGSP.} Uses Gaussian CI-style tests with $\alpha=\alpha_{\mathrm{inv}}=10^{-3}$, depth 4, and 5 runs by default. It reports a directed graph directly.
    \item \textbf{PC.} Uses stable PC with Fisher-$z$ CI tests and $\alpha=0.05$ by default. Undirected CPDAG edges are randomly oriented with a fixed per-file seed for DAG metrics. The reduced real-data repair uses $\alpha=0.001$ and at most 200 rows after the full configuration times out.
    \item \textbf{DAS.} Uses the additive spline wrapper with pruning parameters $\eta_G=\eta_H=0.001$, $\alpha=0.05$, 10 splines, degree 3, and parent range 5--20.
    \item \textbf{LiNGAM.} Uses the standard DirectLiNGAM wrapper and reports the returned directed graph.
    \item \textbf{DAGMA.} Uses the wrapper's automatic linear/nonlinear selection with default optimization settings and default weight threshold 0.3.
    \item \textbf{NOTEARS.} Uses linear NOTEARS with $\lambda_1=0.1$ and fixed adjacency threshold 0.3.
    \item \textbf{NOTEARS-MLP.} Uses hidden size 10, $\lambda_1=\lambda_2=0.01$, max iteration 100, and fixed adjacency threshold 0.3.
    \item \textbf{SDCD.} Uses the two-stage wrapper with the default stage-1/stage-2 schedule and model threshold 0.1; high-dimensional budget reductions are logged when invoked.
    \item \textbf{Arrow.} Uses the released ArrowFM base checkpoint \href{https://huggingface.co/ryan-thompson/arrowfm-base/blob/main/arrowfm-base.pt}{\texttt{arrowfm-base.pt}} and its official decoded binary output.
    \item \textbf{AVICI.} Uses the released AVICI \href{https://huggingface.co/larslorch/avici/tree/main/scm-v0}{\texttt{scm-v0}} pretrained model (checkpoint \texttt{checkpoint\_0300000.pkl}), loaded by \texttt{avici.load\_pretrained(download="scm-v0")}, and fixed threshold 0.5 for probabilistic adjacency outputs.
    \item \textbf{CauScale.} Uses the released synthetic checkpoint \href{https://huggingface.co/OpenCausaLab/causcale-model/blob/main/synthetic/auprc\%3D0.905_migrated.ckpt}{\texttt{synthetic/auprc=0.905\_migrated.ckpt}}, feature sample size 500 by default, and fixed threshold 0.5 for probabilistic outputs.
    \item \textbf{CDFM.} Uses the released CDFM checkpoint \href{https://huggingface.co/DMIRLAB/CDFM/blob/main/model.safetensors}{\texttt{model.safetensors}} and its official output logic.
    \item \textbf{FoundCause.} Uses the released checkpoint \href{https://github.com/amazon-science/foundcause/releases/latest/download/checkpoint.pt}{\texttt{checkpoint.pt}}. The wrapper scores saved edge probabilities with a fixed $0.5$ threshold. The released official binary decision was also scored; we report the $0.5$-threshold output because it had higher average F1/SHD on the main, sample-size, and semantic/formula checks. This is a documented output rule, not a per-dataset hyperparameter search.
    \item \textbf{SEA.} Uses the released obs-only FCI checkpoint \href{https://github.com/rmwu/sea-reproduce/tree/main/checkpoints/fci_synthetic}{\texttt{fci\_synthetic/model\_best\_epoch=373\_auprc=0.842.ckpt}}, with 500 inference batches and FCI batch size 500. We do not run SEA as an interventional method: its interventional recipe uses GIES on sampled batches and is reported to need at least 250 observations per batch~\cite{wu2025sampleestimateaggregaterecipe}, while the main obs+int split has only $n_{\mathrm{int}}=200$ interventional rows in total, further split across several targets. The repaired real-data wrapper drops constant columns internally and maps predictions back to the original variable dimension.
    \item \textbf{TabCausal.} Uses the CausalArena paper-snapshot checkpoint \href{https://huggingface.co/LAMDA-Tabular/TabCausal/blob/main/checkpoints/tabcausal-v2.pt}{\texttt{tabcausal-v2.pt}} through the released inference wrapper and fixed threshold 0.5 for probabilistic adjacency outputs.
\end{itemize}

%% file: complete_results.tex
\section{Complete Results}
\label{sec:appendix-complete-results}

This section reports the numerical result views behind the compact main-text tables and figures. Overall benchmark tables include F1, precision, recall, SHD, SID, normalized SHD (nSHD), AUROC, and AP when the corresponding method output supports the metric; hard-graph methods without edge scores therefore have missing AUROC/AP entries. nSHD is computed as $\mathrm{SHD}/(d(d-1))$. Values are reported as mean$\pm$standard deviation across heterogeneous graph instances or datasets. The reported standard deviation is a descriptive spread and can exceed the mean for nonnegative metrics such as SHD. Bold and underline mark the best and second-best values within each comparable block, separately for every metric. To keep the appendix readable, factor-level decompositions use compact F1 views; the released CSV reports retain the full precision/recall/SHD/SID columns for each slice.

\begin{figure}[!htbp]
    \centering
    \includegraphics[width=0.98\textwidth]{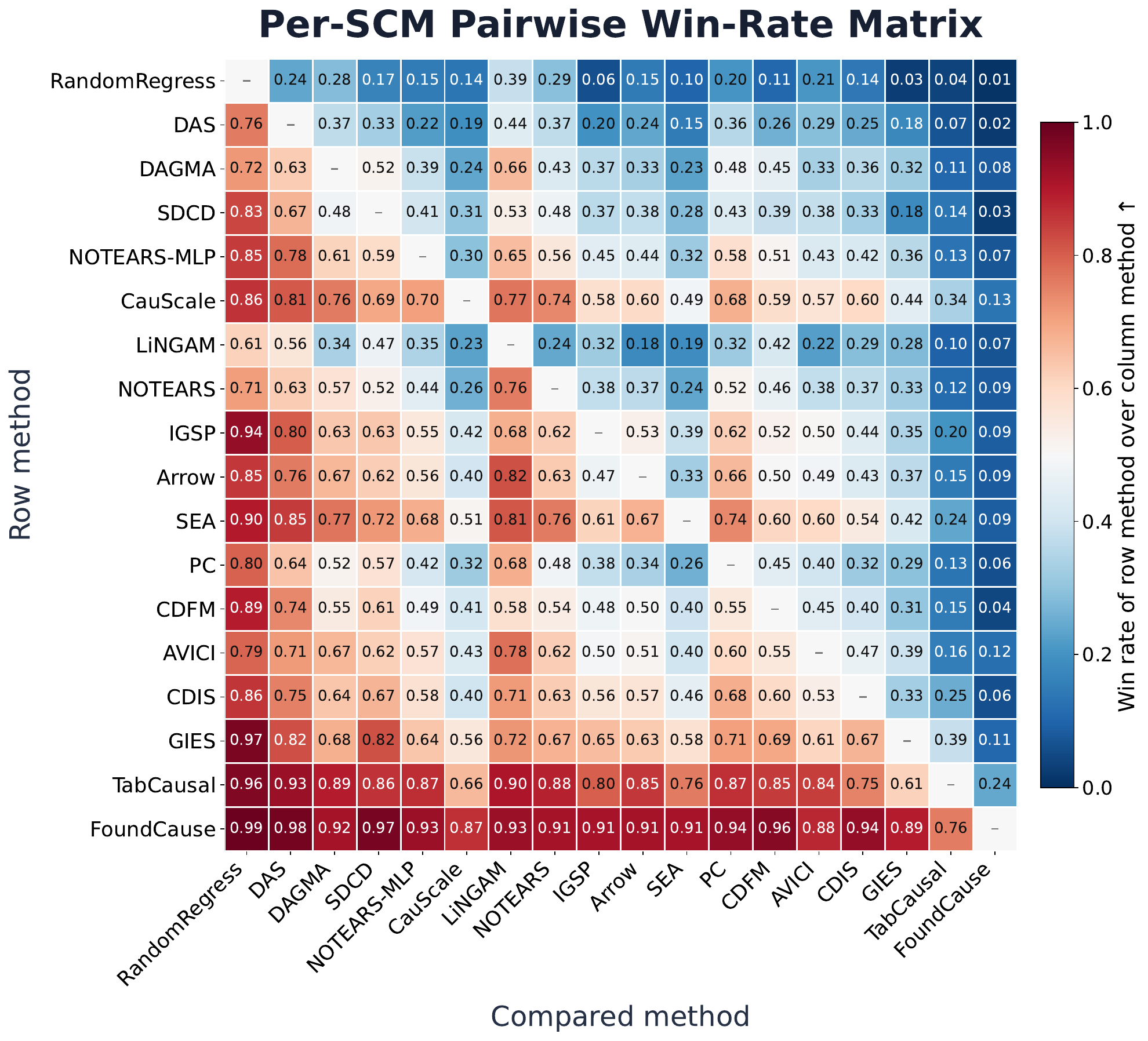}
    \caption{\textbf{Complete observation-only pairwise win-rate matrix.} Each cell reports how often the row method wins over the column method across per-SCM or per-dataset comparisons, after averaging repeated runs within the same unit. On each unit, F1 and SHD are compared separately (higher F1 wins, lower SHD wins, ties count as $0.5$), and wins from both metrics are pooled into one rate. Methods are ordered by mean family-wise win rate, matching \cref{fig:intro-pairwise-win-profile}. Red cells indicate that the row method wins more often; blue cells indicate the opposite.}
    \label{fig:appendix-pairwise-win-heatmap}
\end{figure}

\subsection{Synthetic Benchmark}
\label{sec:appendix-complete-synthetic}
\Cref{tab:appendix-synthetic-overall-complete} reports the overall synthetic benchmark metrics. \Cref{tab:appendix-synthetic-graph-family-compact,tab:appendix-synthetic-dimension-compact,tab:appendix-synthetic-root-dependency-compact,tab:appendix-synthetic-difficulty-component-compact} decompose the same benchmark by graph family, dimension, root-dependency mode, and difficulty component. \Cref{tab:appendix-synthetic-mechanism-edge-compact,tab:appendix-synthetic-noise-edge-compact} give compact edge-level F1 views by mechanism and noise labels.
\input{tables/generated_results/appendix_complete/synthetic_overall_complete_metrics}
\FloatBarrier
\input{tables/generated_results/appendix_complete/synthetic_graph_family_f1_compact}
\input{tables/generated_results/appendix_complete/synthetic_dimension_f1_compact}
\input{tables/generated_results/appendix_complete/synthetic_root_dependency_f1_compact}
\input{tables/generated_results/appendix_complete/synthetic_difficulty_component_f1_compact}
\input{tables/generated_results/appendix_complete/synthetic_mechanism_edge_f1_compact}
\input{tables/generated_results/appendix_complete/synthetic_noise_edge_f1_compact}

\subsection{Semantic Benchmark}
\label{sec:appendix-complete-semantic}
\Cref{tab:appendix-semantic-overall-complete} reports the full Semantic benchmark metrics. Domain-level behavior is analyzed in \cref{sec:semantic-formula-behavior-analysis}.
\input{tables/generated_results/appendix_complete/semantic_overall_complete_metrics}

\subsection{Formula Benchmark}
\label{sec:appendix-complete-formula}
\Cref{tab:appendix-formula-overall-complete} reports the full Formula benchmark metrics. Scientific-domain behavior is analyzed in \cref{sec:semantic-formula-behavior-analysis}.
\input{tables/generated_results/appendix_complete/formula_overall_complete_metrics}

\subsection{Real-data Benchmark}
\label{sec:appendix-complete-real}
\Cref{tab:appendix-real-method-complete} reports the complete real-data method summary. \Cref{tab:appendix-real-dataset-complete} lists every method--dataset pair, including observation-only and observation-plus-intervention splits. The six observation-only sources are the CD-CSG datasets Abalone, Auto MPG, Cardiac Arrhythmia, Concrete Compressive Strength, Deutscher Wetterdienst, and Ozone. The seven interventional sources are Sachs flow-cytometry; PetShop high traffic, low traffic, temporal traffic 1, and temporal traffic 2; and the Causal Chambers Light Tunnel and Wind Tunnel experiments.
\input{tables/generated_results/appendix_complete/real_method_complete_metrics}
\input{tables/generated_results/appendix_complete/real_dataset_complete_metrics}

\subsection{Sample-size Scaling}
\label{sec:appendix-complete-sample}
\Cref{tab:appendix-sample-scaling-f1-shd} reports the numerical F1/SHD values for the observation-only sample-size scaling suite, which fixes $d=30$ and varies the number of samples.
\input{tables/generated_results/appendix_complete/sample_scaling_f1_shd}

\subsection{Intervention-protocol Sensitivity}
\label{sec:appendix-complete-intervention}
\Cref{tab:appendix-intervention-protocol-complete} reports the full numerical values for the intervention-protocol sensitivity suite.
\input{tables/generated_results/appendix_complete/intervention_protocol_complete_metrics}

\subsection{Construction Ablation}
\label{sec:appendix-complete-ablation}
\Cref{tab:construction-ablation-quality} repeats the criterion-level construction-quality scores with standard deviations for the same 20 sampled scenarios summarized in \cref{fig:construction-ablation-stages}.
\input{tables/generated_results/ablation/construction_ablation_quality}

%% file: tables/generated_results/appendix_complete/synthetic_overall_complete_metrics.tex
\begingroup
\scriptsize
\setlength{\tabcolsep}{0.8pt}
\renewcommand{\arraystretch}{1.05}
\begin{longtable}{llr@{\hspace{0.8em}}llllllll}
\caption{\textbf{Complete synthetic benchmark metrics.} Values are mean$\pm$standard deviation over replicate-level rows.}\label{tab:appendix-synthetic-overall-complete}\\
\toprule
\textbf{Split} & \textbf{Method} & \textbf{N} & \textbf{F1} & \textbf{Prec.} & \textbf{Rec.} & \textbf{SHD} & \textbf{nSHD} & \textbf{SID} & \textbf{AUROC} & \textbf{AP} \\
\midrule
\endfirsthead
\toprule
\textbf{Split} & \textbf{Method} & \textbf{N} & \textbf{F1} & \textbf{Prec.} & \textbf{Rec.} & \textbf{SHD} & \textbf{nSHD} & \textbf{SID} & \textbf{AUROC} & \textbf{AP} \\
\midrule
\endhead
\midrule
\multicolumn{11}{r}{Continued on next page}\\
\endfoot
\bottomrule
\endlastfoot
Obs. & RandomRegress & 5k & $0.28{\pm}0.10$ & $0.24{\pm}0.11$ & $0.36{\pm}0.11$ & $218.0{\pm}310.0$ & $0.103{\pm}0.054$ & $1003.6{\pm}1488.4$ & $0.50{\pm}0.06$ & $0.13{\pm}0.07$ \\
Obs. & CDIS & 5k & $0.38{\pm}0.16$ & $0.47{\pm}0.20$ & $0.34{\pm}0.16$ & $143.0{\pm}352.0$ & $0.064{\pm}0.045$ & $826.6{\pm}1357.4$ & -- & -- \\
Obs. & GIES & 5k & $0.48{\pm}0.15$ & $0.46{\pm}0.17$ & \underline{$0.52{\pm}0.14$} & $123.6{\pm}174.9$ & $0.063{\pm}0.039$ & $798.1{\pm}1141.9$ & -- & -- \\
Obs. & IGSP & 5k & $0.41{\pm}0.15$ & $0.43{\pm}0.18$ & $0.40{\pm}0.14$ & $135.6{\pm}192.3$ & $0.063{\pm}0.034$ & $783.3{\pm}1239.3$ & -- & -- \\
Obs. & PC & 5k & $0.31{\pm}0.14$ & $0.35{\pm}0.15$ & $0.29{\pm}0.15$ & $115.4{\pm}149.0$ & $0.062{\pm}0.034$ & $990.2{\pm}1596.1$ & -- & -- \\
Obs. & DAS & 5k & $0.34{\pm}0.12$ & $0.33{\pm}0.13$ & $0.36{\pm}0.14$ & $142.6{\pm}180.1$ & $0.077{\pm}0.039$ & $722.1{\pm}1128.5$ & -- & -- \\
Obs. & LiNGAM & 5k & $0.21{\pm}0.14$ & $0.53{\pm}0.23$ & $0.14{\pm}0.11$ & $109.6{\pm}142.2$ & $0.060{\pm}0.029$ & $664.6{\pm}1139.3$ & -- & -- \\
Obs. & DAGMA & 5k & $0.28{\pm}0.14$ & $0.45{\pm}0.17$ & $0.22{\pm}0.13$ & $107.2{\pm}138.6$ & $0.059{\pm}0.031$ & $697.4{\pm}1136.9$ & -- & -- \\
\end{longtable}
\endgroup

%% file: tables/generated_results/appendix_complete/synthetic_graph_family_f1_compact.tex
\begin{table}[H]
\centering
\scriptsize
\setlength{\tabcolsep}{2pt}
\renewcommand{\arraystretch}{1.06}
\caption{\textbf{Compact synthetic graph-family breakdown.} Cells report Obs./Obs+int F1.}
\label{tab:appendix-synthetic-graph-family-compact}
\begin{tabularx}{\linewidth}{@{}l*{10}{>{\centering\arraybackslash}X}@{}}
\toprule
\textbf{Method} & \textbf{Erdos--Renyi} & \textbf{Scale-free} & \textbf{Chain} & \textbf{Tree} & \textbf{Layered} & \textbf{Collider-rich} & \textbf{Bipartite layered} & \textbf{Dense local modules} & \textbf{Small-world} & \textbf{Block-sparse layered} \\
\midrule
RandomRegress & $0.27$/-- & $0.27$/-- & $0.30$/-- & $0.27$/-- & $0.27$/-- & $0.27$/-- & $0.27$/-- & $0.31$/-- & $0.30$/-- & $0.30$/-- \\
CDIS & $0.42$/$0.42$ & $0.30$/$0.32$ & $0.33$/$0.36$ & $0.31$/$0.35$ & $0.40$/$0.42$ & $0.41$/$0.42$ & $0.58$/{\bfseries\boldmath $0.58$} & $0.34$/$0.39$ & $0.34$/$0.36$ & $0.34$/$0.38$ \\
GIES & \underline{$0.50$}/{\bfseries\boldmath $0.48$} & $0.42$/\underline{$0.46$} & $0.45$/\underline{$0.50$} & $0.44$/$0.48$ & \underline{$0.52$}/{\bfseries\boldmath $0.50$} & \underline{$0.49$}/\underline{$0.47$} & \underline{$0.61$}/\underline{$0.51$} & $0.45$/$0.46$ & $0.44$/{\bfseries\boldmath $0.48$} & $0.47$/$0.47$ \\
IGSP & $0.37$/$0.36$ & $0.41$/$0.39$ & $0.38$/$0.37$ & $0.43$/$0.40$ & $0.39$/$0.37$ & $0.38$/$0.36$ & $0.46$/$0.44$ & $0.44$/$0.41$ & $0.36$/$0.35$ & $0.45$/$0.43$ \\
PC & $0.28$/-- & $0.25$/-- & $0.34$/-- & $0.30$/-- & $0.32$/-- & $0.28$/-- & $0.29$/-- & $0.35$/-- & $0.31$/-- & $0.34$/-- \\
DAS & $0.34$/-- & $0.39$/-- & $0.31$/-- & $0.38$/-- & $0.33$/-- & $0.33$/-- & $0.30$/-- & $0.37$/-- & $0.30$/-- & $0.35$/-- \\
LiNGAM & $0.18$/-- & $0.21$/-- & $0.22$/-- & $0.22$/-- & $0.21$/-- & $0.19$/-- & $0.18$/-- & $0.23$/-- & $0.19$/-- & $0.23$/-- \\
\bottomrule
\end{tabularx}
\end{table}

%% file: tables/generated_results/appendix_complete/synthetic_dimension_f1_compact.tex
\begin{table}[H]
\centering
\scriptsize
\setlength{\tabcolsep}{2pt}
\renewcommand{\arraystretch}{1.06}
\caption{\textbf{Compact synthetic dimension breakdown.} Cells report Obs./Obs+int F1.}
\label{tab:appendix-synthetic-dimension-compact}
\begin{tabularx}{\linewidth}{@{}l*{5}{>{\centering\arraybackslash}X}@{}}
\toprule
\textbf{Method} & \textbf{10} & \textbf{20} & \textbf{30} & \textbf{50} & \textbf{100} \\
\midrule
RandomRegress & $0.34$/-- & $0.31$/-- & $0.28$/-- & $0.25$/-- & $0.23$/-- \\
CDIS & $0.36$/$0.44$ & $0.39$/$0.42$ & $0.39$/$0.40$ & $0.38$/$0.38$ & $0.36$/$0.35$ \\
GIES & $0.49$/\underline{$0.55$} & $0.49$/\underline{$0.49$} & $0.49$/\underline{$0.48$} & \underline{$0.47$}/$0.45$ & $0.47$/\underline{$0.44$} \\
IGSP & $0.49$/$0.47$ & $0.45$/$0.42$ & $0.42$/$0.40$ & $0.36$/$0.34$ & $0.32$/$0.30$ \\
PC & $0.42$/-- & $0.34$/-- & $0.31$/-- & $0.26$/-- & $0.20$/-- \\
DAS & $0.43$/-- & $0.38$/-- & $0.34$/-- & $0.30$/-- & $0.25$/-- \\
LiNGAM & $0.29$/-- & $0.23$/-- & $0.21$/-- & $0.18$/-- & $0.13$/-- \\
DAGMA & $0.34$/-- & $0.30$/-- & $0.29$/-- & $0.27$/-- & $0.22$/-- \\
NOTEARS & $0.33$/-- & $0.30$/-- & $0.28$/-- & $0.26$/-- & $0.21$/-- \\
NOTEARS-MLP & $0.39$/-- & $0.38$/-- & $0.36$/-- & $0.34$/-- & $0.29$/-- \\
SDCD & $0.38$/$0.40$ & $0.39$/$0.39$ & $0.39$/$0.38$ & $0.38$/$0.36$ & $0.37$/$0.36$ \\
Arrow & $0.45$/-- & $0.40$/-- & $0.36$/-- & $0.29$/-- & $0.21$/-- \\
AVICI & $0.41$/$0.46$ & $0.40$/$0.43$ & $0.36$/$0.38$ & $0.28$/$0.29$ & $0.16$/$0.16$ \\
CauScale & $0.35$/$0.43$ & $0.39$/$0.44$ & $0.41$/$0.46$ & $0.44$/\underline{$0.47$} & \underline{$0.48$}/{\bfseries\boldmath $0.49$} \\
CDFM & $0.54$/-- & $0.49$/-- & $0.44$/-- & $0.37$/-- & $0.27$/-- \\
FoundCause & {\bfseries\boldmath $0.67$}/-- & {\bfseries\boldmath $0.65$}/-- & {\bfseries\boldmath $0.63$}/-- & {\bfseries\boldmath $0.61$}/-- & {\bfseries\boldmath $0.56$}/-- \\
SEA & $0.46$/-- & $0.44$/-- & $0.42$/-- & $0.39$/-- & $0.27$/-- \\
TabCausal & \underline{$0.56$}/{\bfseries\boldmath $0.58$} & \underline{$0.56$}/{\bfseries\boldmath $0.56$} & \underline{$0.53$}/{\bfseries\boldmath $0.54$} & $0.47$/{\bfseries\boldmath $0.47$} & $0.36$/$0.35$ \\
\bottomrule
\end{tabularx}
\end{table}

%% file: tables/generated_results/appendix_complete/synthetic_root_dependency_f1_compact.tex
\begin{table}[H]
\centering
\scriptsize
\setlength{\tabcolsep}{2pt}
\renewcommand{\arraystretch}{1.06}
\caption{\textbf{Compact synthetic root-dependency breakdown.} Cells report Obs./Obs+int F1.}
\label{tab:appendix-synthetic-root-dependency-compact}
\begin{tabularx}{\linewidth}{@{}l*{4}{>{\centering\arraybackslash}X}@{}}
\toprule
\textbf{Method} & \textbf{correlated normal} & \textbf{independent roots} & \textbf{random covariance normal} & \textbf{uniform ball} \\
\midrule
RandomRegress & $0.23$/-- & $0.29$/-- & $0.22$/-- & $0.30$/-- \\
CDIS & $0.27$/$0.29$ & $0.38$/$0.41$ & $0.30$/$0.30$ & $0.45$/$0.47$ \\
GIES & $0.37$/$0.40$ & $0.49$/\underline{$0.49$} & $0.39$/$0.40$ & \underline{$0.54$}/{\bfseries\boldmath $0.51$} \\
IGSP & $0.35$/$0.34$ & $0.41$/$0.39$ & $0.37$/$0.35$ & $0.43$/$0.40$ \\
PC & $0.25$/-- & $0.31$/-- & $0.26$/-- & $0.32$/-- \\
DAS & $0.33$/-- & $0.34$/-- & $0.33$/-- & $0.32$/-- \\
LiNGAM & $0.21$/-- & $0.21$/-- & $0.19$/-- & $0.20$/-- \\
DAGMA & $0.32$/-- & $0.28$/-- & $0.31$/-- & $0.24$/-- \\
NOTEARS & $0.30$/-- & $0.28$/-- & $0.29$/-- & $0.24$/-- \\
NOTEARS-MLP & $0.38$/-- & $0.35$/-- & $0.37$/-- & $0.32$/-- \\
SDCD & $0.36$/$0.35$ & $0.39$/$0.38$ & $0.36$/$0.35$ & $0.38$/$0.38$ \\
Arrow & $0.32$/-- & $0.35$/-- & $0.30$/-- & $0.34$/-- \\
AVICI & $0.32$/$0.34$ & $0.33$/$0.35$ & $0.28$/$0.29$ & $0.32$/$0.34$ \\
CauScale & $0.43$/{\bfseries\boldmath $0.46$} & $0.41$/$0.46$ & $0.42$/{\bfseries\boldmath $0.45$} & $0.42$/$0.46$ \\
CDFM & $0.38$/-- & $0.43$/-- & $0.34$/-- & $0.44$/-- \\
FoundCause & {\bfseries\boldmath $0.58$}/-- & {\bfseries\boldmath $0.63$}/-- & {\bfseries\boldmath $0.54$}/-- & {\bfseries\boldmath $0.66$}/-- \\
SEA & $0.36$/-- & $0.40$/-- & $0.36$/-- & $0.40$/-- \\
TabCausal & \underline{$0.43$}/\underline{$0.46$} & \underline{$0.51$}/{\bfseries\boldmath $0.51$} & \underline{$0.42$}/\underline{$0.43$} & $0.51$/\underline{$0.50$} \\
\bottomrule
\end{tabularx}
\end{table}

%% file: tables/generated_results/appendix_complete/synthetic_difficulty_component_f1_compact.tex
\begin{table}[H]
\centering
\scriptsize
\setlength{\tabcolsep}{2pt}
\renewcommand{\arraystretch}{1.06}
\caption{\textbf{Compact synthetic difficulty-component breakdown.} Cells report Obs./Obs+int F1; multi-component cases contribute to each active component.}
\label{tab:appendix-synthetic-difficulty-component-compact}
\begin{tabularx}{\linewidth}{@{}l*{9}{>{\centering\arraybackslash}X}@{}}
\toprule
\textbf{Method} & \textbf{Baseline} & \textbf{Distribution shift} & \textbf{Feature corruption} & \textbf{Feature warping} & \textbf{High noise} & \textbf{Missingness} & \textbf{Outliers} & \textbf{Soft interventions} & \textbf{Weak signal} \\
\midrule
RandomRegress & $0.28$/-- & $0.28$/-- & $0.27$/-- & $0.27$/-- & $0.30$/-- & $0.27$/-- & $0.26$/-- & $0.28$/-- & $0.30$/-- \\
CDIS & $0.40$/$0.42$ & $0.39$/$0.43$ & $0.35$/$0.37$ & $0.37$/$0.40$ & $0.39$/$0.40$ & $0.36$/$0.38$ & $0.31$/$0.32$ & $0.38$/$0.41$ & $0.38$/$0.41$ \\
GIES & $0.51$/\underline{$0.52$} & $0.50$/$0.46$ & $0.45$/\underline{$0.46$} & \underline{$0.46$}/{\bfseries\boldmath $0.48$} & \underline{$0.52$}/{\bfseries\boldmath $0.53$} & $0.45$/\underline{$0.46$} & $0.36$/$0.37$ & $0.49$/\underline{$0.45$} & \underline{$0.50$}/{\bfseries\boldmath $0.51$} \\
IGSP & $0.42$/$0.41$ & $0.43$/$0.41$ & $0.38$/$0.35$ & $0.39$/$0.38$ & $0.46$/$0.43$ & $0.39$/$0.37$ & $0.30$/$0.29$ & $0.41$/$0.40$ & $0.43$/$0.41$ \\
PC & $0.33$/-- & $0.32$/-- & $0.31$/-- & $0.30$/-- & $0.31$/-- & $0.31$/-- & $0.23$/-- & $0.31$/-- & $0.31$/-- \\
DAS & $0.34$/-- & $0.35$/-- & $0.32$/-- & $0.33$/-- & $0.41$/-- & $0.32$/-- & $0.30$/-- & $0.34$/-- & $0.35$/-- \\
LiNGAM & $0.26$/-- & $0.23$/-- & $0.18$/-- & $0.21$/-- & $0.16$/-- & $0.20$/-- & $0.06$/-- & $0.22$/-- & $0.22$/-- \\
\bottomrule
\end{tabularx}
\end{table}

%% file: tables/generated_results/appendix_complete/synthetic_mechanism_edge_f1_compact.tex
\begin{table}[H]
\centering
\scriptsize
\setlength{\tabcolsep}{2pt}
\renewcommand{\arraystretch}{1.06}
\caption{\textbf{Compact synthetic edge-level mechanism breakdown.} Cells report Obs./Obs+int edge-F1 from aggregated TP/FP/FN counts.}
\label{tab:appendix-synthetic-mechanism-edge-compact}
\begin{tabularx}{\linewidth}{@{}l*{10}{>{\centering\arraybackslash}X}@{}}
\toprule
\textbf{Method} & \textbf{activation additive} & \textbf{aggregation nonlin.} & \textbf{discrete assign.} & \textbf{exp log} & \textbf{linear additive} & \textbf{linear interact.} & \textbf{mixture regime} & \textbf{mult.} & \textbf{neural random} & \textbf{piecewise threshold} \\
\midrule
RandomRegress & $0.06$/-- & $0.10$/-- & $0.04$/-- & $0.08$/-- & $0.08$/-- & $0.08$/-- & $0.07$/-- & $0.04$/-- & $0.05$/-- & $0.08$/-- \\
CDIS & $0.34$/$0.34$ & $0.32$/$0.32$ & $0.32$/$0.32$ & $0.37$/$0.36$ & $0.39$/$0.39$ & $0.40$/$0.39$ & $0.38$/$0.38$ & $0.34$/$0.35$ & $0.37$/$0.36$ & $0.37$/$0.36$ \\
GIES & \underline{$0.48$}/{\bfseries\boldmath $0.47$} & $0.43$/$0.42$ & \underline{$0.45$}/{\bfseries\boldmath $0.44$} & \underline{$0.52$}/\underline{$0.51$} & \underline{$0.59$}/{\bfseries\boldmath $0.58$} & \underline{$0.55$}/\underline{$0.53$} & \underline{$0.56$}/\underline{$0.55$} & \underline{$0.47$}/\underline{$0.47$} & \underline{$0.51$}/{\bfseries\boldmath $0.50$} & \underline{$0.54$}/\underline{$0.54$} \\
IGSP & $0.34$/$0.32$ & $0.32$/$0.31$ & $0.31$/$0.29$ & $0.38$/$0.36$ & $0.42$/$0.40$ & $0.39$/$0.38$ & $0.40$/$0.39$ & $0.34$/$0.33$ & $0.36$/$0.34$ & $0.39$/$0.38$ \\
PC & $0.23$/-- & $0.25$/-- & $0.20$/-- & $0.27$/-- & $0.30$/-- & $0.29$/-- & $0.29$/-- & $0.21$/-- & $0.23$/-- & $0.28$/-- \\
DAS & $0.28$/-- & $0.30$/-- & $0.29$/-- & $0.30$/-- & $0.31$/-- & $0.29$/-- & $0.32$/-- & $0.31$/-- & $0.33$/-- & $0.29$/-- \\
LiNGAM & $0.07$/-- & $0.12$/-- & $0.04$/-- & $0.09$/-- & $0.13$/-- & $0.12$/-- & $0.11$/-- & $0.05$/-- & $0.07$/-- & $0.11$/-- \\
\bottomrule
\end{tabularx}
\vspace{0.45em}
\begin{tabularx}{\linewidth}{@{}l*{6}{>{\centering\arraybackslash}X}@{}}
\toprule
\textbf{Method} & \textbf{polynomial power} & \textbf{rational ratio} & \textbf{rff gp} & \textbf{saturation clip} & \textbf{tree based} & \textbf{trig periodic} \\
\midrule
RandomRegress & $0.07$/-- & $0.07$/-- & $0.05$/-- & $0.07$/-- & $0.03$/-- & $0.05$/-- \\
CDIS & $0.31$/$0.31$ & $0.32$/$0.33$ & $0.32$/$0.31$ & $0.37$/$0.36$ & $0.21$/$0.21$ & $0.33$/$0.33$ \\
GIES & $0.42$/$0.41$ & \underline{$0.47$}/{\bfseries\boldmath $0.46$} & $0.40$/$0.39$ & \underline{$0.54$}/{\bfseries\boldmath $0.53$} & $0.27$/$0.29$ & \underline{$0.45$}/\underline{$0.44$} \\
IGSP & $0.32$/$0.31$ & $0.33$/$0.32$ & $0.29$/$0.28$ & $0.39$/$0.37$ & $0.22$/$0.22$ & $0.33$/$0.31$ \\
PC & $0.23$/-- & $0.23$/-- & $0.19$/-- & $0.27$/-- & $0.15$/-- & $0.22$/-- \\
DAS & $0.25$/-- & $0.31$/-- & $0.30$/-- & $0.32$/-- & $0.20$/-- & $0.29$/-- \\
LiNGAM & $0.07$/-- & $0.10$/-- & $0.05$/-- & $0.10$/-- & $0.02$/-- & $0.06$/-- \\
\bottomrule
\end{tabularx}
\end{table}

%% file: tables/generated_results/appendix_complete/synthetic_noise_edge_f1_compact.tex
\begin{table}[H]
\centering
\scriptsize
\setlength{\tabcolsep}{2pt}
\renewcommand{\arraystretch}{1.06}
\caption{\textbf{Compact synthetic edge-level noise breakdown.} Cells report Obs./Obs+int edge-F1 from aggregated TP/FP/FN counts.}
\label{tab:appendix-synthetic-noise-edge-compact}
\begin{tabularx}{\linewidth}{@{}l*{10}{>{\centering\arraybackslash}X}@{}}
\toprule
\textbf{Method} & \textbf{beta} & \textbf{censored} & \textbf{exponential} & \textbf{gaussian} & \textbf{gumbel} & \textbf{heterosk.} & \textbf{laplace} & \textbf{mixture outlier} & \textbf{mult. lognormal} & \textbf{quantized} \\
\midrule
RandomRegress & $0.09$/-- & $0.06$/-- & $0.08$/-- & $0.06$/-- & $0.06$/-- & $0.08$/-- & $0.06$/-- & $0.05$/-- & $0.09$/-- & $0.06$/-- \\
CDIS & $0.39$/$0.40$ & $0.34$/$0.34$ & $0.35$/$0.35$ & $0.34$/$0.34$ & $0.34$/$0.33$ & $0.37$/$0.37$ & $0.33$/$0.33$ & $0.32$/$0.32$ & $0.37$/$0.37$ & $0.32$/$0.33$ \\
GIES & $0.51$/$0.50$ & \underline{$0.49$}/{\bfseries\boldmath $0.48$} & $0.49$/$0.48$ & \underline{$0.48$}/{\bfseries\boldmath $0.47$} & \underline{$0.48$}/{\bfseries\boldmath $0.47$} & \underline{$0.51$}/\underline{$0.50$} & \underline{$0.48$}/\underline{$0.47$} & $0.46$/\underline{$0.45$} & $0.51$/\underline{$0.50$} & \underline{$0.48$}/{\bfseries\boldmath $0.47$} \\
IGSP & $0.39$/$0.39$ & $0.34$/$0.32$ & $0.38$/$0.37$ & $0.35$/$0.33$ & $0.34$/$0.32$ & $0.37$/$0.35$ & $0.34$/$0.32$ & $0.32$/$0.30$ & $0.38$/$0.38$ & $0.34$/$0.32$ \\
PC & $0.28$/-- & $0.23$/-- & $0.27$/-- & $0.24$/-- & $0.24$/-- & $0.26$/-- & $0.23$/-- & $0.22$/-- & $0.27$/-- & $0.23$/-- \\
DAS & $0.28$/-- & $0.31$/-- & $0.29$/-- & $0.31$/-- & $0.30$/-- & $0.32$/-- & $0.29$/-- & $0.22$/-- & $0.29$/-- & $0.31$/-- \\
LiNGAM & $0.11$/-- & $0.06$/-- & $0.11$/-- & $0.07$/-- & $0.08$/-- & $0.09$/-- & $0.08$/-- & $0.09$/-- & $0.11$/-- & $0.07$/-- \\
\bottomrule
\end{tabularx}
\vspace{0.45em}
\begin{tabularx}{\linewidth}{@{}l*{4}{>{\centering\arraybackslash}X}@{}}
\toprule
\textbf{Method} & \textbf{RFF heterosk.} & \textbf{student t} & \textbf{target r2 gaussian} & \textbf{uniform} \\
\midrule
RandomRegress & $0.02$/-- & $0.06$/-- & $0.06$/-- & $0.08$/-- \\
CDIS & $0.25$/$0.25$ & $0.34$/$0.35$ & $0.34$/$0.33$ & $0.35$/$0.36$ \\
GIES & \underline{$0.37$}/{\bfseries\boldmath $0.36$} & \underline{$0.48$}/\underline{$0.48$} & \underline{$0.48$}/{\bfseries\boldmath $0.46$} & $0.50$/$0.49$ \\
IGSP & $0.22$/$0.20$ & $0.35$/$0.33$ & $0.36$/$0.33$ & $0.37$/$0.36$ \\
PC & $0.14$/-- & $0.24$/-- & $0.24$/-- & $0.26$/-- \\
DAS & $0.21$/-- & $0.30$/-- & $0.34$/-- & $0.33$/-- \\
LiNGAM & $0.03$/-- & $0.08$/-- & $0.07$/-- & $0.10$/-- \\
DAGMA & $0.11$/-- & $0.24$/-- & $0.26$/-- & $0.28$/-- \\
NOTEARS & $0.02$/-- & $0.07$/-- & $0.06$/-- & $0.08$/-- \\
NOTEARS-MLP & $0.13$/-- & $0.26$/-- & $0.28$/-- & $0.30$/-- \\
SDCD & $0.23$/$0.21$ & $0.39$/$0.39$ & $0.36$/$0.34$ & $0.44$/$0.44$ \\
Arrow & $0.11$/-- & $0.24$/-- & $0.24$/-- & $0.29$/-- \\
AVICI & $0.08$/$0.08$ & $0.23$/$0.23$ & $0.17$/$0.17$ & $0.25$/$0.27$ \\
CauScale & $0.25$/\underline{$0.26$} & $0.47$/{\bfseries\boldmath $0.50$} & $0.42$/\underline{$0.43$} & $0.49$/{\bfseries\boldmath $0.51$} \\
CDFM & $0.17$/-- & $0.22$/-- & $0.20$/-- & $0.23$/-- \\
FoundCause & {\bfseries\boldmath $0.40$}/-- & {\bfseries\boldmath $0.58$}/-- & {\bfseries\boldmath $0.54$}/-- & {\bfseries\boldmath $0.62$}/-- \\
SEA & $0.16$/-- & $0.34$/-- & $0.32$/-- & $0.38$/-- \\
TabCausal & $0.24$/$0.23$ & $0.42$/$0.42$ & $0.40$/$0.37$ & \underline{$0.51$}/\underline{$0.50$} \\
\bottomrule
\end{tabularx}
\end{table}

%% file: tables/generated_results/appendix_complete/semantic_overall_complete_metrics.tex
\begingroup
\scriptsize
\setlength{\tabcolsep}{0.8pt}
\renewcommand{\arraystretch}{1.05}
\begin{longtable}{llr@{\hspace{0.8em}}llllllll}
\caption{\textbf{Complete Semantic benchmark metrics.} Values are mean$\pm$standard deviation over replicate-level rows.}\label{tab:appendix-semantic-overall-complete}\\
\toprule
\textbf{Split} & \textbf{Method} & \textbf{N} & \textbf{F1} & \textbf{Prec.} & \textbf{Rec.} & \textbf{SHD} & \textbf{nSHD} & \textbf{SID} & \textbf{AUROC} & \textbf{AP} \\
\midrule
\endfirsthead
\toprule
\textbf{Split} & \textbf{Method} & \textbf{N} & \textbf{F1} & \textbf{Prec.} & \textbf{Rec.} & \textbf{SHD} & \textbf{nSHD} & \textbf{SID} & \textbf{AUROC} & \textbf{AP} \\
\midrule
\endhead
\midrule
\multicolumn{11}{r}{Continued on next page}\\
\endfoot
\bottomrule
\endlastfoot
Obs. & RandomRegress & 500 & $0.25{\pm}0.09$ & $0.20{\pm}0.09$ & $0.38{\pm}0.10$ & $93.2{\pm}69.3$ & $0.183{\pm}0.124$ & $192.7{\pm}58.2$ & $0.55{\pm}0.05$ & $0.14{\pm}0.05$ \\
Obs. & CDIS & 500 & $0.40{\pm}0.17$ & \underline{$0.55{\pm}0.23$} & $0.32{\pm}0.15$ & \underline{$31.7{\pm}9.2$} & \underline{$0.063{\pm}0.016$} & $123.5{\pm}31.2$ & -- & -- \\
Obs. & GIES & 500 & \underline{$0.55{\pm}0.14$} & $0.54{\pm}0.16$ & \underline{$0.57{\pm}0.14$} & $33.4{\pm}15.7$ & $0.066{\pm}0.028$ & \underline{$107.9{\pm}53.4$} & -- & -- \\
Obs. & IGSP & 500 & $0.31{\pm}0.18$ & $0.34{\pm}0.22$ & $0.30{\pm}0.16$ & $40.8{\pm}14.4$ & $0.081{\pm}0.025$ & $148.0{\pm}48.5$ & -- & -- \\
Obs. & PC & 500 & $0.33{\pm}0.16$ & $0.43{\pm}0.21$ & $0.27{\pm}0.14$ & $33.5{\pm}9.7$ & $0.067{\pm}0.017$ & $139.5{\pm}36.6$ & -- & -- \\
Obs. & DAS & 500 & $0.21{\pm}0.07$ & $0.19{\pm}0.07$ & $0.24{\pm}0.08$ & $59.5{\pm}15.9$ & $0.119{\pm}0.026$ & $185.4{\pm}42.9$ & -- & -- \\
Obs. & LiNGAM & 500 & $0.20{\pm}0.09$ & $0.40{\pm}0.18$ & $0.14{\pm}0.06$ & $38.9{\pm}9.5$ & $0.078{\pm}0.017$ & $137.2{\pm}31.7$ & -- & -- \\
Obs. & DAGMA & 500 & $0.18{\pm}0.07$ & $0.25{\pm}0.10$ & $0.15{\pm}0.06$ & $42.0{\pm}9.2$ & $0.084{\pm}0.014$ & $166.8{\pm}43.1$ & -- & -- \\
\end{longtable}
\endgroup

%% file: tables/generated_results/appendix_complete/formula_overall_complete_metrics.tex
\begingroup
\scriptsize
\setlength{\tabcolsep}{0.8pt}
\renewcommand{\arraystretch}{1.05}
\begin{longtable}{llr@{\hspace{0.8em}}llllllll}
\caption{\textbf{Complete Formula benchmark metrics.} Values are mean$\pm$standard deviation over replicate-level rows.}\label{tab:appendix-formula-overall-complete}\\
\toprule
\textbf{Split} & \textbf{Method} & \textbf{N} & \textbf{F1} & \textbf{Prec.} & \textbf{Rec.} & \textbf{SHD} & \textbf{nSHD} & \textbf{SID} & \textbf{AUROC} & \textbf{AP} \\
\midrule
\endfirsthead
\toprule
\textbf{Split} & \textbf{Method} & \textbf{N} & \textbf{F1} & \textbf{Prec.} & \textbf{Rec.} & \textbf{SHD} & \textbf{nSHD} & \textbf{SID} & \textbf{AUROC} & \textbf{AP} \\
\midrule
\endhead
\midrule
\multicolumn{11}{r}{Continued on next page}\\
\endfoot
\bottomrule
\endlastfoot
Obs. & RandomRegress & 500 & $0.20{\pm}0.08$ & $0.14{\pm}0.07$ & $0.38{\pm}0.12$ & $108.3{\pm}70.1$ & $0.221{\pm}0.134$ & $187.2{\pm}60.8$ & $0.56{\pm}0.05$ & $0.12{\pm}0.05$ \\
Obs. & CDIS & 500 & $0.23{\pm}0.19$ & $0.35{\pm}0.28$ & $0.18{\pm}0.16$ & \underline{$31.0{\pm}6.9$} & \underline{$0.064{\pm}0.015$} & $106.8{\pm}33.1$ & -- & -- \\
Obs. & GIES & 500 & $0.47{\pm}0.14$ & $0.42{\pm}0.15$ & \underline{$0.54{\pm}0.15$} & $36.7{\pm}14.8$ & $0.076{\pm}0.031$ & $95.0{\pm}51.6$ & -- & -- \\
Obs. & IGSP & 500 & $0.21{\pm}0.16$ & $0.20{\pm}0.17$ & $0.23{\pm}0.17$ & $43.3{\pm}17.3$ & $0.089{\pm}0.035$ & $137.9{\pm}49.0$ & -- & -- \\
Obs. & PC & 500 & $0.22{\pm}0.18$ & $0.29{\pm}0.23$ & $0.18{\pm}0.15$ & $31.1{\pm}7.0$ & $0.064{\pm}0.015$ & $111.9{\pm}32.9$ & -- & -- \\
Obs. & DAS & 500 & $0.16{\pm}0.07$ & $0.13{\pm}0.06$ & $0.21{\pm}0.08$ & $64.8{\pm}16.5$ & $0.133{\pm}0.034$ & $171.5{\pm}46.6$ & -- & -- \\
Obs. & LiNGAM & 500 & $0.24{\pm}0.10$ & $0.34{\pm}0.15$ & $0.19{\pm}0.08$ & $34.7{\pm}8.6$ & $0.071{\pm}0.017$ & $122.2{\pm}36.4$ & -- & -- \\
Obs. & DAGMA & 500 & $0.18{\pm}0.07$ & $0.20{\pm}0.09$ & $0.16{\pm}0.07$ & $39.7{\pm}8.7$ & $0.082{\pm}0.018$ & $163.5{\pm}50.4$ & -- & -- \\
\end{longtable}
\endgroup

%% file: tables/generated_results/appendix_complete/real_method_complete_metrics.tex
\begingroup
\scriptsize
\setlength{\tabcolsep}{0.8pt}
\renewcommand{\arraystretch}{1.05}
\begin{longtable}{llr@{\hspace{0.8em}}llllllll}
\caption{\textbf{Complete real-data method metrics.} Obs.-only includes the six native observation-only datasets plus observation-only conversions of the seven interventional datasets.}\label{tab:appendix-real-method-complete}\\
\toprule
\textbf{Split} & \textbf{Method} & \textbf{N} & \textbf{F1} & \textbf{Prec.} & \textbf{Rec.} & \textbf{SHD} & \textbf{nSHD} & \textbf{SID} & \textbf{AUROC} & \textbf{AP} \\
\midrule
\endfirsthead
\toprule
\textbf{Split} & \textbf{Method} & \textbf{N} & \textbf{F1} & \textbf{Prec.} & \textbf{Rec.} & \textbf{SHD} & \textbf{nSHD} & \textbf{SID} & \textbf{AUROC} & \textbf{AP} \\
\midrule
\endhead
\midrule
\multicolumn{11}{r}{Continued on next page}\\
\endfoot
\bottomrule
\endlastfoot
Obs. & RandomRegress & 13 & $0.20{\pm}0.14$ & $0.15{\pm}0.11$ & $0.31{\pm}0.21$ & $59.8{\pm}55.9$ & $0.208{\pm}0.148$ & $144.2{\pm}186.5$ & -- & -- \\
Obs. & CDIS & 13 & $0.28{\pm}0.22$ & $0.24{\pm}0.20$ & $0.40{\pm}0.23$ & $99.5{\pm}133.7$ & $0.204{\pm}0.118$ & $179.8{\pm}222.8$ & -- & -- \\
Obs. & GIES & 13 & $0.26{\pm}0.15$ & $0.21{\pm}0.15$ & $0.39{\pm}0.22$ & $48.0{\pm}43.5$ & $0.201{\pm}0.153$ & $164.5{\pm}210.2$ & -- & -- \\
Obs. & IGSP & 13 & $0.18{\pm}0.22$ & $0.15{\pm}0.21$ & $0.23{\pm}0.27$ & $31.1{\pm}23.5$ & $0.175{\pm}0.165$ & $65.9{\pm}71.8$ & -- & -- \\
Obs. & PC & 13 & $0.18{\pm}0.14$ & $0.23{\pm}0.17$ & $0.18{\pm}0.15$ & $28.8{\pm}21.0$ & $0.183{\pm}0.159$ & $63.8{\pm}55.2$ & -- & -- \\
Obs. & DAS & 13 & $0.19{\pm}0.23$ & $0.14{\pm}0.17$ & $0.30{\pm}0.36$ & $61.9{\pm}55.1$ & $0.207{\pm}0.141$ & $130.7{\pm}126.7$ & -- & -- \\
Obs. & LiNGAM & 13 & $0.13{\pm}0.17$ & $0.24{\pm}0.30$ & $0.09{\pm}0.12$ & {\bfseries\boldmath $24.3{\pm}17.8$} & $0.164{\pm}0.151$ & {\bfseries\boldmath $50.5{\pm}49.0$} & -- & -- \\
Obs. & DAGMA & 13 & $0.16{\pm}0.14$ & $0.17{\pm}0.14$ & $0.18{\pm}0.19$ & $31.8{\pm}27.1$ & $0.197{\pm}0.184$ & $62.4{\pm}54.5$ & -- & -- \\
\end{longtable}
\endgroup

%% file: tables/generated_results/appendix_complete/real_dataset_complete_metrics.tex
\begingroup
\scriptsize
\setlength{\tabcolsep}{2pt}
\renewcommand{\arraystretch}{1.05}
\begin{longtable}{p{0.16\textwidth}llrrrrrr}
\caption{\textbf{Complete real-data dataset-level metrics.} Each row is one method on one scored real dataset.}\label{tab:appendix-real-dataset-complete}\\
\toprule
\textbf{Dataset} & \textbf{Split} & \textbf{Method} & \textbf{F1} & \textbf{Precision} & \textbf{Recall} & \textbf{SHD} & \textbf{SID} & \textbf{nSHD} \\
\midrule
\endfirsthead
\toprule
\textbf{Dataset} & \textbf{Split} & \textbf{Method} & \textbf{F1} & \textbf{Precision} & \textbf{Recall} & \textbf{SHD} & \textbf{SID} & \textbf{nSHD} \\
\midrule
\endhead
\midrule
\multicolumn{9}{r}{Continued on next page}\\
\endfoot
\bottomrule
\endlastfoot
Abalone & Obs. & RandomRegress & $0.00$ & $0.00$ & $0.00$ & $24.0$ & $34.0$ & $0.429$ \\
Abalone & Obs. & CDIS & $0.27$ & $0.20$ & $0.43$ & $16.0$ & $26.0$ & $0.286$ \\
Abalone & Obs. & GIES & $0.36$ & $0.24$ & \underline{$0.71$} & $18.0$ & $20.0$ & $0.321$ \\
Abalone & Obs. & IGSP & $0.23$ & $0.16$ & $0.43$ & $19.0$ & $24.0$ & $0.339$ \\
Abalone & Obs. & PC & $0.09$ & $0.07$ & $0.14$ & $18.0$ & $27.0$ & $0.321$ \\
Abalone & Obs. & DAS & $0.00$ & $0.00$ & $0.00$ & $23.0$ & $30.0$ & $0.411$ \\
Abalone & Obs. & LiNGAM & $0.00$ & $0.00$ & $0.00$ & $11.0$ & $14.0$ & $0.196$ \\
Abalone & Obs. & DAGMA & $0.00$ & $0.00$ & $0.00$ & \underline{$8.0$} & \underline{$11.0$} & \underline{$0.143$} \\
\end{longtable}
\endgroup

%% file: tables/generated_results/appendix_complete/sample_scaling_f1_shd.tex
\begin{table}[!htbp]
\centering
\scriptsize
\setlength{\tabcolsep}{2.5pt}
\renewcommand{\arraystretch}{1.05}
\begin{threeparttable}
\caption{\textbf{Complete sample-size sensitivity results.} Cells report F1/SHD for the observation-only $d=30$ synthetic suite.}
\label{tab:appendix-sample-scaling-f1-shd}
\begin{tabular*}{\textwidth}{@{\extracolsep{\fill}}lccccccc@{}}
\toprule
\textbf{Method} & \textbf{$n=100$} & \textbf{$n=200$} & \textbf{$n=500$} & \textbf{$n=1000$} & \textbf{$n=2000$} & \textbf{$n=5000$} & \textbf{$n=10000$} \\
\midrule
RandomRegress & 0.25/76.6 & 0.28/76.8 & 0.28/84.1 & 0.29/88.8 & 0.28/99.2 & 0.27/113.0 & 0.26/125.7 \\
CDIS & 0.23/51.1 & 0.29/49.4 & 0.36/48.4 & 0.39/50.8 & 0.40/54.2 & 0.42/60.0 & 0.43/61.7 \\
GIES & \underline{0.35}/61.8 & 0.41/56.2 & 0.46/54.2 & 0.49/54.6 & 0.50/57.9 & 0.49/65.3 & 0.49/70.6 \\
IGSP & 0.25/52.3 & 0.31/51.8 & 0.38/52.2 & 0.41/54.3 & 0.43/58.9 & 0.43/68.8 & 0.43/76.8 \\
PC & 0.25/50.6 & 0.27/50.2 & 0.30/50.8 & 0.31/52.5 & 0.32/54.3 & 0.31/57.9 & 0.31/60.1 \\
DAS & 0.22/61.5 & 0.28/61.5 & 0.32/65.4 & 0.35/67.5 & 0.36/71.0 & 0.35/104.7 & 0.36/87.1 \\
LiNGAM & 0.20/51.3 & 0.20/50.5 & 0.21/49.7 & 0.22/49.5 & 0.21/49.4 & 0.21/49.5 & 0.21/49.5 \\
DAGMA & 0.27/53.0 & 0.29/50.1 & 0.29/49.6 & 0.29/49.1 & 0.29/48.8 & 0.29/49.0 & 0.29/49.0 \\
NOTEARS & 0.26/49.2 & 0.28/48.2 & 0.28/47.7 & 0.28/47.4 & 0.28/47.5 & 0.28/47.3 & 0.27/47.5 \\
NOTEARS-MLP & 0.26/78.2 & 0.32/57.9 & 0.35/51.5 & 0.36/49.9 & 0.36/49.9 & 0.37/49.4 & 0.36/49.6 \\
SDCD & 0.22/118.1 & 0.26/100.0 & 0.35/75.3 & 0.38/68.7 & 0.40/62.8 & 0.41/59.2 & 0.43/54.2 \\
Arrow & 0.32/\textbf{47.5} & 0.34/46.5 & 0.35/46.3 & 0.36/46.6 & 0.36/47.5 & 0.34/50.0 & 0.33/53.5 \\
AVICI & 0.22/58.0 & 0.31/47.5 & 0.35/44.9 & 0.35/44.5 & 0.35/44.3 & 0.36/43.7 & 0.36/40.4 \\
CauScale & 0.26/66.5 & 0.33/49.7 & 0.40/44.6 & 0.41/44.6 & 0.40/47.0 & 0.37/52.6 & 0.35/56.3 \\
CDFM & 0.34/70.5 & 0.38/65.5 & 0.42/61.2 & 0.44/58.0 & 0.46/54.5 & 0.48/52.4 & 0.48/51.2 \\
FoundCause & \textbf{0.43}/49.7 & \textbf{0.52}/\underline{45.2} & \textbf{0.61}/\textbf{39.0} & \textbf{0.64}/\textbf{35.6} & \textbf{0.65}/\textbf{34.7} & \textbf{0.65}/\textbf{34.1} & \textbf{0.66}/\textbf{33.7} \\
SEA & 0.26/80.8 & 0.34/58.6 & 0.41/45.7 & 0.42/45.0 & 0.42/45.1 & 0.42/44.7 & 0.42/44.8 \\
TabCausal & 0.33/\underline{47.6} & \underline{0.42}/\textbf{43.9} & \underline{0.49}/\underline{41.4} & \underline{0.53}/\underline{39.2} & \underline{0.55}/\underline{39.0} & \underline{0.57}/\underline{38.1} & \underline{0.57}/\underline{38.6} \\
\bottomrule
\end{tabular*}
\begin{tablenotes}[flushleft]
\footnotesize
\item The sample-size suite fixes $d=30$ and varies only the number of observational samples. Higher F1 and lower SHD are better; bold and underline mark the best and second-best F1 and SHD separately in each column.
\end{tablenotes}
\end{threeparttable}
\end{table}

%% file: tables/generated_results/appendix_complete/intervention_protocol_complete_metrics.tex
\begingroup
\scriptsize
\setlength{\tabcolsep}{0.8pt}
\renewcommand{\arraystretch}{1.05}
\begin{longtable}{llr@{\hspace{0.8em}}llllllll}
\caption{\textbf{Complete intervention-protocol sensitivity metrics.} All rows use the mixed observational/interventional split.}\label{tab:appendix-intervention-protocol-complete}\\
\toprule
\textbf{Protocol} & \textbf{Method} & \textbf{N} & \textbf{F1} & \textbf{Prec.} & \textbf{Rec.} & \textbf{SHD} & \textbf{nSHD} & \textbf{SID} & \textbf{AUROC} & \textbf{AP} \\
\midrule
\endfirsthead
\toprule
\textbf{Protocol} & \textbf{Method} & \textbf{N} & \textbf{F1} & \textbf{Prec.} & \textbf{Rec.} & \textbf{SHD} & \textbf{nSHD} & \textbf{SID} & \textbf{AUROC} & \textbf{AP} \\
\midrule
\endhead
\midrule
\multicolumn{11}{r}{Continued on next page}\\
\endfoot
\bottomrule
\endlastfoot
Default & CDIS & 500 & $0.40{\pm}0.12$ & $0.48{\pm}0.16$ & $0.35{\pm}0.10$ & $48.1{\pm}19.4$ & $0.055{\pm}0.022$ & $237.9{\pm}117.7$ & -- & -- \\
Default & GIES & 500 & \underline{$0.48{\pm}0.11$} & $0.43{\pm}0.13$ & {\bfseries\boldmath $0.56{\pm}0.11$} & $59.7{\pm}31.4$ & $0.069{\pm}0.036$ & $251.3{\pm}78.4$ & -- & -- \\
Default & IGSP & 500 & $0.40{\pm}0.11$ & $0.44{\pm}0.15$ & $0.38{\pm}0.11$ & $53.7{\pm}26.9$ & $0.062{\pm}0.031$ & $232.6{\pm}107.2$ & -- & -- \\
Default & SDCD & 500 & $0.38{\pm}0.09$ & $0.33{\pm}0.11$ & $0.48{\pm}0.11$ & $73.0{\pm}25.5$ & $0.084{\pm}0.029$ & $246.0{\pm}92.0$ & -- & -- \\
Default & AVICI & 500 & $0.39{\pm}0.13$ & {\bfseries\boldmath $0.66{\pm}0.15$} & $0.29{\pm}0.12$ & \underline{$43.4{\pm}17.3$} & \underline{$0.050{\pm}0.020$} & \underline{$189.6{\pm}109.6$} & $0.77{\pm}0.09$ & $0.42{\pm}0.12$ \\
Default & CauScale & 500 & $0.45{\pm}0.11$ & $0.56{\pm}0.12$ & $0.39{\pm}0.11$ & $43.6{\pm}18.7$ & $0.050{\pm}0.022$ & $201.3{\pm}108.4$ & {\bfseries\boldmath $0.86{\pm}0.05$} & \underline{$0.47{\pm}0.11$} \\
Default & TabCausal & 500 & {\bfseries\boldmath $0.54{\pm}0.12$} & \underline{$0.62{\pm}0.14$} & \underline{$0.49{\pm}0.13$} & {\bfseries\boldmath $40.5{\pm}17.8$} & {\bfseries\boldmath $0.047{\pm}0.020$} & {\bfseries\boldmath $187.9{\pm}96.7$} & \underline{$0.85{\pm}0.08$} & {\bfseries\boldmath $0.56{\pm}0.14$} \\
\midrule
Fixed setpoint & CDIS & 500 & $0.40{\pm}0.12$ & $0.48{\pm}0.16$ & $0.35{\pm}0.10$ & $48.3{\pm}19.6$ & $0.056{\pm}0.022$ & $237.7{\pm}119.2$ & -- & -- \\
Fixed setpoint & GIES & 500 & \underline{$0.48{\pm}0.12$} & $0.43{\pm}0.14$ & {\bfseries\boldmath $0.56{\pm}0.12$} & $59.7{\pm}31.7$ & $0.069{\pm}0.036$ & $253.1{\pm}79.2$ & -- & -- \\
Fixed setpoint & IGSP & 500 & $0.40{\pm}0.12$ & $0.43{\pm}0.15$ & $0.38{\pm}0.11$ & $54.2{\pm}26.7$ & $0.062{\pm}0.031$ & $233.2{\pm}105.7$ & -- & -- \\
Fixed setpoint & SDCD & 500 & $0.38{\pm}0.09$ & $0.33{\pm}0.11$ & $0.48{\pm}0.11$ & $72.7{\pm}26.5$ & $0.084{\pm}0.030$ & $242.8{\pm}91.2$ & -- & -- \\
Fixed setpoint & AVICI & 500 & $0.38{\pm}0.12$ & {\bfseries\boldmath $0.65{\pm}0.15$} & $0.28{\pm}0.11$ & $43.7{\pm}17.1$ & $0.050{\pm}0.020$ & \underline{$189.3{\pm}109.8$} & $0.77{\pm}0.09$ & $0.41{\pm}0.12$ \\
Fixed setpoint & CauScale & 500 & $0.46{\pm}0.10$ & $0.57{\pm}0.11$ & $0.39{\pm}0.11$ & \underline{$43.0{\pm}18.6$} & \underline{$0.049{\pm}0.021$} & $198.7{\pm}106.2$ & \underline{$0.85{\pm}0.05$} & \underline{$0.47{\pm}0.11$} \\
Fixed setpoint & TabCausal & 500 & {\bfseries\boldmath $0.54{\pm}0.12$} & \underline{$0.61{\pm}0.14$} & \underline{$0.49{\pm}0.13$} & {\bfseries\boldmath $40.6{\pm}17.7$} & {\bfseries\boldmath $0.047{\pm}0.020$} & {\bfseries\boldmath $186.6{\pm}94.4$} & {\bfseries\boldmath $0.85{\pm}0.08$} & {\bfseries\boldmath $0.55{\pm}0.14$} \\
\midrule
Parameter shift & CDIS & 500 & $0.40{\pm}0.12$ & $0.48{\pm}0.16$ & $0.35{\pm}0.10$ & $48.3{\pm}19.6$ & $0.055{\pm}0.023$ & $240.0{\pm}120.0$ & -- & -- \\
Parameter shift & GIES & 500 & \underline{$0.43{\pm}0.11$} & $0.39{\pm}0.12$ & {\bfseries\boldmath $0.50{\pm}0.10$} & $64.1{\pm}32.1$ & $0.074{\pm}0.037$ & $279.8{\pm}79.1$ & -- & -- \\
Parameter shift & IGSP & 500 & $0.39{\pm}0.11$ & $0.43{\pm}0.15$ & $0.38{\pm}0.10$ & $54.4{\pm}27.0$ & $0.062{\pm}0.031$ & $235.9{\pm}107.2$ & -- & -- \\
Parameter shift & SDCD & 500 & $0.38{\pm}0.09$ & $0.33{\pm}0.11$ & $0.47{\pm}0.11$ & $73.2{\pm}25.5$ & $0.084{\pm}0.029$ & $244.8{\pm}88.5$ & -- & -- \\
Parameter shift & AVICI & 500 & $0.37{\pm}0.13$ & {\bfseries\boldmath $0.62{\pm}0.15$} & $0.27{\pm}0.11$ & \underline{$44.6{\pm}17.7$} & \underline{$0.051{\pm}0.020$} & \underline{$192.5{\pm}109.1$} & $0.76{\pm}0.09$ & $0.39{\pm}0.13$ \\
Parameter shift & CauScale & 500 & $0.42{\pm}0.10$ & $0.52{\pm}0.12$ & $0.36{\pm}0.11$ & $45.2{\pm}18.8$ & $0.052{\pm}0.022$ & $209.8{\pm}106.1$ & \underline{$0.84{\pm}0.06$} & \underline{$0.41{\pm}0.11$} \\
Parameter shift & TabCausal & 500 & {\bfseries\boldmath $0.52{\pm}0.12$} & \underline{$0.59{\pm}0.15$} & \underline{$0.47{\pm}0.13$} & {\bfseries\boldmath $41.8{\pm}18.0$} & {\bfseries\boldmath $0.048{\pm}0.021$} & {\bfseries\boldmath $191.4{\pm}94.1$} & {\bfseries\boldmath $0.85{\pm}0.08$} & {\bfseries\boldmath $0.53{\pm}0.15$} \\
\midrule
Multi-target & CDIS & 500 & $0.38{\pm}0.12$ & $0.47{\pm}0.16$ & $0.33{\pm}0.10$ & $48.9{\pm}19.4$ & $0.056{\pm}0.022$ & $240.6{\pm}120.8$ & -- & -- \\
Multi-target & GIES & 500 & \underline{$0.48{\pm}0.12$} & $0.44{\pm}0.14$ & {\bfseries\boldmath $0.56{\pm}0.11$} & $59.3{\pm}31.2$ & $0.068{\pm}0.036$ & $252.8{\pm}79.0$ & -- & -- \\
Multi-target & IGSP & 500 & $0.13{\pm}0.19$ & $0.14{\pm}0.22$ & $0.12{\pm}0.19$ & $54.6{\pm}21.7$ & $0.063{\pm}0.025$ & $210.2{\pm}114.3$ & -- & -- \\
Multi-target & SDCD & 500 & $0.38{\pm}0.09$ & $0.34{\pm}0.11$ & $0.48{\pm}0.11$ & $72.4{\pm}25.9$ & $0.083{\pm}0.030$ & $243.8{\pm}93.4$ & -- & -- \\
Multi-target & AVICI & 500 & $0.39{\pm}0.12$ & {\bfseries\boldmath $0.64{\pm}0.14$} & $0.28{\pm}0.11$ & \underline{$43.7{\pm}17.2$} & \underline{$0.050{\pm}0.020$} & \underline{$189.9{\pm}109.7$} & $0.77{\pm}0.08$ & $0.41{\pm}0.12$ \\
Multi-target & CauScale & 500 & $0.44{\pm}0.10$ & $0.51{\pm}0.12$ & $0.39{\pm}0.11$ & $45.7{\pm}18.7$ & $0.053{\pm}0.021$ & $210.0{\pm}104.7$ & \underline{$0.84{\pm}0.06$} & \underline{$0.44{\pm}0.11$} \\
Multi-target & TabCausal & 500 & {\bfseries\boldmath $0.54{\pm}0.12$} & \underline{$0.62{\pm}0.14$} & \underline{$0.49{\pm}0.13$} & {\bfseries\boldmath $40.6{\pm}17.6$} & {\bfseries\boldmath $0.047{\pm}0.020$} & {\bfseries\boldmath $186.3{\pm}97.5$} & {\bfseries\boldmath $0.85{\pm}0.08$} & {\bfseries\boldmath $0.55{\pm}0.14$} \\
\midrule
Dense mixed & CDIS & 500 & $0.36{\pm}0.10$ & $0.52{\pm}0.15$ & $0.29{\pm}0.09$ & $47.2{\pm}17.9$ & $0.054{\pm}0.021$ & $210.4{\pm}113.2$ & -- & -- \\
Dense mixed & GIES & 500 & \underline{$0.50{\pm}0.11$} & $0.45{\pm}0.13$ & {\bfseries\boldmath $0.57{\pm}0.11$} & $56.3{\pm}28.0$ & $0.065{\pm}0.032$ & $251.3{\pm}76.9$ & -- & -- \\
Dense mixed & IGSP & 500 & $0.25{\pm}0.16$ & $0.35{\pm}0.22$ & $0.20{\pm}0.13$ & $51.8{\pm}21.0$ & $0.060{\pm}0.024$ & $214.3{\pm}112.1$ & -- & -- \\
Dense mixed & SDCD & 500 & $0.40{\pm}0.09$ & $0.36{\pm}0.12$ & \underline{$0.49{\pm}0.11$} & $68.3{\pm}23.6$ & $0.079{\pm}0.027$ & $232.1{\pm}88.9$ & -- & -- \\
Dense mixed & AVICI & 500 & $0.37{\pm}0.12$ & {\bfseries\boldmath $0.64{\pm}0.15$} & $0.27{\pm}0.11$ & \underline{$44.3{\pm}17.1$} & \underline{$0.051{\pm}0.020$} & \underline{$190.3{\pm}111.1$} & $0.77{\pm}0.08$ & $0.40{\pm}0.11$ \\
Dense mixed & CauScale & 500 & $0.42{\pm}0.10$ & $0.50{\pm}0.13$ & $0.38{\pm}0.10$ & $46.7{\pm}18.8$ & $0.054{\pm}0.022$ & $211.4{\pm}104.3$ & \underline{$0.82{\pm}0.06$} & \underline{$0.41{\pm}0.11$} \\
Dense mixed & TabCausal & 500 & {\bfseries\boldmath $0.53{\pm}0.11$} & \underline{$0.61{\pm}0.13$} & $0.48{\pm}0.12$ & {\bfseries\boldmath $41.0{\pm}16.9$} & {\bfseries\boldmath $0.047{\pm}0.019$} & {\bfseries\boldmath $186.9{\pm}98.2$} & {\bfseries\boldmath $0.85{\pm}0.07$} & {\bfseries\boldmath $0.53{\pm}0.13$} \\
\midrule
High fraction & CDIS & 500 & $0.36{\pm}0.10$ & $0.51{\pm}0.15$ & $0.28{\pm}0.09$ & $47.6{\pm}17.7$ & $0.055{\pm}0.020$ & $211.5{\pm}114.9$ & -- & -- \\
High fraction & GIES & 500 & \underline{$0.51{\pm}0.12$} & $0.46{\pm}0.14$ & {\bfseries\boldmath $0.58{\pm}0.12$} & $56.7{\pm}29.9$ & $0.065{\pm}0.034$ & $238.2{\pm}82.5$ & -- & -- \\
High fraction & IGSP & 500 & $0.31{\pm}0.10$ & $0.43{\pm}0.15$ & $0.25{\pm}0.09$ & $52.0{\pm}22.4$ & $0.060{\pm}0.026$ & $224.1{\pm}109.5$ & -- & -- \\
High fraction & SDCD & 500 & $0.40{\pm}0.10$ & $0.35{\pm}0.12$ & $0.49{\pm}0.10$ & $70.3{\pm}25.5$ & $0.081{\pm}0.029$ & $237.3{\pm}93.0$ & -- & -- \\
High fraction & AVICI & 500 & $0.40{\pm}0.13$ & {\bfseries\boldmath $0.67{\pm}0.14$} & $0.30{\pm}0.12$ & $42.9{\pm}17.2$ & $0.049{\pm}0.020$ & \underline{$187.2{\pm}109.4$} & $0.77{\pm}0.09$ & $0.43{\pm}0.12$ \\
High fraction & CauScale & 500 & $0.48{\pm}0.10$ & $0.60{\pm}0.12$ & $0.41{\pm}0.11$ & \underline{$40.8{\pm}17.6$} & \underline{$0.047{\pm}0.020$} & $193.3{\pm}107.6$ & {\bfseries\boldmath $0.86{\pm}0.05$} & \underline{$0.51{\pm}0.11$} \\
High fraction & TabCausal & 500 & {\bfseries\boldmath $0.54{\pm}0.11$} & \underline{$0.62{\pm}0.13$} & \underline{$0.49{\pm}0.12$} & {\bfseries\boldmath $40.0{\pm}16.8$} & {\bfseries\boldmath $0.046{\pm}0.019$} & {\bfseries\boldmath $181.7{\pm}97.0$} & \underline{$0.86{\pm}0.07$} & {\bfseries\boldmath $0.56{\pm}0.13$} \\
\end{longtable}
\endgroup

%% file: tables/generated_results/ablation/construction_ablation_quality.tex
\begin{table}[!htbp]
\centering
\scriptsize
\setlength{\tabcolsep}{1.2pt}
\renewcommand{\arraystretch}{1.05}
\begin{threeparttable}
\caption{\textbf{Agentic construction ablation quality scores.}}
\label{tab:construction-ablation-quality}
\begin{tabular}{lcccccccccc}
\toprule
\textbf{Stage} & \textbf{Variant} & \textbf{Cases} & \textbf{Plaus.} & \textbf{Mech.} & \textbf{Bench.} & \textbf{Spec.} & \textbf{Single avg.} & \textbf{Struct. div.} & \textbf{Sem./mech. div.} & \textbf{Avg.} \\
\midrule
A0 & Seed only & 20 & 3.650\scorestd{0.462} & 3.375\scorestd{0.626} & 3.400\scorestd{0.447} & 3.375\scorestd{0.535} & 3.450\scorestd{0.492} & \textbf{4.500} & \underline{4.500} & 3.800 \\
A1 & + reference pack & 20 & 3.650\scorestd{0.328} & 3.275\scorestd{0.499} & 3.375\scorestd{0.358} & 3.425\scorestd{0.520} & 3.431\scorestd{0.398} & \textbf{4.500} & \textbf{5.000} & 3.871 \\
A2 & + planning & 20 & \underline{4.225\scorestd{0.302}} & \underline{4.275\scorestd{0.302}} & \underline{4.000\scorestd{0.363}} & \underline{4.225\scorestd{0.302}} & \underline{4.181\scorestd{0.291}} & \textbf{4.500} & \textbf{5.000} & \underline{4.371} \\
A3 & + graph review & 20 & \textbf{4.375\scorestd{0.275}} & \textbf{4.400\scorestd{0.262}} & \textbf{4.200\scorestd{0.377}} & \textbf{4.400\scorestd{0.262}} & \textbf{4.344\scorestd{0.262}} & \textbf{4.500} & \textbf{5.000} & \textbf{4.479} \\
\bottomrule
\end{tabular}
\begin{tablenotes}[flushleft]
\footnotesize
\item Scores are on a 1--5 scale judged by an external LLM evaluator. Single-graph criteria are mean\scorestd{standard deviation} over 20 cases. The final average also includes two batch-level diversity criteria. Higher is better; bold and underline mark the best and second-best values in each score column. Ties share the same mark.
\end{tablenotes}
\end{threeparttable}
\end{table}

%% file: cite.bib
@article{kalisch2005estimatinghighdimensionaldirectedacyclic,
  title   = {Estimating High-Dimensional Directed Acyclic Graphs with the {PC}-Algorithm},
  author  = {Markus Kalisch and Peter B{\"u}hlmann},
  journal = {Journal of Machine Learning Research},
  volume  = {8},
  number  = {22},
  pages   = {613--636},
  year    = {2007}
}

@inproceedings{wang2017permutationbasedcausalinferencealgorithms,
  author       = {Yuhao Wang and
                  Liam Solus and
                  Karren D. Yang and
                  Caroline Uhler},
  title        = {Permutation-based Causal Inference Algorithms with Interventions},
  booktitle    = {{NeurIPS}},
  year         = {2017}
}

@article{hauser2012characterizationgreedylearninginterventional,
  author       = {Alain Hauser and
                  Peter B{\"{u}}hlmann},
  title        = {Characterization and greedy learning of interventional {Markov} equivalence
                  classes of directed acyclic graphs},
  journal      = {Journal of Machine Learning Research},
  volume       = {13},
  pages        = {2409--2464},
  year         = {2012}
}

@article{colombo2013orderindependentconstraintbasedcausalstructure,
      author       = {Diego Colombo and
                  Marloes H. Maathuis},
  title        = {Order-independent constraint-based causal structure learning},
  journal      = {Journal of Machine Learning Research},
  volume       = {15},
  number       = {1},
  pages        = {3741--3782},
  year         = {2014}
}

@inproceedings{zheng2018dagstearscontinuousoptimization,
  author       = {Xun Zheng and
                  Bryon Aragam and
                  Pradeep Ravikumar and
                  Eric P. Xing},
  title        = {{DAGs} with {NO} {TEARS}: Continuous Optimization for Structure Learning},
  booktitle    = {{NeurIPS}},
  year         = {2018}
}

@inproceedings{zheng2020learning,
  author       = {Xun Zheng and
                  Chen Dan and
                  Bryon Aragam and
                  Pradeep Ravikumar and
                  Eric P. Xing},
  title        = {Learning Sparse Nonparametric {DAGs}},
  booktitle    = {{AISTATS}},
  year         = {2020}
}

@article{buhlmann2014cam,
  author  = {Peter B{\"u}hlmann and Jonas Peters and Jan Ernest},
  title   = {{CAM}: Causal Additive Models, High-dimensional Order Search and Penalized Regression},
  journal = {The Annals of Statistics},
  volume  = {42},
  number  = {6},
  pages   = {2526--2556},
  year    = {2014}
}

@inproceedings{sethuraman2023nodagsflownonlinearcycliccausal,
  author       = {Muralikrishnna G. Sethuraman and
                  Romain Lopez and
                  Rahul Mohan and
                  Faramarz Fekri and
                  Tommaso Biancalani and
                  Jan-Christian H{\"{u}}tter},
  title        = {{NODAGS-Flow}: Nonlinear Cyclic Causal Structure Learning},
  booktitle    = {{AISTATS}},
  pages        = {6371--6387},
  year         = {2023}
}

@inproceedings{lorch2022amortizedinferencecausalstructure,
  author       = {Lars Lorch and
                  Scott Sussex and
                  Jonas Rothfuss and
                  Andreas Krause and
                  Bernhard Sch{\"{o}}lkopf},
  title        = {Amortized Inference for Causal Structure Learning},
  booktitle    = {{NeurIPS}},
  year         = {2022}
}

@book{Peters2017,
author = {Jonas Peters and Dominik Janzing and Bernhard Sch{\"o}lkopf},
title = {Elements of Causal Inference: Foundations and Learning Algorithms},
publisher = {MIT Press},
year = {2017}
}

@inproceedings{dai2025selectionmeetsinterventionadditional,
  author       = {Haoyue Dai and
                  Ignavier Ng and
                  Jianle Sun and
                  Zeyu Tang and
                  Gongxu Luo and
                  Xinshuai Dong and
                  Peter Spirtes and
                  Kun Zhang},
  title        = {When Selection Meets Intervention: Additional Complexities in Causal
                  Discovery},
  booktitle    = {{ICLR}},
  year         = {2025}
}

@inproceedings{brouillard2020differentiablecausaldiscoveryinterventional,
  author       = {Philippe Brouillard and
                  S{\'{e}}bastien Lachapelle and
                  Alexandre Lacoste and
                  Simon Lacoste-Julien and
                  Alexandre Drouin},
  title        = {Differentiable Causal Discovery from Interventional Data},
  booktitle    = {{NeurIPS}},
  year         = {2020}
}

@article{sachs,
author = {Karen Sachs and Omar Perez and Dana Pe'er and Douglas A. Lauffenburger and Garry P. Nolan},
title = {Causal Protein-Signaling Networks Derived from Multiparameter Single-Cell Data},
journal = {Science},
volume = {308},
number = {5721},
pages = {523--529},
year = {2005}
}

@article{zhao2023causalstar,
  author  = {Boxiang Zhao and Shuliang Wang and Lianhua Chi and Qi Li and Xiaojia Liu and Jing Geng},
  title   = {Causal Discovery via Causal Star Graphs},
  journal = {ACM Transactions on Knowledge Discovery from Data},
  volume  = {17},
  number  = {7},
  pages   = {98:1--98:24},
  year    = {2023},
}

@inproceedings{hardt2024petshop,
  author    = {Michaela Hardt and William Roy Orchard and Patrick Bl{\"o}baum and Elke Kirschbaum and Shiva Kasiviswanathan},
  title     = {The {PetShop} Dataset: Finding Causes of Performance Issues across Microservices},
  booktitle = {{CLeaR}},
  pages     = {957--978},
  year      = {2024}
}

@article{lingam,
  author  = {Shohei Shimizu and Patrik O. Hoyer and Aapo Hyv{\"a}rinen and Antti Kerminen},
  title   = {A Linear Non-Gaussian Acyclic Model for Causal Discovery},
  journal = {Journal of Machine Learning Research},
  year    = {2006},
  volume  = {7},
  number  = {72},
  pages   = {2003--2030}
}

@book{books/spirtes2000causation,
  title={Causation, Prediction, and Search},
  author={Peter Spirtes and Clark N. Glymour and Richard Scheines},
  year={2000},
  publisher={MIT Press},
  edition={2}
}

@inproceedings{conf/nips/HoyerJMPS08,
  author    = {Patrik O. Hoyer and
               Dominik Janzing and
               Joris M. Mooij and
               Jonas Peters and
               Bernhard Sch{\"{o}}lkopf},
  title     = {Nonlinear causal discovery with additive noise models},
  booktitle = {{NeurIPS}},
  year      = {2008}
}

@inproceedings{NEURIPS2021_e987eff4,
 author = {Reisach, Alexander and Seiler, Christof and Weichwald, Sebastian},
 booktitle = {{NeurIPS}},
 title = {Beware of the Simulated {DAG}! {Causal} Discovery Benchmarks May Be Easy to Game},
 year = {2021}
}

@InProceedings{pmlr-v213-montagna23b,
  title = 	 {Scalable Causal Discovery with Score Matching},
  author =       {Montagna, Francesco and Noceti, Nicoletta and Rosasco, Lorenzo and Zhang, Kun and Locatello, Francesco},
  booktitle = 	 {{CLeaR}},
  year = 	 {2023}
}

@article{wu2025sampleestimateaggregaterecipe,
  title   = {Sample, Estimate, Aggregate: A Recipe for Causal Discovery Foundation Models},
  author  = {Menghua Wu and Yujia Bao and Regina Barzilay and Tommi Jaakkola},
  journal = {Transactions on Machine Learning Research},
  year    = {2025}
}

@inproceedings{NEURIPS2022_36e2967f,
 author = {Kevin Bello and Bryon Aragam and Pradeep Ravikumar},
 title = {{DAGMA}: Learning {DAGs} via M-matrices and a Log-Determinant Acyclicity Characterization},
 booktitle = {{NeurIPS}},
 year = {2022}
}

@inproceedings{nazaret2024stabledifferentiablecausaldiscovery,
  title     = {Stable Differentiable Causal Discovery},
  author    = {Achille Nazaret and Justin Hong and Elham Azizi and David Blei},
  booktitle = {{ICML}},
  pages     = {37413--37445},
  year      = {2024}
}

@misc{thompson2026arrowfoundationmodelcausal,
  title        = {Arrow: A Foundation Model for Causal Discovery},
  author       = {Ryan Thompson and He Zhao and Daniel M. Steinberg and Edwin V. Bonilla},
  year         = {2026},
  eprint       = {2605.07204},
  archivePrefix= {arXiv},
  primaryClass = {cs.LG},
  note         = {arXiv preprint arXiv:2605.07204}
}

@inproceedings{peng2026causcaleneuralcausaldiscovery,
  title     = {{CauScale}: Neural Causal Discovery at Scale},
  author    = {Bo Peng and Sirui Chen and Jiaguo Tian and Yu Qiao and Chaochao Lu},
  booktitle = {{ICML}},
  year      = {2026},
  eprint    = {2602.08629},
  archivePrefix = {arXiv},
  primaryClass  = {cs.LG}
}

@article{mooij2016causeeffectpairs,
  title   = {Distinguishing Cause from Effect Using Observational Data: Methods and Benchmarks},
  author  = {Mooij, Joris M. and Peters, Jonas and Janzing, Dominik and Zscheischler, Jakob and Sch{\"o}lkopf, Bernhard},
  journal = {Journal of Machine Learning Research},
  volume  = {17},
  number  = {32},
  pages   = {1--102},
  year    = {2016}
}

@article{rios2025benchpress,
  title   = {{Benchpress}: A Versatile Platform for Structure Learning in Causal and Probabilistic Graphical Models},
  author  = {Rios, Felix L. and Moffa, Giusi and Kuipers, Jack},
  journal = {Journal of Statistical Software},
  volume  = {114},
  number  = {12},
  pages   = {1--43},
  year    = {2025},
}

@misc{geffner2022deci,
  title         = {Deep End-to-end Causal Inference},
  author        = {Tomas Geffner and Javier Antoran and Adam Foster and Wenbo Gong and Chao Ma and Emre Kiciman and Amit Sharma and Angus Lamb and Martin Kukla and Nick Pawlowski and Miltiadis Allamanis and Cheng Zhang},
  year          = {2022},
  eprint        = {2202.02195},
  archivePrefix = {arXiv},
  primaryClass  = {cs.LG},
  note          = {arXiv preprint arXiv:2202.02195}
}

@misc{panayiotou2025causalprofiler,
  title         = {{CausalProfiler}: Generating Synthetic Benchmarks for Rigorous and Transparent Evaluation of Causal Machine Learning},
  author        = {Panayiotou, Panayiotis and Poinsot, Audrey and Leite, Alessandro and Chesneau, Nicolas and Schoenauer, Marc and {\c{S}}im{\c{s}}ek, {\"O}zg{\"u}r},
  year          = {2025},
  eprint        = {2511.22842},
  archivePrefix = {arXiv},
  primaryClass  = {cs.LG},
  note          = {arXiv preprint arXiv:2511.22842}
}

@inproceedings{herdeanu2025causaldynamics,
  title     = {{CausalDynamics}: A Large-scale Benchmark for Structural Discovery of Dynamical Causal Models},
  author    = {Benjamin Herdeanu and Juan Nathaniel and Carla Roesch and Jatan Buch and Gregor Ramien and Johannes Haux and Pierre Gentine},
  booktitle = {{NeurIPS}},
  year      = {2025}
}

@inproceedings{babakov2025causalgraphbench,
  title     = {{CausalGraphBench}: a Benchmark for Evaluating Language Models capabilities of Causal Graph discovery},
  author    = {Babakov, Nikolay and Reiter, Ehud and Bugar{\'i}n, Alberto},
  booktitle = {{ACL}},
  pages     = {240--258},
  year      = {2025},
}

@inproceedings{stein2025causalrivers,
  title     = {{CausalRivers} - Scaling up Benchmarking of Causal Discovery for Real-World Time-Series},
  author    = {Gideon Stein and Maha Shadaydeh and Jan Blunk and Niklas Penzel and Joachim Denzler},
  booktitle = {{ICLR}},
  year      = {2025}
}

@article{chevalley2025causalbench,
  title   = {A large-scale benchmark for network inference from single-cell perturbation data},
  author  = {Chevalley, Mathieu and Roohani, Yusuf H. and Mehrjou, Arash and Leskovec, Jure and Schwab, Patrick},
  journal = {Communications Biology},
  volume  = {8},
  number  = {1},
  pages   = {412},
  year    = {2025},
}

@misc{li2026tabcausal,
  title         = {{TabCausal}: Pretraining Across Causal Environments for Tabular Causal Discovery},
  author        = {Li, Zi-Rong and Liu, Si-Yang and Wang, Tian-Zuo and Ye, Han-Jia},
  year          = {2026},
  eprint        = {2605.31156},
  archivePrefix = {arXiv},
  primaryClass  = {cs.LG},
  note          = {arXiv preprint arXiv:2605.31156}
}

@misc{qiao2026cdfm,
  title         = {{CDFM}: Towards a General-Purpose Causal Discovery Foundation Model},
  author        = {Jie Qiao and Ruichu Cai and Zijian Li and Weilin Chen and Pengfei Hua and Boyan Xu and Zhengming Chen and Zhifeng Hao and Peng Cui},
  year          = {2026},
  eprint        = {2607.11508},
  archivePrefix = {arXiv},
  primaryClass  = {cs.LG},
  note          = {arXiv preprint arXiv:2607.11508}
}

@misc{bloebaum2026foundcause,
  title         = {{FoundCause}: Causal Discovery with Latent Confounders from Observational Data},
  author        = {Patrick Bl{\"o}baum and Krishnakumar Balasubramanian and Shiva Prasad Kasiviswanathan},
  year          = {2026},
  eprint        = {2606.17516},
  archivePrefix = {arXiv},
  primaryClass  = {cs.LG},
  note          = {arXiv preprint arXiv:2606.17516}
}

@article{scutari2010bnlearn,
  title   = {Learning {Bayesian} Networks with the {bnlearn} {R} Package},
  author  = {Scutari, Marco},
  journal = {Journal of Statistical Software},
  volume  = {35},
  number  = {3},
  pages   = {1--22},
  year    = {2010},
}

@misc{zhou2024ocdb,
  title         = {{OCDB}: Revisiting Causal Discovery with a Comprehensive Benchmark and Evaluation Framework},
  author        = {Wei Zhou and Hong Huang and Guowen Zhang and Ruize Shi and Kehan Yin and Yuanyuan Lin and Bang Liu},
  year          = {2024},
  eprint        = {2406.04598},
  archivePrefix = {arXiv},
  primaryClass  = {cs.LG},
  note          = {arXiv preprint arXiv:2406.04598}
}

@inproceedings{herman2025uumc,
  title     = {Unitless Unrestricted {Markov}-Consistent {SCM} Generation: Better Benchmark Datasets for Causal Discovery},
  author    = {Rebecca J. Herman and Jonas Wahl and Urmi Ninad and Jakob Runge},
  booktitle = {{CLeaR}},
  pages     = {1506--1531},
  year      = {2025}
}

@article{gamella2024causalchambers,
  title   = {Causal Chambers as a Real-World Physical Testbed for {AI} Methodology},
  author  = {Juan L. Gamella and Jonas Peters and Peter B{\"u}hlmann},
  journal = {Nature Machine Intelligence},
  volume  = {7},
  pages   = {107--118},
  year    = {2025},
}

@inproceedings{cheng2024causaltime,
  title     = {{CausalTime}: Realistically Generated Time-series for Benchmarking of Causal Discovery},
  author    = {Cheng, Yuxiao and Wang, Ziqian and Xiao, Tingxiong and Zhong, Qin and He, Jinliang},
  booktitle = {{ICLR}},
  year      = {2024}
}

@inproceedings{ferdous2025timegraph,
  title     = {{TimeGraph}: Synthetic Benchmark Datasets for Robust Time-Series Causal Discovery},
  author    = {Muhammad Hasan Ferdous and Emam Hossain and Md Osman Gani},
  booktitle = {{KDD}},
  pages     = {5425--5435},
  year      = {2025},
}

@inproceedings{gobler2024causalassembly,
  title     = {{causalAssembly}: Generating Realistic Production Data for Benchmarking Causal Discovery},
  author    = {Konstantin G{\"o}bler and Tobias Windisch and Mathias Drton and Tim Pychynski and Martin Roth and Steffen Sonntag},
  booktitle = {{CLeaR}},
  pages     = {609--642},
  year      = {2024}
}

@inproceedings{wan2025llmcdsurvey,
  title     = {Large Language Models for Causal Discovery: Current Landscape and Future Directions},
  author    = {Wan, Guangya and Lu, Yunsheng and Wu, Yuqi and Hu, Mengxuan and Li, Sheng},
  booktitle = {{IJCAI}},
  pages     = {10687--10695},
  year      = {2025},
}

@article{zhang2025promptbn,
  title   = {Bayesian Network Structure Discovery Using Large Language Models},
  author  = {Yinghuan Zhang and Yufei Zhang and Parisa Kordjamshidi and Zijun Cui},
  journal = {Transactions on Machine Learning Research},
  year    = {2026},
  note    = {Also available as arXiv:2511.00574}
}
